%% file: main.tex
\documentclass[10pt,twocolumn,letterpaper]{article}

\usepackage{iccv}              

\input{preamble}

\definecolor{iccvblue}{rgb}{0.21,0.49,0.74}
\usepackage[pagebackref,breaklinks,colorlinks,allcolors=iccvblue]{hyperref}
\usepackage{stfloats}

\usepackage[numbers]{natbib}
\usepackage[resetlabels]{multibib}
\def\paperID{ } 
\def\confName{CVPR}
\def\confYear{2026}

\title{Differentiable Adaptive 4D Structured Illumination for Joint Capture of Shape and Reflectance}

\author{Huakeng Ding\textsuperscript{1} \quad 
Yaowen Chen\textsuperscript{1} \quad 
Kun Zhou\textsuperscript{1,2}\addtocounter{footnote}{1}\thanks{Corresponding authors: \{kunzhou,hwu\}@acm.org.} \quad 
Hongzhi Wu\textsuperscript{1}\footnotemark[2]\\
\textsuperscript{1}State Key Lab of CAD\&CG, Zhejiang University \\
\textsuperscript{2}Hangzhou Research Institute of Holographic and AI Technology\\
}

\begin{document}

\maketitle

\input{abstract}    
\input{adaptive}
\textbf{Acknowledgements.} We would like to thank Fan Pei, Kaizhang Kang, and Xianmin Xu for their help.
This work is partially supported by NSF China (62227806, 62332015, \& 62421003), the XPLORER PRIZE, Information Technology Center, State Key Lab of CAD\&CG, Zhejiang University and a gift from Adobe.
{
    \small
    \bibliographystyle{ieeenat_fullname}
    \bibliography{main}
}


\end{document}

%% file: preamble.tex
\usepackage{amsmath}


%% file: abstract.tex
\begin{abstract}
We present a differentiable framework to adaptively compute 4D illumination conditions with respect to an object, for efficient, high-quality simultaneous acquisition of its shape and reflectance, with a unified spatial-angular structured light and a single camera. Using a simple histogram-based pixel-level probability model for depth and reflectance, we differentiably link the next illumination condition(s) with a loss that encourages the reduction in depth uncertainty. As new structured illumination is cast, corresponding image measurements are used to update the uncertainty at each pixel. Finally, a fine-tuning-based approach reconstructs the depth map and reflectance parameter maps, by minimizing the differences between all physical measurements and their simulated counterparts. The effectiveness of our framework is demonstrated on physical objects with wide variations in shape and appearance. Our depth results compare favorably with state-of-the-art techniques, while our reflectance results are comparable when validated against photographs.
\end{abstract}

%% file: adaptive.tex
\section{Introduction}
Capturing the geometry and appearance of physical objects is a key problem in computer vision and graphics. Often represented as a 3D mesh and a 6D spatially-varying bidirectional reflectance distribution function (SVBRDF), the reconstruction result enables photo-realistic image synthesis under any view and lighting conditions. This benefits a wide range of applications, including cultural heritage, e-commerce, special effects and video games.

Active, structured illumination is commonly employed in high-quality acquisition of geometry \emph{or} reflectance. It probes the physical domain to obtain measurements strongly correlated with target information, resulting in a high signal-to-noise ratio (SNR). For shape, spatially-varying light pattern(s) are projected into the 3D space to distinguish rays for triangulation~\cite{gupta_2012_mpsSL,Schar_2003_triangulationSL}. For reflectance, angular-domain illumination patterns are cast onto the object surface, physically convolving with BRDF slices at different locations. Accurate appearance can be deduced from corresponding measurements~\cite{Ghosh_2009_MultiplexSphericalLight,Paul_2013_UnifiedArc}.

\input{figs/prototype}

To facilitate \emph{active} capture of both shape \emph{and} reflectance, the first 4D structured light is recently proposed in~\cite{xxm_2023_unified}, unifying the aforementioned two types of illumination in a compact form factor. Despite its physical sampling capability in the spatial-angular domain, the acquisition efficiency is not satisfactory: it takes 24 minutes to scan a single view, the majority of which is spent on geometry capture, with only one LED on at a time. Moreover, the illumination conditions in~\cite{xxm_2023_unified} are pre-optimized, which could be \emph{suboptimal} for acquiring a specific object.





To actively capture shape and reflectance at the same time with \emph{high efficiency and quality}, we propose a differentiable framework to automatically optimize the spatial-angular lighting conditions during acquisition, which are tailored to the object being digitized. This is achieved by differentiably connecting the next lighting condition(s) to a loss that minimizes the overall depth uncertainty. By using multiple LEDs instead of one at a time as in previous work, we reduce the exposure time by up to 100$\times$ and the total acquisition time by 2$\times$. After acquisition, we further fine-tune the results by minimizing the differences between physical measurements and simulated ones. The final output is a depth map and multiple texture maps that store GGX BRDF parameters.

The main contributions of this paper are as follows:
\begin{itemize}
    \item We propose a learned multiplexing scheme for 4D spatial-angular structured illumination, to simultaneously capture shape and reflectance with high efficiency.
    \item We propose a differentiable framework to on-the-fly optimize complex lighting conditions, which are \emph{adaptive} to an object of interest.
\end{itemize}

\section{Related Work}
This section reviews representative work on structured illumination for geometry and/or reflectance, as well as adaptive acquisition, which are two topics most related to our paper. Passive approaches are not discussed here. Interested readers are directed to excellent surveys~\cite{DONG201959survey,SALVI2004827survey,Guarnera2016BRDFRAsurvey,10.5555/3059320.305932_survey,https://doi.org/10.1111/cgf.15199survey} for a broader view of the topic.

\subsection{Structured Illumination}

\textbf{Spatial Lighting.} Laser-stripe triangulation~\cite{levoy2000scanSL} and spatial structured lighting~\cite{gupta_2012_mpsSL,kutu_2018_optimalSL,Schar_2003_triangulationSL} are widely used for capturing highly accurate geometry. These techniques  project one or more spatially distinctive, discrete or continuous patterns onto the object surface, encoding light rays to facilitate 3D triangulation. Different pattern designs have been explored to enhance robustness~\cite{moreno_2015_robustSL,gupta_2012_mpsSL}, improve computational efficiency~\cite{Fanello_2016_efficientSL,Fanello_2017_efficientSL,ZhaHb_2024_NRefficientSL}, and increase acquisition speed~\cite{Koppal_2011_speedupSL}. While the spatial structures in illumination are historically hand-crafted, recent work demonstrates enhanced quality and efficiency with automatically designed ones~\cite{kutu_2020_autoSL,zyx_2024_volumeSI}. 

\textbf{Angular Lighting.} 
Despite its quality, exhaustively sampling the physical appearance domain with a gantry is prohibitively time-consuming~\cite{dana1999reflectance,lawrence2006inverse}. On the other hand, illumination multiplexing is a highly successful class of approaches, which modulates the intensities of lights at multiple angles, to reconstruct reflectance based on measurements taken under different lighting conditions. The lightstage recovers appearance using a pre-computed inverse lookup table~\cite{Ghosh_2009_MultiplexSphericalLight}. Planar SVBRDF is estimated by analyzing appearance changes relative to a moving linear light source~\cite{Tong_2014_linearlightBRDF,Gard_2003_LinearLightAppr}. Isotropic reflectance can be captured via a frequency domain analysis, utilizing an LCD panel as the light source~\cite{aittala_2013_frequency_SVBRDF}. Similar to spatial illumination, recent trend in angular lighting moves from manual to automatic design of light patterns, substantially increasing the acquisition efficiency~\cite{kang_2018_autoencoder,Ma_2021_FreeScan,ma2023svbrdf,Choi2024DifferentiablePS}.

The closest work to ours is~\cite{xxm_2023_unified}, which proposes a unified structured light in both spatial and angular domain, with an LED array and an LCD mask. Despite hardware novelty, the prototype \emph{separately} captures geometry (acting like a conventional projector) and then appearance (like a lightstage), with pre-optimized 2D lighting conditions. In comparison, we propose the first learned multiplexing approach for 4D spatial-angular domain to \emph{simultaneously} capture shape and reflectance, with an improved efficiency in both the acquisition time and the number of input photographs. We also establish a differentiable adaptive framework to optimize lighting conditions that are tailored to an object.

\subsection{Adaptive Acquisition}

\textbf{Geometry Capture.} 
Real-time generation of a parametric~\cite{Koninckx_2006_realtimeRangeAcquisition} or code-based~\cite{Maurice_2013_RealtimeCoding} single-shot spatial pattern can be performed on-the-fly with respect to scene content. In~\cite{LiQiang_2011_DenseDepth_Estiamtion}, structured patterns with adaptive color is computed using principal component analysis of the scene image with a projector and two cameras. Adaptation strategy for focus and exposure is proposed in~\cite{ZHANG2014robustDepthSensing} to generate corresponding pattern for robust depth sensing. In~\cite{Rosman_2016_Information-Driven}, the next pattern is selected from an pre-defined set to maximize information gain. Recently, another class of approaches optimize \emph{subject-independent} structured patterns are optimized to adapt to a specific hardware configuration~\cite{kutu_2018_optimalSL,kutu_2020_autoSL}.


\textbf{Appearance Capture.} 
Lensch et al.~\cite{Lensch2003PlannedBRDF} propose a function to measure the reduction in
uncertainty added by one view, to guide view planning for SVBRDF acquisition. In~\cite{Fuchs2007AdaptiveReflect}, a 
sampling algorithm is developed to incrementally choose light directions adapted to the properties of a reflectance field. Exploiting a database prior, an adaptive method of accurate interpolation of sparsely measured BRDF is introduced~\cite{filip_2013_adaptiveBRDFslice}.
A sampling scheme is derived from a new BRDF parameterization that automatically adapts to the behavior of a material~\cite{Dupuy_2018_adaptiveParam}. Liu et al.~\cite{Liu_2023_learn_to_sample} apply meta-learning to optimize the physical sampling pattern. Stochastic particle-optimization sampling is adopted in~\cite{egsr_2024_uncertainty} to sample uncertain material parameters to guide acquisition process. Note that a recent approach learns a divide-and-conquer strategy for reflectance reconstruction, with fixed illumination patterns~\cite{Ma_2024_AdaptiveReflect}. 

Unlike the majority existing work in this category, we propose a \emph{differentiable} framework to automatically compute high-dimensional structured illumination for \emph{simultaneous} acquisition of both shape and reflectance.

\section{Hardware}
\label{sec:hardware}
We use a single-camera acquisition prototype, similar to the one proposed in~\cite{xxm_2023_unified}.
The spatial-angular structured light includes a rectangular RGB LED array (with 64$\times$48 = 3,072 LEDs) and an LCD mask (with a spatial resolution of 1,920$\times$1,080). A 45MP Canon EOS R5 camera is mounted above the mask. We use a focal length of 24mm and an aperture of f/22. A valid volume is defined as a cube of 15cm$\times$15cm$\times$15cm, whose center is 15cm in front of the center of the mask. All intrinsic and extrinsic parameters of cameras and lights are differentiably calibrated in an end-to-end fashion (detailed in the supplemental material). Following existing terminology, we refer to the intensity distribution of the LED array as a \textbf{light pattern}, and the image displayed on the LCD as a \textbf{mask pattern}. Please refer to~\cite{xxm_2023_unified} for more details.

\section{Preliminaries}
Below we describe the relationship between a pixel measurement, the depth/reflectance corresponding to that pixel and a light/mask pattern, which will be used to drive differentiable optimization described in the next section.
\begin{align}
I_{j,k}
=&\sum_{l} \int_{A}L_{j}(\mathbf{x}_{l}, -\omega^i_{k})M_{j}(\mathbf{x}_{l}\leftrightarrow\mathbf{x}_{k}) f_{k,l} FdA, \nonumber \\
\approx &\sum_{l} f_{k,l} F\int_{A}L_{j}(\mathbf{x}_{l}, -\omega^i_{k}) M_{j}(\mathbf{x}_{l}\leftrightarrow\mathbf{x}_{k})dA, \nonumber \\
\approx &\sum_{l} f_{k,l} FL_{j}(l)\Psi(-\omega^i_{k})\int_{A}L(\mathbf{x}_{l})M_{j}(\mathbf{x}_{l}\leftrightarrow\mathbf{x}_{k}^{p})dA.
\label{eq:render}
\end{align}
Here $I_{j,k}$ is an image measurement at a pixel $k$ under $j$-th light/mask pattern. \(\mathbf{x}_l\) is a point on an LED with an index of \(l\), modeled as an area light of 2mm \(\times\) 2mm according to calibration. \(\mathbf{x}_k\) is the 3D position corresponding to the current pixel. The lighting direction $\omega^i_{k}$ is a unit vector pointing from  $\mathbf{x}_k$ to $\mathbf{x}_l$, and the view direction $\omega^o_{k}$ from $\mathbf{x}_k$ to the center of the camera. Moreover, $L_{j}(\mathbf{x}_{l}, -\omega^i_{k})$ is the radiance emitted from the LED $l$ along the direction $-\omega^i_{k}$. $M_{j}(\mathbf{x}_{l}\leftrightarrow\mathbf{x}_{k})$ is the mask value, where the ray from $x_l$ to $x_k$ intersects our liquid crystal panel. In addition, $f_{k,l}$ is a BRDF value for $(\omega^i_{k},\omega^o_{k})$. We use parametric GGX model~\cite{Walter2007GGXmodel} in this paper, while other models can also be plugged in here. \(F = \frac{(\omega^i_k\cdot\mathbf{n}_k)^+(-\omega^i_k\cdot\mathbf{n}_l)^+}{||\mathbf{x}_l-\mathbf{x}_k||^2}\) is the form factor, where $\mathbf{n}_k/\mathbf{n}_l$ is the surface normal of $\mathbf{x}_k/\mathbf{x}_l$, respectively. 
The above integral is calculated over the surface area $A$ of the LED. Due to the small solid angle subtended by $A$ with respect to $\mathbf{x}_k$, we assume constant $f_r/F/\omega^i_{k}$ in the integral, and factor $L$ as:
\begin{equation}
L_{j}(\mathbf{x}_{l}, -\omega^i_{k}) \approx L_{j}(l)\Psi(-\omega^i_{k})L(\mathbf{x}_{l}), 
\end{equation}
where $L_{j}(l)$ is the relative intensity of the LED $l$ in the $j$-th light pattern, in the range of [0,1]; $\Psi$ is a pre-calibrated angular distribution function for an LED; and $L(\mathbf{x}_{l})$ is implemented as a $5\times5$ kernel, satisfying $\int_A L(\mathbf{x}_{l}) dA = 1$, whose values are also pre-calibrated, similar to~\cite{xxm_2023_unified}.




\section{Overview}
Our pipeline consists of two stages: differentiable adaptive acquisition and fine-tuning. First, for a physical object, we compute the next light/mask pattern(s) by minimizing depth uncertainty, take photograph(s) with these patterns, update the uncertainty measure with new measurements, and repeat this process until a termination condition is met~(\cref{sec:capture}). Next, we use the depth/reflectance estimate in the previous stage as initial values, and fine-tune the results by minimizing the differences between physical measurements and corresponding simulated ones~(\cref{sec:reconstruction}). The final output is a depth map and several texture maps that store parameters of the GGX BRDF model. Similar to the majority of related work, we focus on high-quality reconstruction from a single view, and leave the scanning of a complete 3D object to orthogonal techniques. Please refer to~\cref{fig:pipeline} for a graphical illustration of the pipeline, and~\cref{fig:pattern_visualize} for an example. 
\input{figs/pipeline}

\section{Differentiable Adaptive Acquisition}
\label{sec:capture}
To quantitatively measure uncertainty, we first build simple histogram-based probability models for depth and BRDF parameters at a valid pixel~(\cref{sec:model}). During adaptive acquisition, we differentiably optimize the next light/mask pattern(s), by minimizing the uncertainty computed from our models~(\cref{sec:pattern}). We then update the models with the new measurements under optimized patterns~(\cref{sec:model}). This process is repeated until 72 light/mask patterns are projected in our experiments, while other termination conditions can also be employed here.

\subsection{Probability Model}
\label{sec:model}

\textbf{Definition.} For a \emph{single} pixel, we assume that the joint distribution of its depth and reflectance can be modeled as independent probability models for each individual parameter (i.e., depth/BRDF parameters) based on probability mass functions defined with histograms. A \textbf{candidate} has its depth and BRDF parameters independently sampled from these probability models.

\begin{figure}
    \centering
\includegraphics[width=\linewidth]{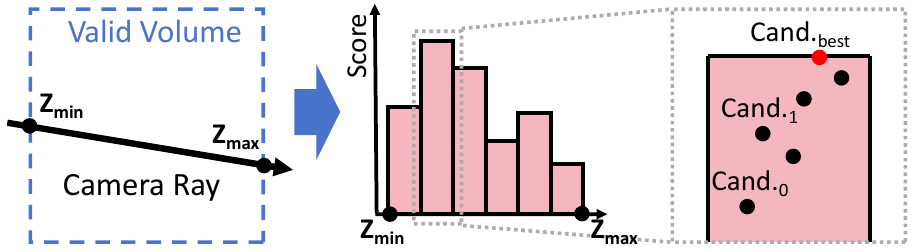}
    \caption{Graphical illustration of our probability model for depth. To build this model, we first determine its range $z_{\operatorname{min}}/z_{\operatorname{max}}$ by intersecting the valid volume with a corresponding camera ray. This range is then split into $n_{\operatorname{bin}}$ bins. Each bin stores the highest ZNCC score, computed between physical measurements and simulated ones from each candidate. Cand. = candidate.}
    \label{fig:prob}
\end{figure}

Specifically, to build a probability model for depth, we first determine its minimal and maximal value by intersecting the valid volume~(\cref{sec:hardware}) with the camera ray corresponding to the current pixel. Next, we uniformly split the range into $n_{\operatorname{bin}}$ bins ($n_{\operatorname{bin}}$ = 100 in our experiments). For each bin, it stores the highest Zero-Normalized Cross-Correlation~(ZNCC) score among all randomly sampled candidates, whose depth falls within the current bin: for each candidate, the score is computed between the physical measurements at the current pixel under the projected light/mask pattern(s) so far, and the corresponding simulated measurements~(\cref{eq:render}). Finally, the scores in each bin are normalized to generate a probability distribution. We use ZNCC as it is a common choice of metric in depth acquisition~\cite{Martin2007ExperimentalCO,kutu_2018_optimalSL}.
Please refer to~\cref{fig:prob} for an illustration.

For a BRDF parameter, its probability model is similar to that of depth, except that its range is determined by the statistics reported in OpenSVBRDF~\cite{ma2023svbrdf}. Moreover, we calculate the inverse of $\mathcal{L}_1$ distance instead of ZNCC, as $\mathcal{L}_1$ norm is commonly used in state-of-the-art appearance reconstruction~\cite{bi2024rgs,Zeng2023nrhints}.


\textbf{Sample.} 
To sample the depth/reflectance parameter of a candidate from our model, we randomly select a bin based on the scores after normalization (i.e., the probability mass function). Next, we continuously sample a value in the range of the current bin uniformly at random, as the final result (\cref{fig:prob}).


\textbf{Update.} All histograms are initialized as a uniform probability distribution. We adopt a Monte-Carlo approach to update them. First, we randomly sample the depth/BRDF parameters for $n_{\operatorname{sample}}$ candidates ($n_{\operatorname{sample}}$ = 600 in our experiments), according to the current probability models. For each candidate, we compute its ZNCC score and store the highest one among all candidates in the corresponding bin in the probability model for depth. Similar operations are carried out by updating with the inverse of $\mathcal{L}_1$ distance of the current candidate, for the probability models for BRDF parameters.
We hope that as more measurements under adaptive illumination are received, the probability distribution on either depth or each reflectance parameter at a valid pixel becomes more concentrated (i.e., we become more certain about the depth and appearance). The bottom row of~\cref{fig:pattern_visualize} shows an example of the histogram of depth at a single pixel that gradually concentrates around the ground truth.

\subsection{Optimizing the Next Pattern(s)}
\label{sec:pattern}

\textbf{Loss function.} It is defined as the sum of the depth uncertainty at each valid pixel. For a single pixel, we compute its depth uncertainty as the cross entropy:
\begin{equation}
-\sum_{a,b} y_{a,b} \log(\hat{y}_{a,b}),
\label{eq:cross_entropy}
\end{equation}
where we cast the problem as standard multi-class classification~\cite{xxm_2023_unified}. Specifically, we randomly sample candidates according to the current probability distributions (\cref{sec:model}), and treat each candidate as a class of its own. $y_{a,b}$ is the ideal likelihood for classifying candidate $a$ to class $b$. It is 1 if $a=b$, and 0 otherwise. $\hat{y}_{a,b}$ is the computed likelihood for classifying candidate $a$ to class $b$ from the simulated measurements under existing and the next pattern(s), as:
\begin{equation}
\hat{y}_{a,b} = \frac{e^{\operatorname{ZNCC}(\{I_{j,a}\}_j,\{I_{j,b}\}_j)}}{\sum_{b} e^{\operatorname{ZNCC}(\{I_{j,a}\}_j,\{I_{j,b}\}_j)}}.
\end{equation}
Here $\{I_{j,a}\}_j$/$\{I_{j,b}\}_j$ is a vector of measurements, each of whose elements is calculated with~\cref{eq:render}, corresponding to a particular light/mask pattern.

Intuitively, our loss function in~\cref{eq:cross_entropy} encourages the distinctiveness from one candidate to another in terms of the corresponding measurements under the patterns being optimized. Note that the loss is defined with respect to depth uncertainty only, without explicit consideration of the uncertainty in reflectance. The reason is that per-pixel reflectance can be faithfully recovered by using even a small number of photographs under varying illumination, as in state-of-the-art techniques~\cite{xxm_2023_unified,ma2023svbrdf}. So we would like to devote our effort to the more difficult problem of depth reconstruction, as reflectance can be easily recovered as a by-product: our experiments confirm that photographs taken with adaptive illumination that minimizes depth uncertainty are sufficient to produce high-quality reflectance results (see \cref{sec:results} and additional results in supplementary material).


\textbf{Optimization.} According to the relationship in~\cref{eq:render}, we can differentiably optimize the next light and mask pattern(s) adaptive to the current situation, by minimizing the loss function in \cref{eq:cross_entropy}. Please refer to~\cref{fig:pipeline} for a graphical illustration.

For physical plausibility and ease of calibration, the value of each light/mask pattern pixel goes through a sigmoid function to fit in the range of [0,1]. In particular, to generate a pixel in a mask pattern, we multiply a free variable with a large scalar ($10^8$ in our experiments) before going to the sigmoid, to encourage the final result to be either 0 or 1. Note that in our experiments, we optimize $n_{\operatorname{batch}}$ light/mask patterns at a time, to amortize the training cost ($n_{\operatorname{batch}}$ = 3 in our experiments). Please refer to~\cref{fig:pattern_visualize} for a visualization.





We find that each candidate is not equally important in pattern optimization. For candidates whose ZNCC score is considerably smaller than the current best, it is unlikely that any of them will be the final solution. Moreover, for candidates whose depth is very close to the one with the highest ZNCC score, it is not feasible to obtain disambiguating pattern(s), as it is beyond the resolution of our LCD panel. Therefore, in practice, we find it effective to use only the candidates with top $n_{\operatorname{peak}}$ peak ZNCC scores in computing the cross entropy ($n_{\operatorname{peak}}$ = 3 in our experiments). Local maximum filtering with adaptive thresholding is applied for peak detection.

\input{figs/patterns}

\input{figs/texmap}

\section{Fine-Tuning}
\label{sec:reconstruction}

This is a two-step process. The first step is to initialize depth and reflectance for each valid pixel, from the corresponding probability models after differentiable adaptive acquisition. Specifically, for each histogram, we subdivide each of its bins into 5 smaller ones, pick the bin with the highest score, and uniformly at random sample a value within its range as the initial value for the corresponding parameter (i.e., depth or a GGX parameter). 


The second step is simultaneous fine-tuning of both depth and reflectance parameters, by minimizing the differences between physical and synthetic measurements, which are simulated according to~\cref{eq:render}. As directly optimizing native GGX parameters is difficult~\cite{xxm_2023_unified}, we reparameterize the BRDF model with a 16D neural latent vector, along with 5 MLPs. Each MLP converts the latent to respective GGX parameters,  to facilitate differentiable optimization. Before fine-tuning, for each valid pixel, we precompute a latent that closely predicts the initial values of GGX parameters from the first step. Please refer to~\cite{xxm_2023_unified} for details.

\section{Implementation Details}
\label{sec:details}

We apply~\cite{kirillov2023segany} to perform foreground segmentation to determine the pixels of interest. For efficient computation during adaptive acquisition, we downsample each photograph from its original resolution to 127$\times$64. We find that this resolution is suitable for our prototype, achieving decent quality without incurring an excessive computational burden~(\cref{sec:ablate}).
A higher resolution can be used when more computational power is available. After acquisition, we start with the low resolution of 127$\times$64, and gradually upsample it to the original resolution during fine-tuning~(\cref{sec:reconstruction}).
Our differentiable pipeline is implemented with PyTorch. Adam optimizer is used in all experiments, with a learning rate of \(10^{-3}\) and a weight decay of \(10^{-6}\).



\input{figs/geo_main}

\section{Results and Discussions}
\label{sec:results}

We capture the shape and reflectance of 10 physical objects from a single view. The maximum dimension of each object ranges from 9 to 15cm. The appearance ranges from diffuse-dominant clay, wood, plastic, metallic paint, to highly specular natural wax coating. An exposure time of 0.2s is set for taking one photograph, and only LDR input images are used in our pipeline. All computation is conducted on a workstation with dual Intel Xeon 4210 CPUs, 256GB DDR4 memory and a GeForce RTX 3090 GPU. It takes about 10 minutes for adaptive acquisition (the majority of which is spent on pattern optimization, with only 15 seconds of total  exposure time) and 2 hours for joint fine-tuning of about 1024$\times$1024 depth map and multiple textures map for GGX parameters. Note that we focus on the former, as the latter is relatively easier to scale up via means like parallelization.

\subsection{Comparisons}
\label{sec:comp}

\textbf{Geometry.} We compare our shape reconstruction with related work in~\cref{fig:geo_main}. According to~\cite{sundar2022SinglePhotonSL}, we measure geometric quality in the following metrics: RMSE in estimated depth, percentage of inliers (absolute depth error $\textless$ 3mm), and RMSE among the inliers. The ground-truth shape is captured with a commercial 3D scanner~\cite{shining3d}.

In the 1\textsuperscript{st},3\textsuperscript{rd} and 4\textsuperscript{th} column of the figure, we compare with~\cite{xxm_2023_unified}, the closest state-of-the-art technique to ours, and~\cite{gupta_2012_mpsSL}, a representative method on traditional structured illumination. Our approach employs 3$\times$24 = 72 adaptive light/mask patterns. For~\cite{xxm_2023_unified}, the same number of patterns are pre-trained for a fair comparison. Note that each of their lighting conditions requires 20s of exposure time, due to the limited power of a single LED, which is two orders of magnitude longer than our approach with multiple LEDs on simultaneously. For~\cite{gupta_2012_mpsSL}, based on the effective spatial resolution of our LCD panel, a frequency band consisting of 16 pixels and 15 frequencies are used to generate 34 patterns. Our geometric results quantitatively outperform the two related techniques. For {\sc{Orange}}, we are unable to scan a ground-truth shape as it is non-rigid. Yet one can qualitatively inspect the results: ours exhibit less banding artifacts.


In~\cref{fig:cmp_shadow}, we further demonstrate the benefit of making full use of the LED array during acquisition, compared with~\cite{xxm_2023_unified} using a single light source, as is common in the majority of work on structured illumination. Our result is more complete, as we are able to cast light from different sources to reduce the shadow region, where no reliable depth estimation can be made.

\textbf{Reflectance.}~\cref{fig:texturemap} shows texture maps that store the GGX BRDF parameters. Next, in the last 3 columns of~\cref{fig:geo_main}, we compare our relighting results with~\cite{xxm_2023_unified}, while validating against corresponding photographs taken under novel lighting conditions. In all cases, our results are comparable to~\cite{xxm_2023_unified} and closely resemble the photographs. Please refer to the supplementary material for more details.



\input{figs/cmp_shadow}


 
\subsection{Ablations}
\label{sec:ablate}
In first two columns of~\cref{fig:geo_main}, we demonstrate the effectiveness of adaptive acquisition, by comparing to a variant that pre-trains 72 fixed light/mask patterns from synthetic training data, sampled
according to the statistics in~\cite{ma2023svbrdf}. In~\cref{fig:abl_pattern_num}, the depth accuracy improves with the increasing number of patterns, as more information is acquired to disambiguate current uncertainty. 



\input{figs/abl_pattern_num}

We ablate various parameters of our approach as follows. In~\cref{fig:abl_sample_num}, higher quality results are obtained with a larger number of Monte Carlo samples, at the expense of higher computational burden.~\cref{fig:abl_balance} evaluates the impact of the batch size of simutaneously optimized next patterns in the same amount of total acquisition time. Our current choice of $n_{\operatorname{batch}}$ is empirically determined from this experiment. ~\cref{fig:abl_best_peak} ablates $n_{\operatorname{peak}}$, which needs to strike a balance between missing good solutions and the speed to find the optimal value.  In~\cref{fig:abl_bin_num}, finer bins improve the quality with more computations.


Finally, we validate the effectiveness of using a low image resolution for computing adaptive patterns in~\cref{fig:abl_mipmap}. The current choice is made after balancing accuracy and performance.

\input{figs/abl_sample_num}

\input{figs/abl_balance}

\input{figs/abl_best_peak}
 
\input{figs/abl_bin_num}

\input{figs/abl_mipmap}




\section{Limitations \& Future Work}


Our work is subject to a number of limitations, which could inspire exciting future research possibilities. First, our depth uncertainty does not consider indirect illumination during adaptive pattern optimization. Second, the expressive power of depth maps and parametric BRDFs are limited. It will be intriguing to combine our approach with more advanced representations, such as Gaussian Splatting representation that supports high-quality relighting~\cite{bi2024rgs}. Finally, it seems promising to apply our idea to free-form scanning with a hand-held device~\cite{Ma_2021_FreeScan}, equipped with a miniature version of the spatial-angular structured light.

%% file: figs/prototype.tex
\begin{figure}[t] 
    \centering
    \begin{minipage}{\linewidth}
        \centering
        \begin{minipage}{.31\linewidth}
            \centering
            \includegraphics[width=\linewidth]{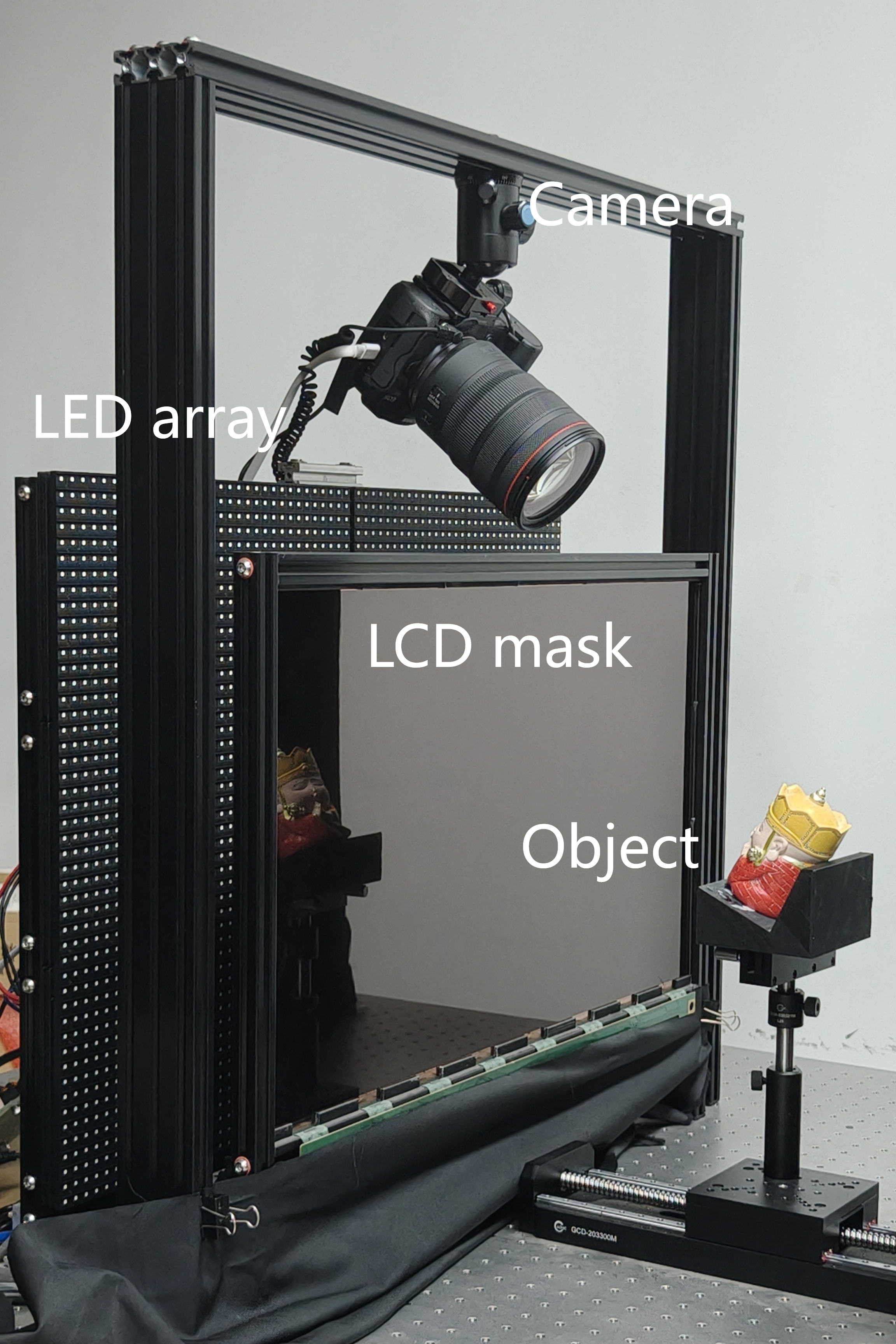}
            \put(-13,4){\scalebox{.9}{\color{white} (a)}}
        \end{minipage}
        \begin{minipage}{.31\linewidth}
            \centering
            \includegraphics[width=\linewidth]{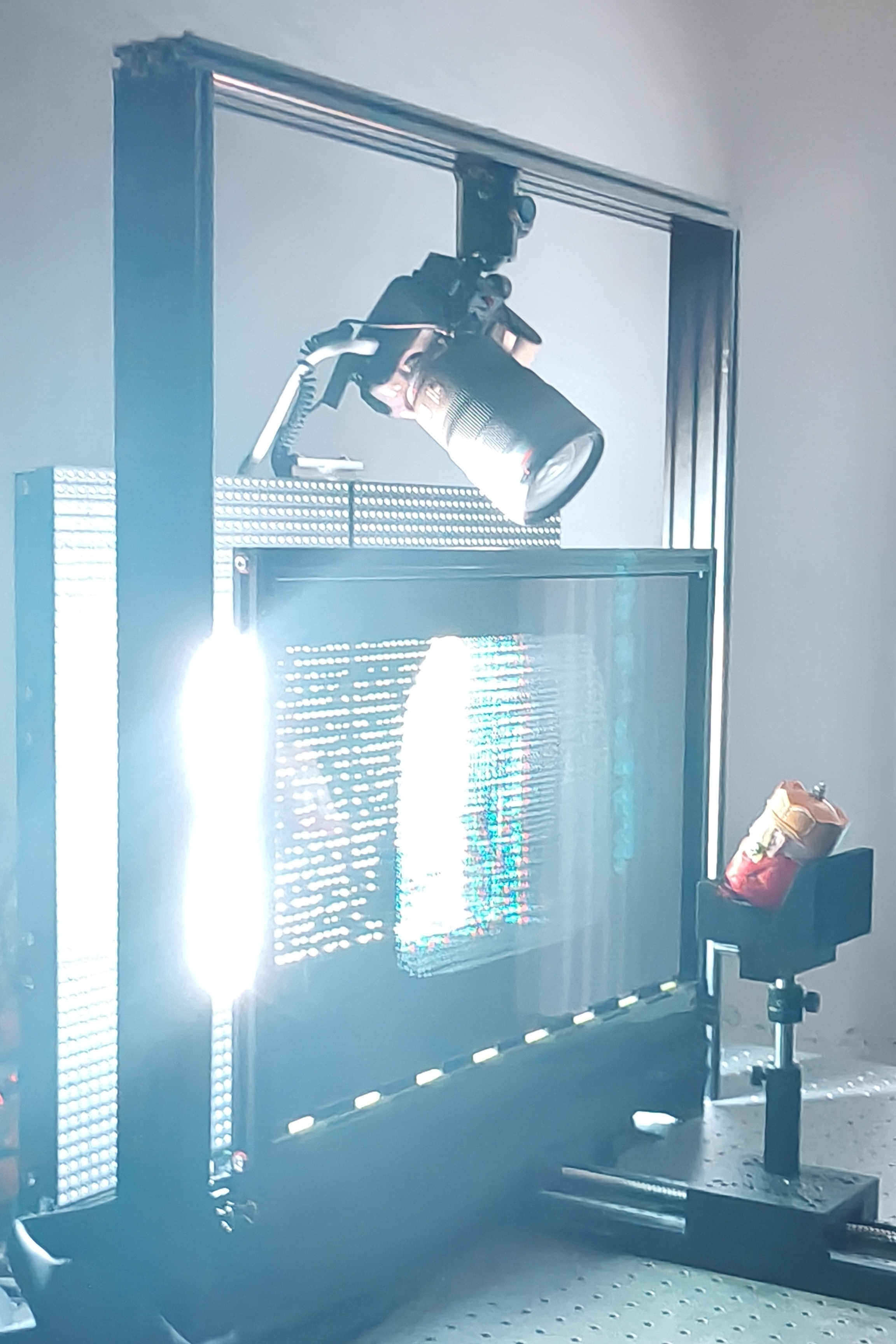}
            \put(-13,4){\scalebox{.9}{\color{white} (b)}}
        \end{minipage}
        \begin{minipage}{.31\linewidth}
            \centering
            \includegraphics[width=\linewidth]{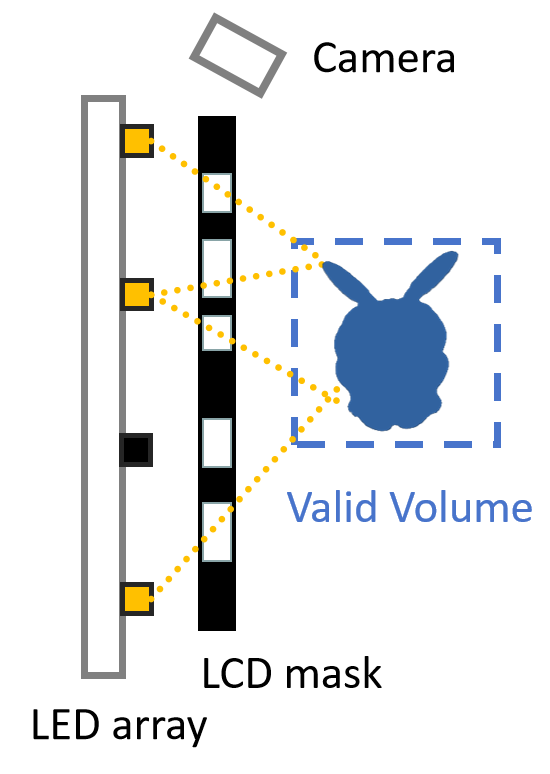}
            \put(-13,1){\scalebox{.9}{\color{black} (c)}}
        \end{minipage}
    \end{minipage}
    
    \caption{Our acquisition setup. It consists of a camera, an LED array and an LCD mask (a). The setup is working with optimized light/mask pattern (b). A side view is illustrated in (c).}
    \label{fig:prototype}
\end{figure}

%% file: figs/pipeline.tex
\begin{figure*}
    \centering
\includegraphics[width=\textwidth]{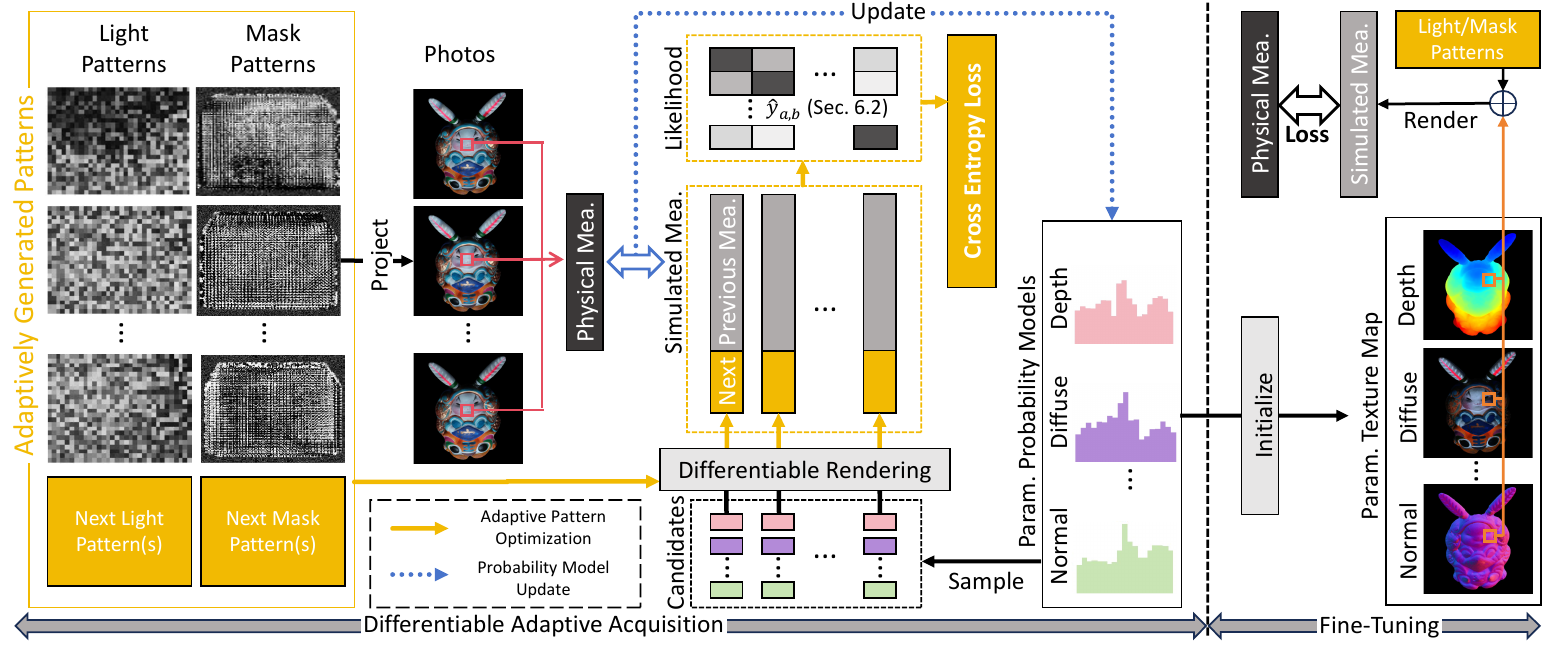}
    \caption{Our pipeline consists of two stages. First, for a physical object, we compute the next light/mask pattern(s) by minimizing the cross entropy among possible candidates sampled from histogram-based probability models. We then take photograph(s) with these patterns, and update probability distributions based on new measurements. This process is repeated until a termination condition is met. Next, we use the depth/reflectance estimate from previous stage as initial values, and fine-tune the results by minimizing the differences between physical measurements and corresponding simulated ones. The final output is a depth map and several texture maps that store parameters of the GGX BRDF model. Params. = parameters, mea. = measurements.
    ~
    }    
    \label{fig:pipeline}
\end{figure*}

%% file: figs/patterns.tex
\begin{figure}[htbp]
    \centering
    \captionsetup[subfigure]{justification=centering}
    \begin{minipage}{\linewidth}
        \centering
        \begin{minipage}{\textwidth}
            \centering
            \begin{minipage}{.24\textwidth}
                \centering 
                \subcaption*{\small Initial}
            \end{minipage}
            \begin{minipage}{.24\textwidth}
                \centering
                \subcaption*{\small Pattern\#3}
            \end{minipage}
            \begin{minipage}{.24\textwidth}
                \centering
                \subcaption*{\small Pattern\#12}
            \end{minipage}
            \begin{minipage}{.24\textwidth}
                \centering
                \subcaption*{\small Pattern\#30}
            \end{minipage}
        \end{minipage}
    \end{minipage}  
    \begin{minipage}{\linewidth}
        \centering
        \begin{minipage}{.24\linewidth}
            \centering
            \includegraphics[width=\linewidth]{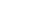}
        \end{minipage}
        \begin{minipage}{.24\linewidth}
            \centering
            \includegraphics[width=\linewidth]{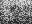}
        \end{minipage}
        \begin{minipage}{.24\linewidth}
            \centering
            \includegraphics[width=\linewidth]{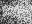}
        \end{minipage}
         \begin{minipage}{.24\linewidth}
            \centering
            \includegraphics[width=\linewidth]{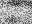}
        \end{minipage}
    \end{minipage}
    \begin{minipage}{\linewidth}
        \centering
        \begin{minipage}{.24\linewidth}
            \centering
            \includegraphics[width=\linewidth]{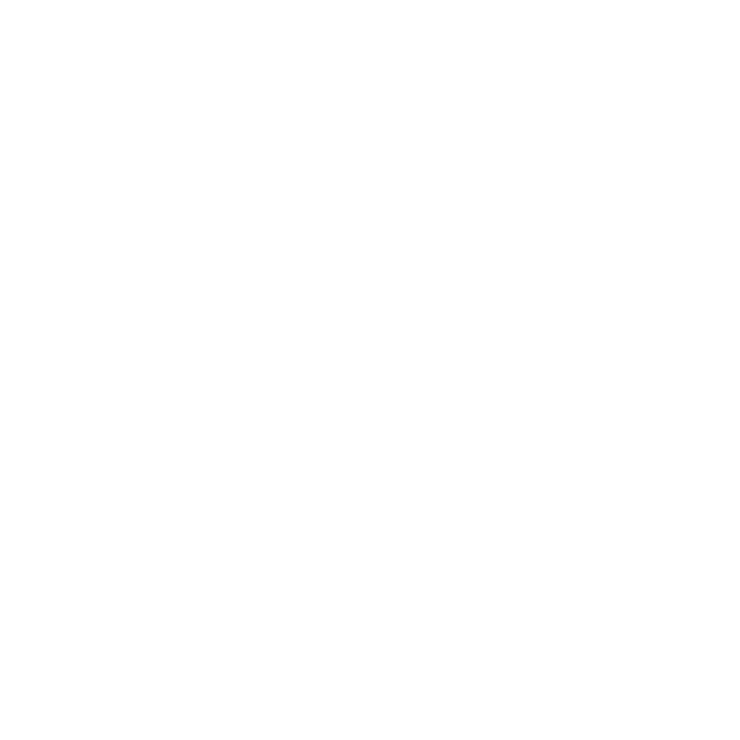}
        \end{minipage}
        \begin{minipage}{.24\linewidth}
            \centering
            \includegraphics[width=\linewidth]{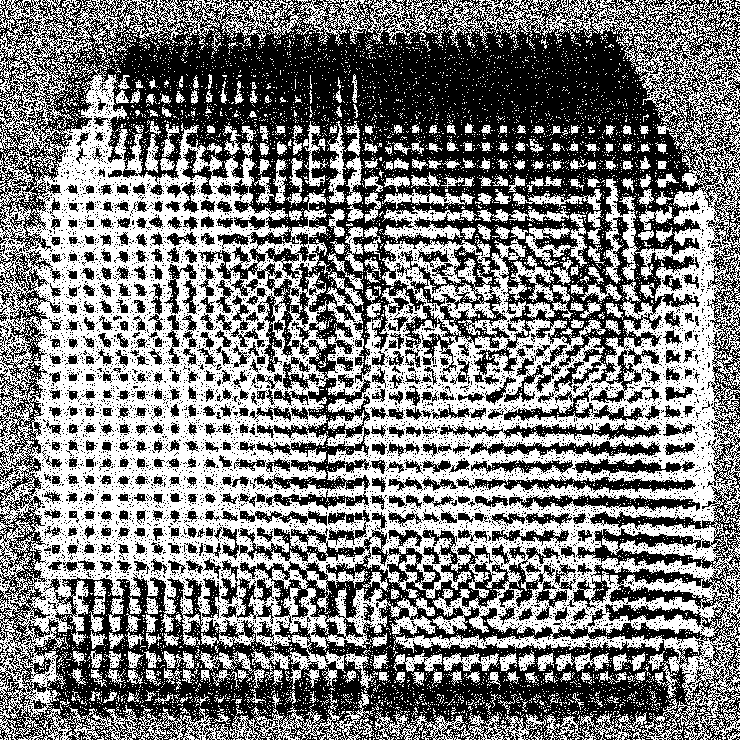}
        \end{minipage}
        \begin{minipage}{.24\linewidth}
            \centering
            \includegraphics[width=\linewidth]{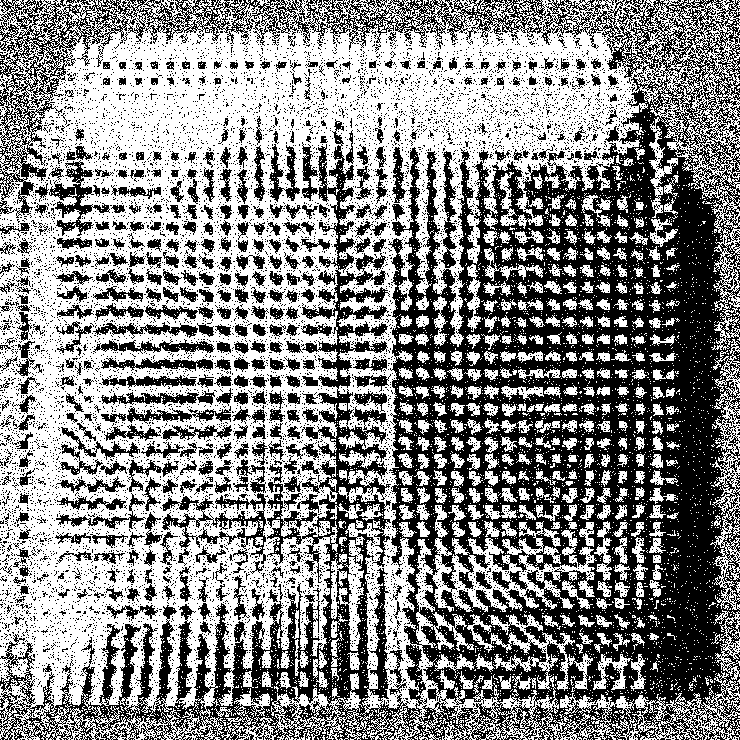}
        \end{minipage}
        \begin{minipage}{.24\linewidth}
            \centering
            \includegraphics[width=\linewidth]{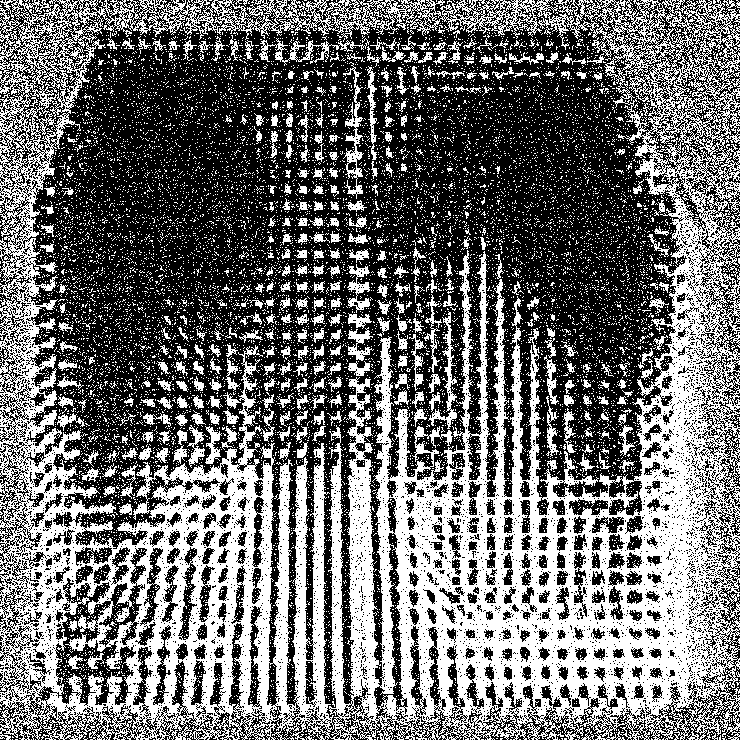}
        \end{minipage}
    \end{minipage}
    \begin{minipage}{\linewidth}
        \centering
        \begin{minipage}{.24\linewidth}
            \centering
            \includegraphics[width=\linewidth]{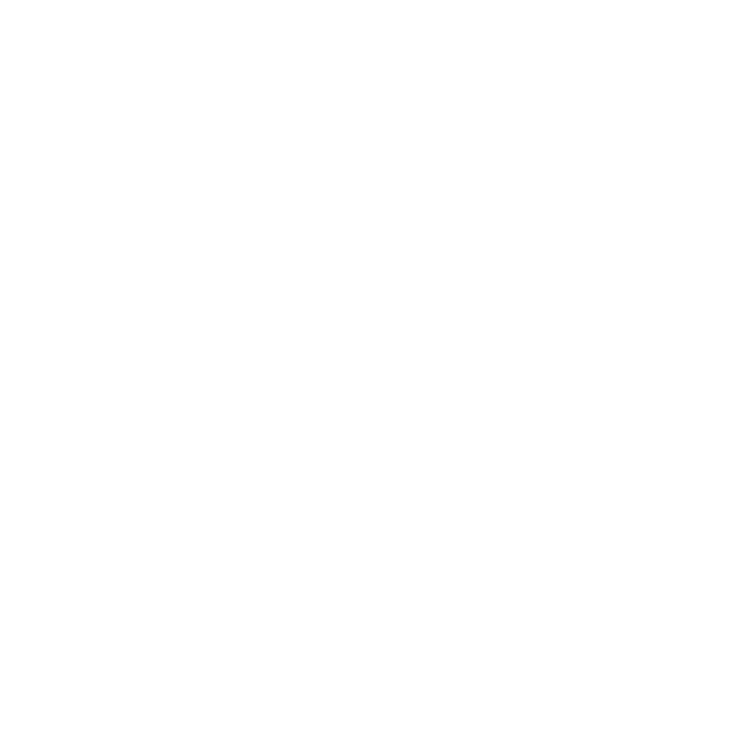}
        \end{minipage}
        \begin{minipage}{.24\linewidth}
            \centering
            \includegraphics[width=\linewidth]{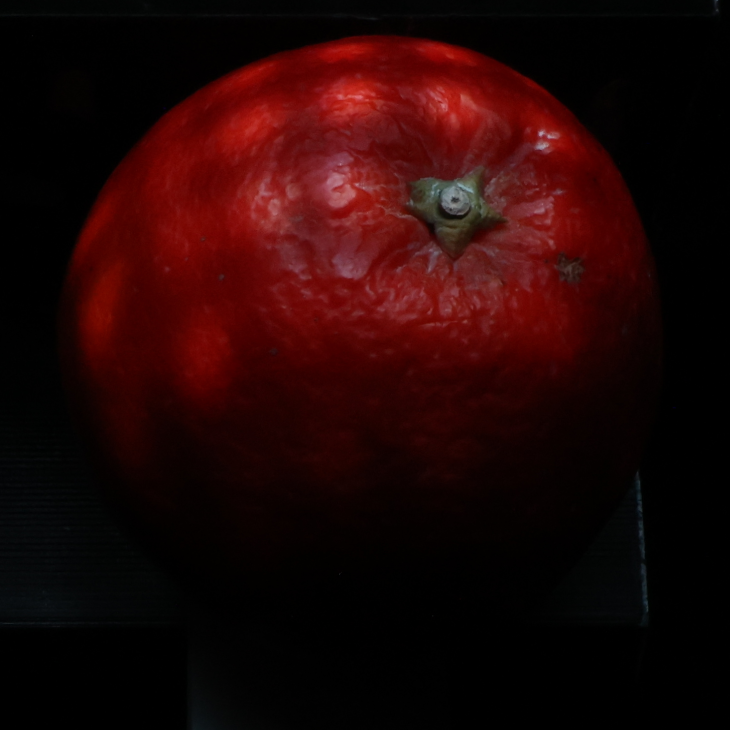}
        \end{minipage}
        \begin{minipage}{.24\linewidth}
            \centering
            \includegraphics[width=\linewidth]{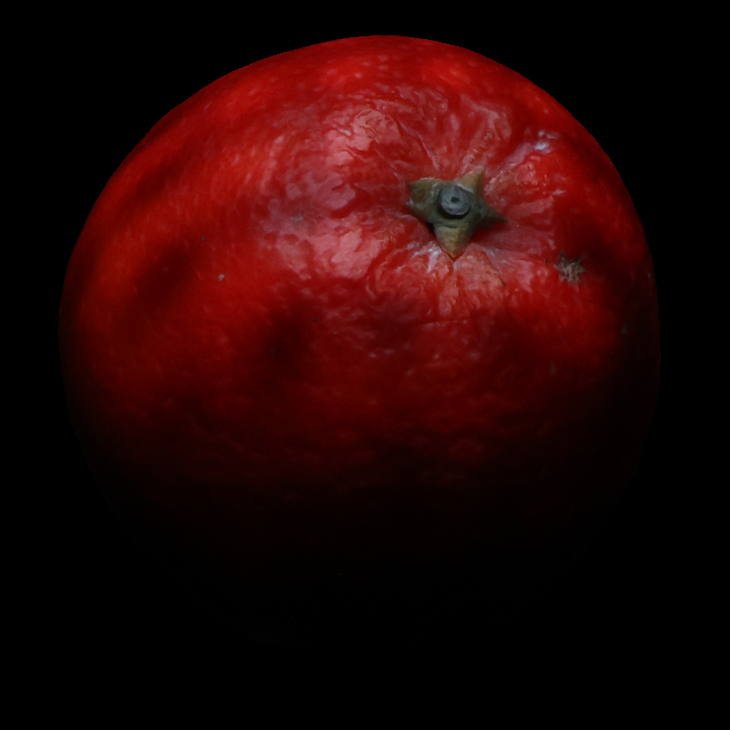}
        \end{minipage}
        \begin{minipage}{.24\linewidth}
            \centering
            \includegraphics[width=\linewidth]{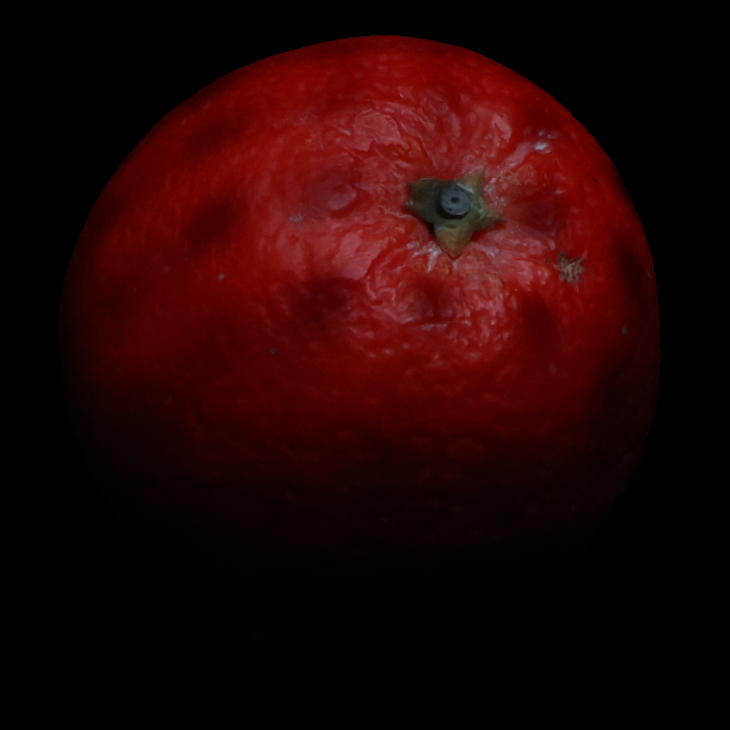}
        \end{minipage}

    \end{minipage}  
    \begin{minipage}{\linewidth}
        \centering
        \begin{minipage}{.24\linewidth}
            \centering
            \includegraphics[width=\linewidth]{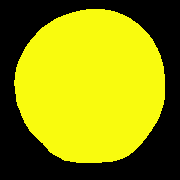}
        \end{minipage}
        \begin{minipage}{.24\linewidth}
            \centering
            \includegraphics[width=\linewidth]{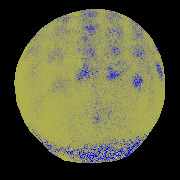}
        \end{minipage}
        \begin{minipage}{.24\linewidth}
            \centering
            \includegraphics[width=\linewidth]{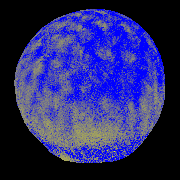}
        \end{minipage}
         \begin{minipage}{.24\linewidth}
            \centering
        \includegraphics[width=\linewidth]{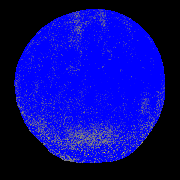}
        \end{minipage}
    \end{minipage}

    \begin{minipage}{\linewidth}
        \centering
        \begin{minipage}{.251\linewidth}
            \centering
            \includegraphics[width=\linewidth]{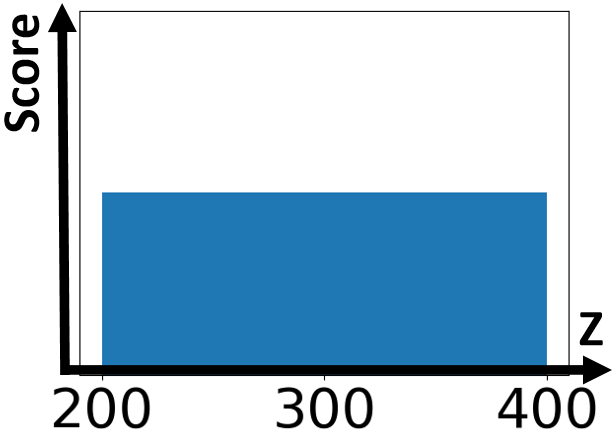}
        \end{minipage}
        \begin{minipage}{.24\linewidth}
            \centering
            \includegraphics[width=\linewidth]{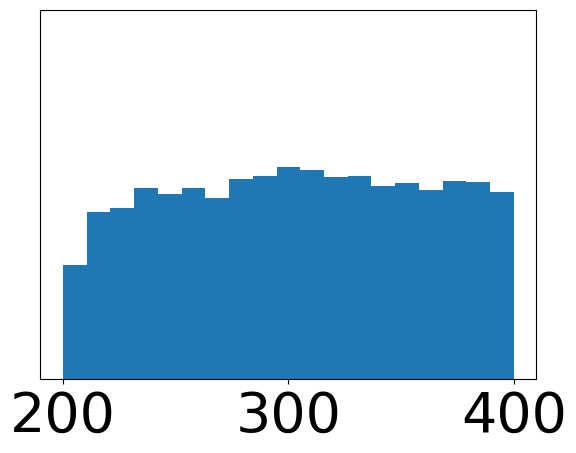}
        \end{minipage}
       \begin{minipage}{.24\linewidth}
            \centering
            \includegraphics[width=\linewidth]{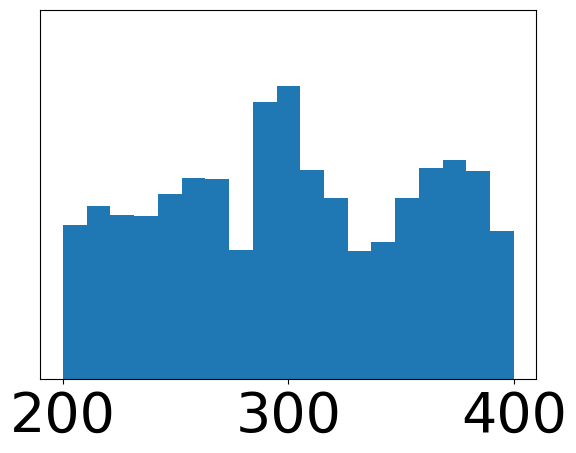}
        \end{minipage}
        \begin{minipage}{.24\linewidth}
            \centering
            \includegraphics[width=\linewidth]{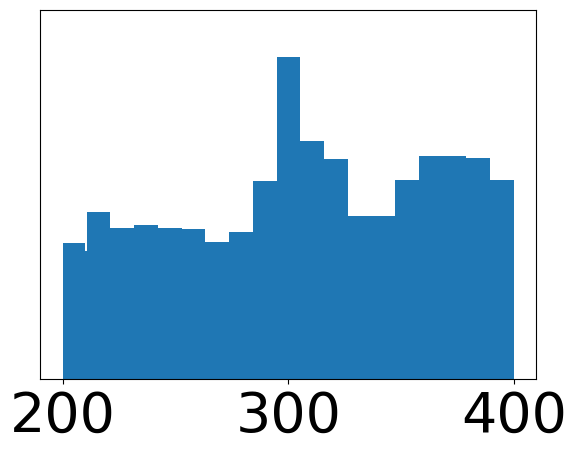}
        \end{minipage}
    \end{minipage}

    \caption{Visualization of various parts in adaptive acquisition. From the left column to right, after the initialization, after pattern\#3, \#12 and \#30 is projected. From the top row to bottom, light pattern, mask pattern, corresponding photograph, depth uncertainty visualization (yellow = uncertain, blue = certain), and the visualization of the probability model at a single pixel.}   
    \label{fig:pattern_visualize}
\end{figure}

%% file: figs/texmap.tex
\begin{figure}[htbp]
    \centering
    \captionsetup[subfigure]{justification=centering}
    \begin{minipage}{\linewidth}
        \centering
        \begin{minipage}{\textwidth}
            \centering
            \begin{minipage}{.24\textwidth}
                \centering 
                \subcaption*{\small Diff. Albedo}
            \end{minipage}
            \begin{minipage}{.24\textwidth}
                \centering
                \subcaption*{\small Spec. Albedo}
            \end{minipage}
            \begin{minipage}{.24\textwidth}
                \centering
                \subcaption*{\small Normal}
            \end{minipage}
            \begin{minipage}{.24\textwidth}
                \centering
                \subcaption*{\small Roughness}
            \end{minipage}
        \end{minipage}
    \end{minipage}  
    \begin{minipage}{\linewidth}
        \center
        \begin{minipage}{.24\linewidth}
            \centering
            \includegraphics[width=\linewidth]{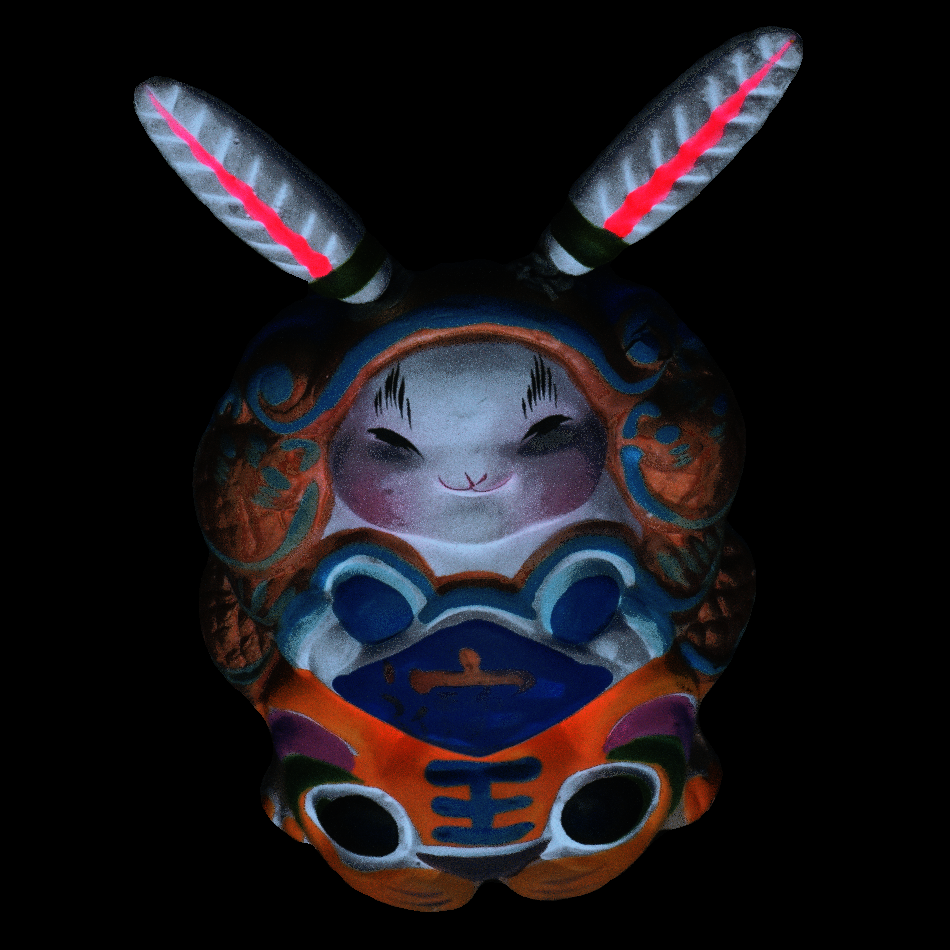}
        \end{minipage}
        \begin{minipage}{.24\linewidth}
            \centering
            \includegraphics[width=\linewidth]{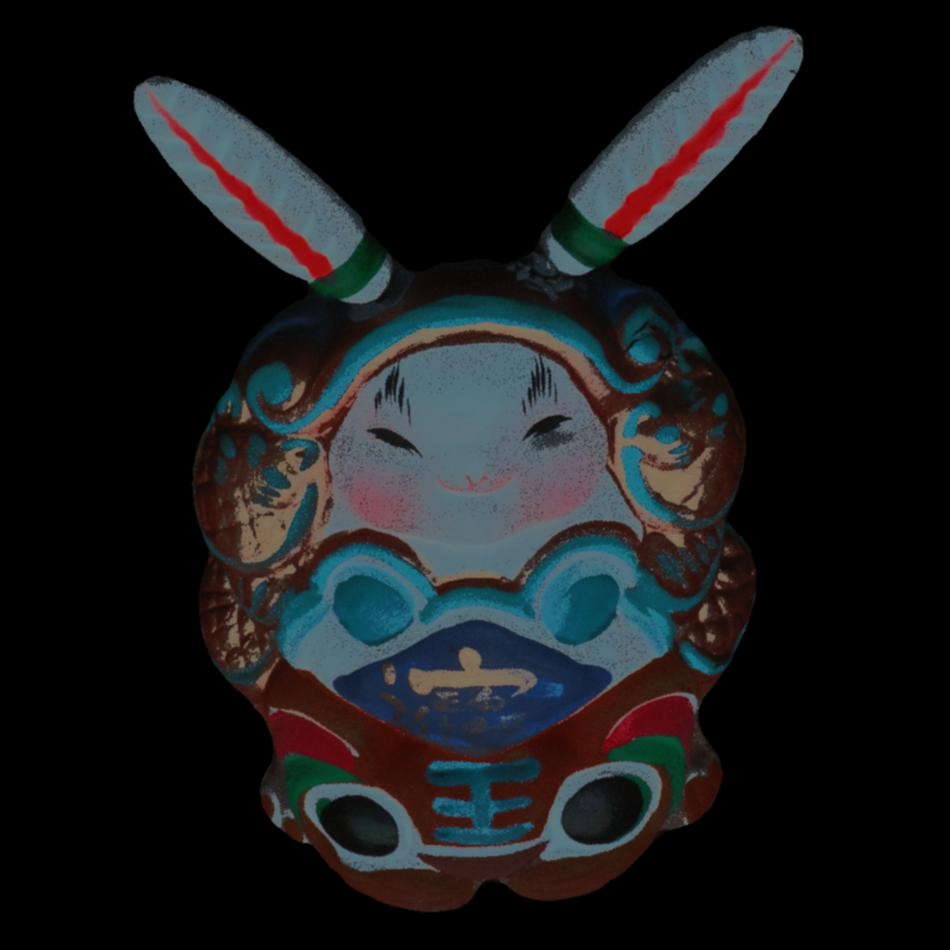}
        \end{minipage}
        \begin{minipage}{.24\linewidth}
            \centering
            \includegraphics[width=\linewidth]{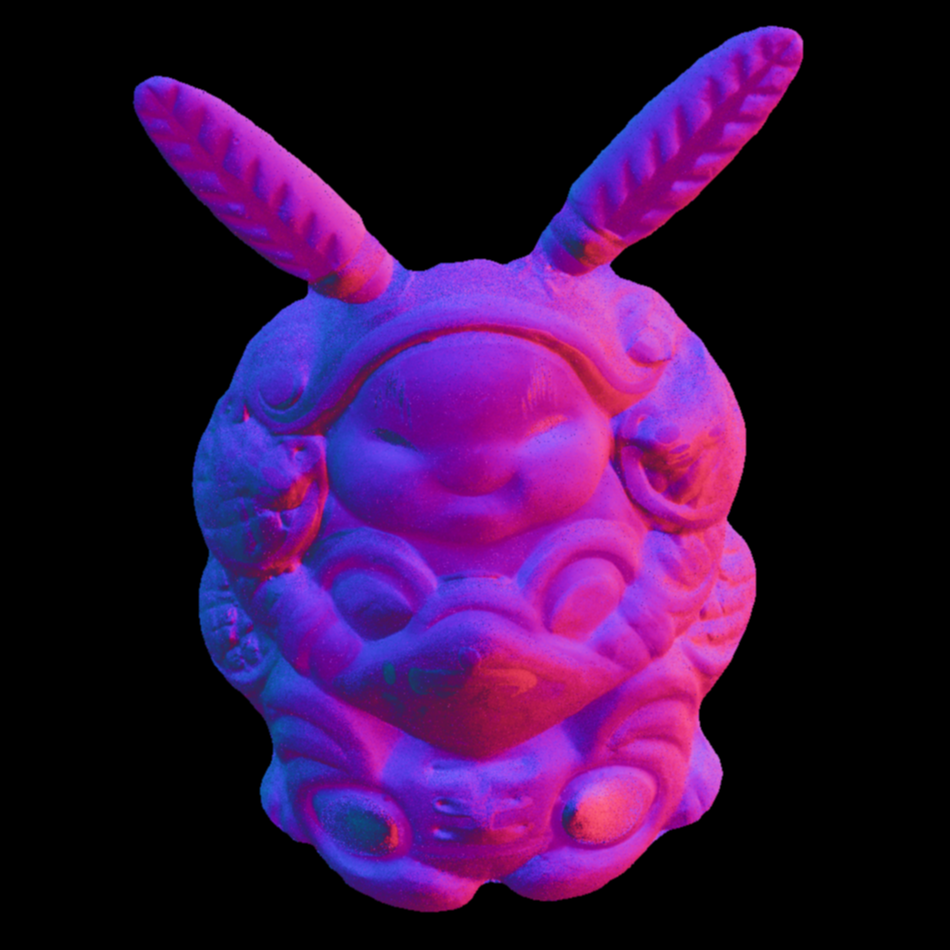}
        \end{minipage}
        \begin{minipage}{.24\linewidth}
            \centering
            \includegraphics[width=\linewidth]{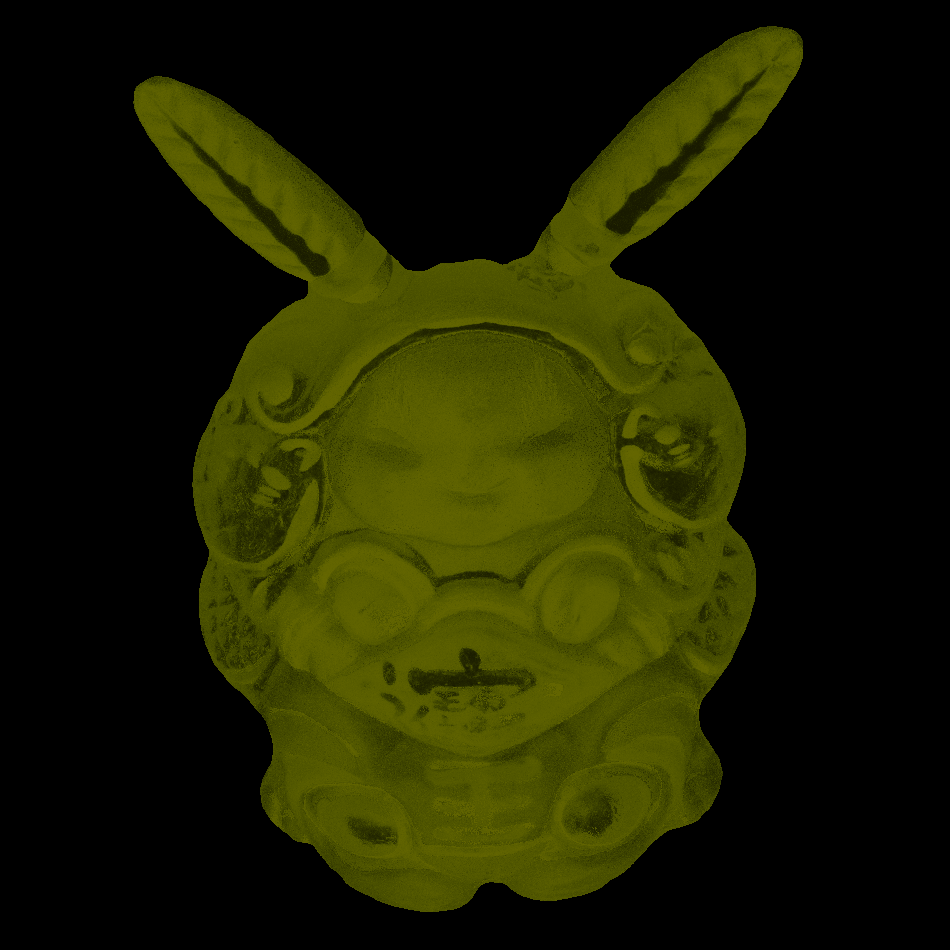}
        \end{minipage}
    \end{minipage}
   \begin{minipage}{\linewidth}
        \center
        \begin{minipage}{.24\linewidth}
            \centering
            \includegraphics[width=\linewidth]{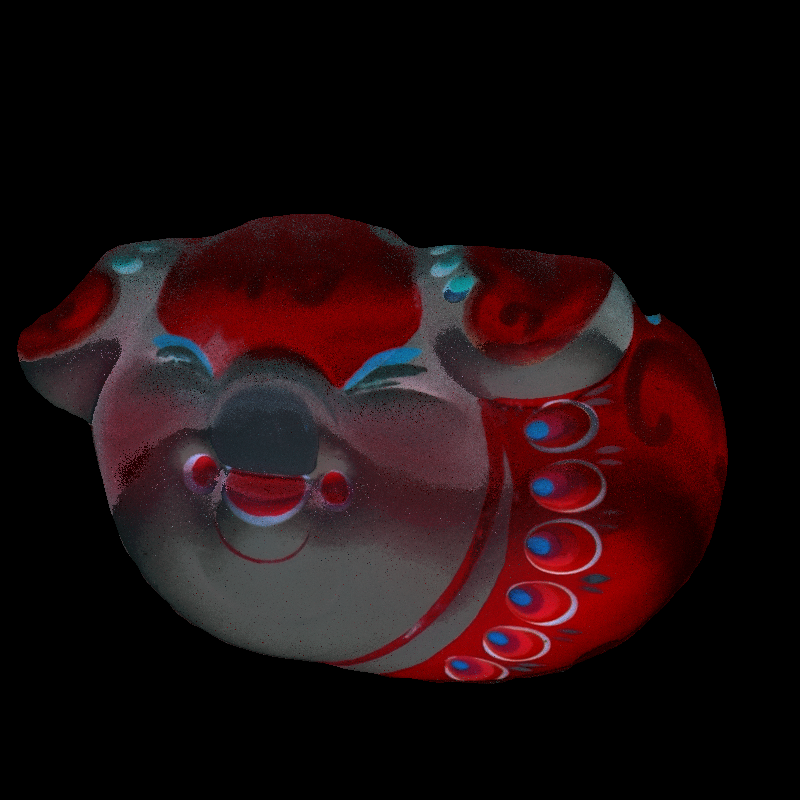}
        \end{minipage}
        \begin{minipage}{.24\linewidth}
            \centering
            \includegraphics[width=\linewidth]{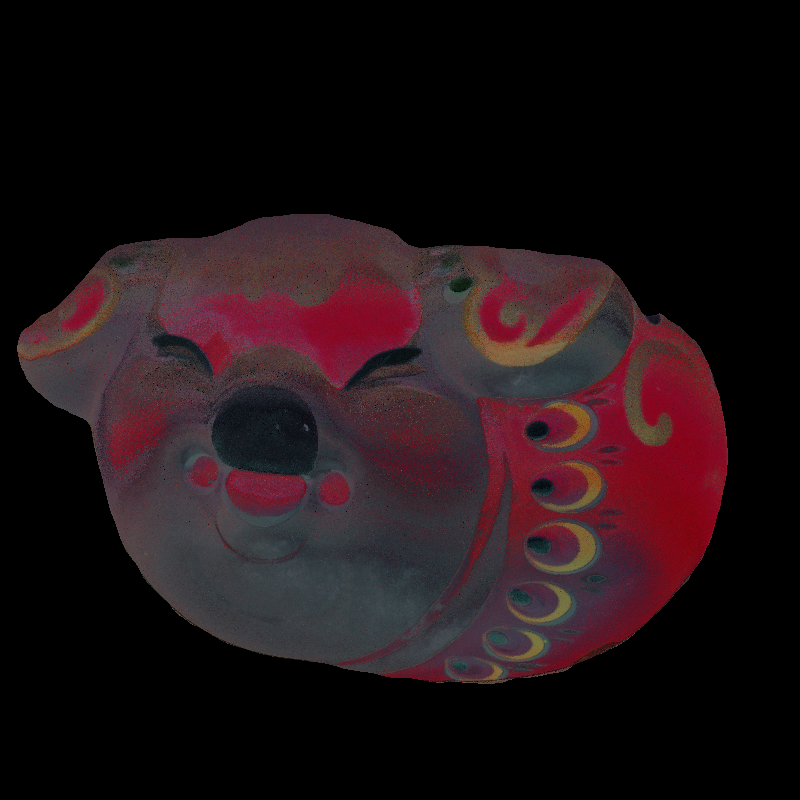}
        \end{minipage}
        \begin{minipage}{.24\linewidth}
            \centering
            \includegraphics[width=\linewidth]{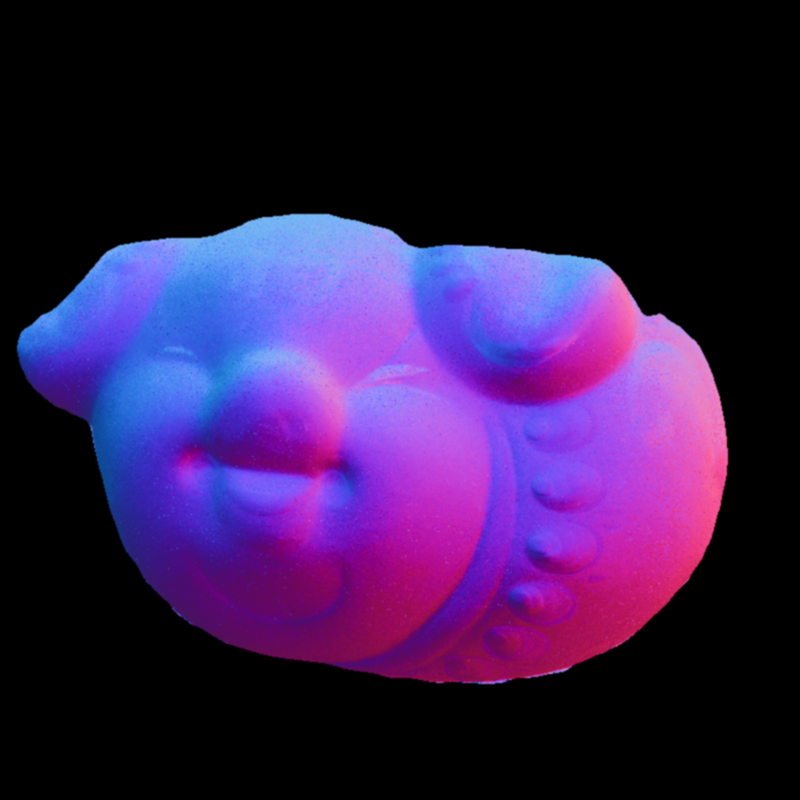}
        \end{minipage}
        \begin{minipage}{.24\linewidth}
            \centering
            \includegraphics[width=\linewidth]{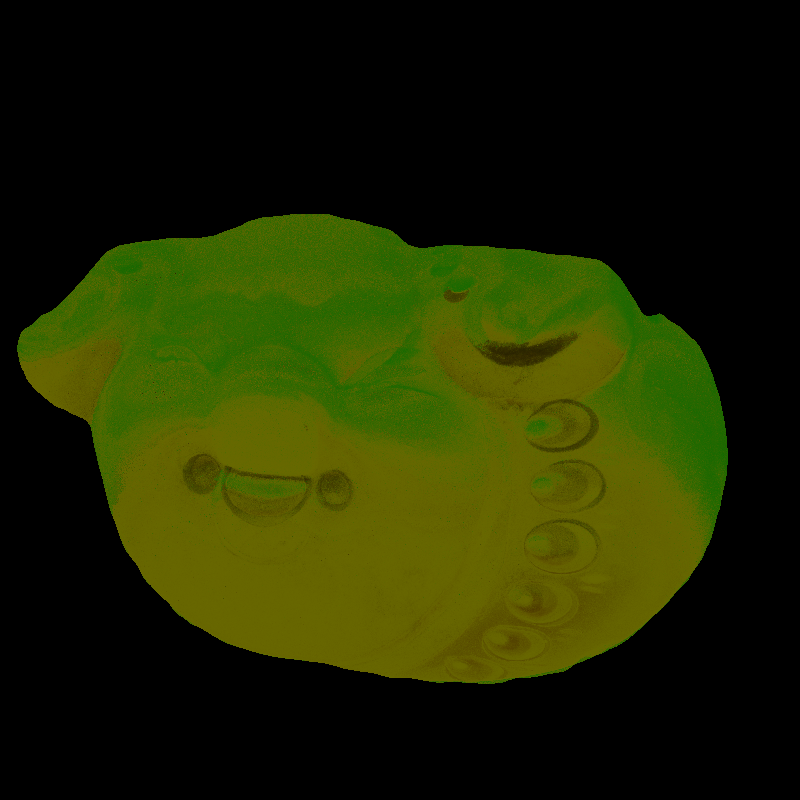}
        \end{minipage}
    \end{minipage}
       \begin{minipage}{\linewidth}
        \center
        \begin{minipage}{.24\linewidth}
            \centering
            \includegraphics[width=\linewidth]{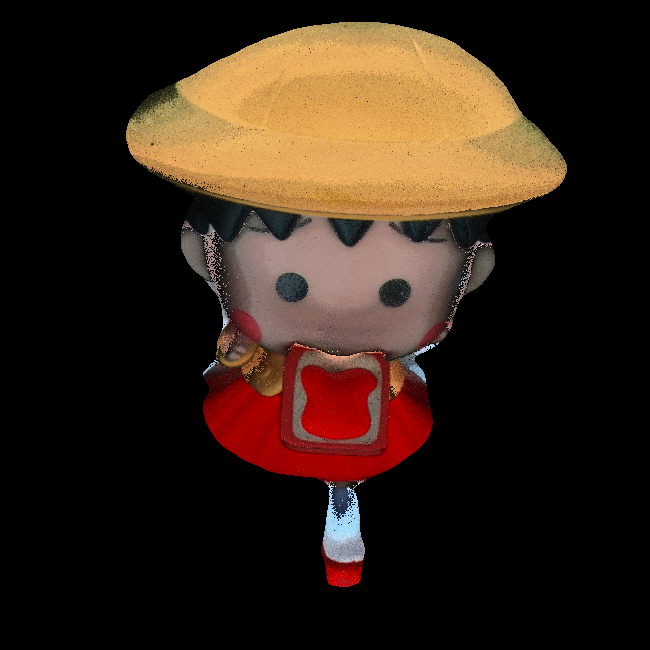}
        \end{minipage}
        \begin{minipage}{.24\linewidth}
            \centering
            \includegraphics[width=\linewidth]{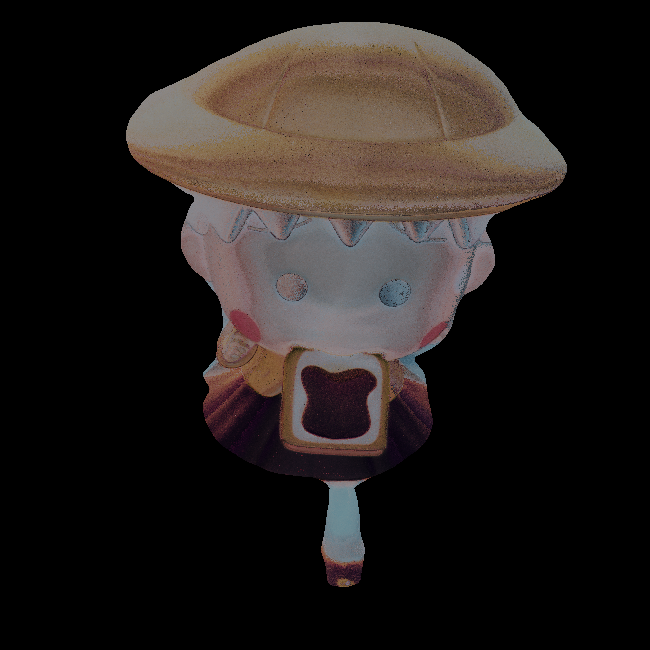}
        \end{minipage}
        \begin{minipage}{.24\linewidth}
            \centering
            \includegraphics[width=\linewidth]{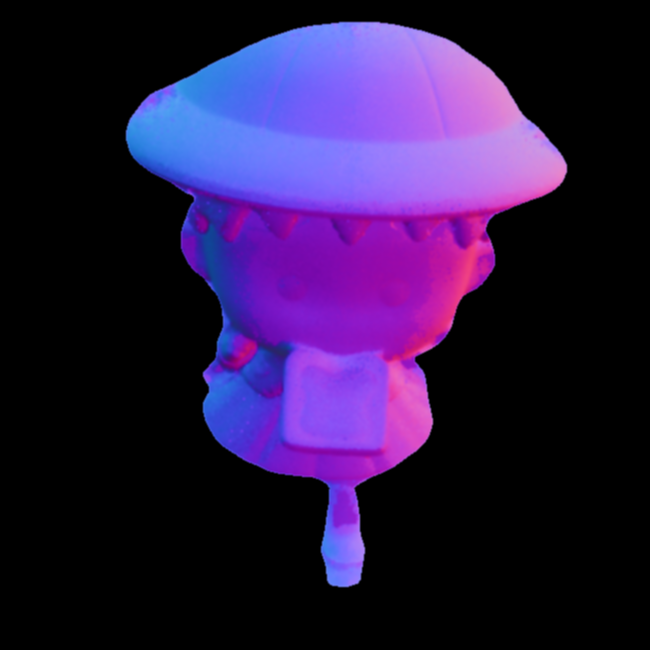}
        \end{minipage}
        \begin{minipage}{.24\linewidth}
            \centering
            \includegraphics[width=\linewidth]{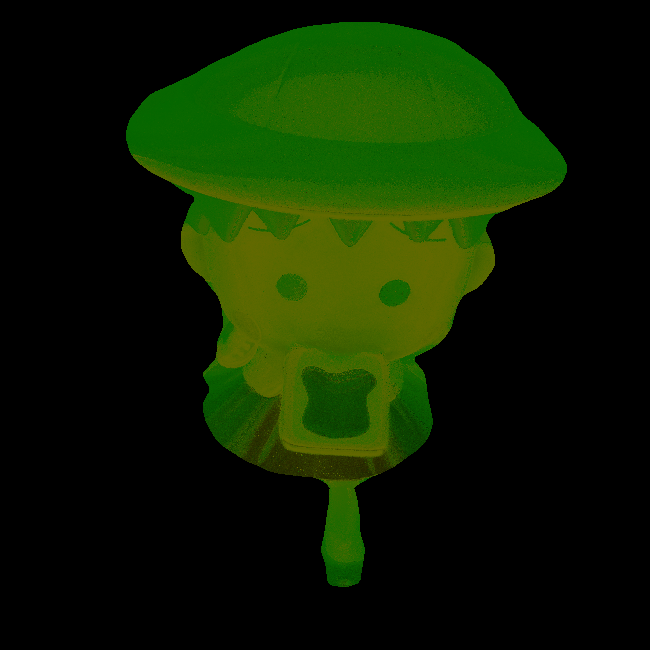}
        \end{minipage}
    \end{minipage}
       \begin{minipage}{\linewidth}
        \center
        \begin{minipage}{.24\linewidth}
            \centering
            \includegraphics[width=\linewidth]{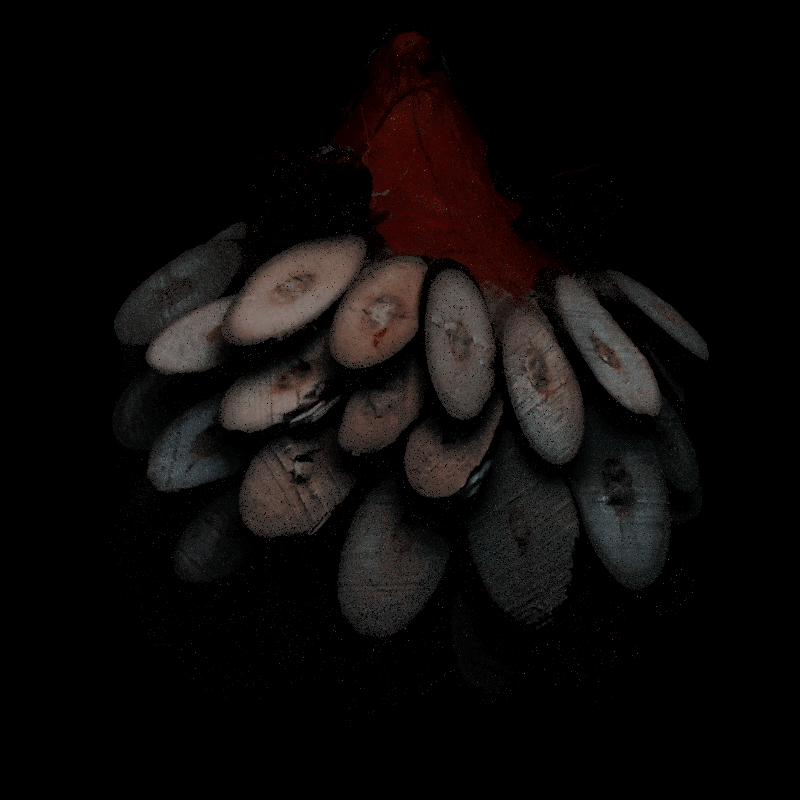}
        \end{minipage}
        \begin{minipage}{.24\linewidth}
            \centering
            \includegraphics[width=\linewidth]{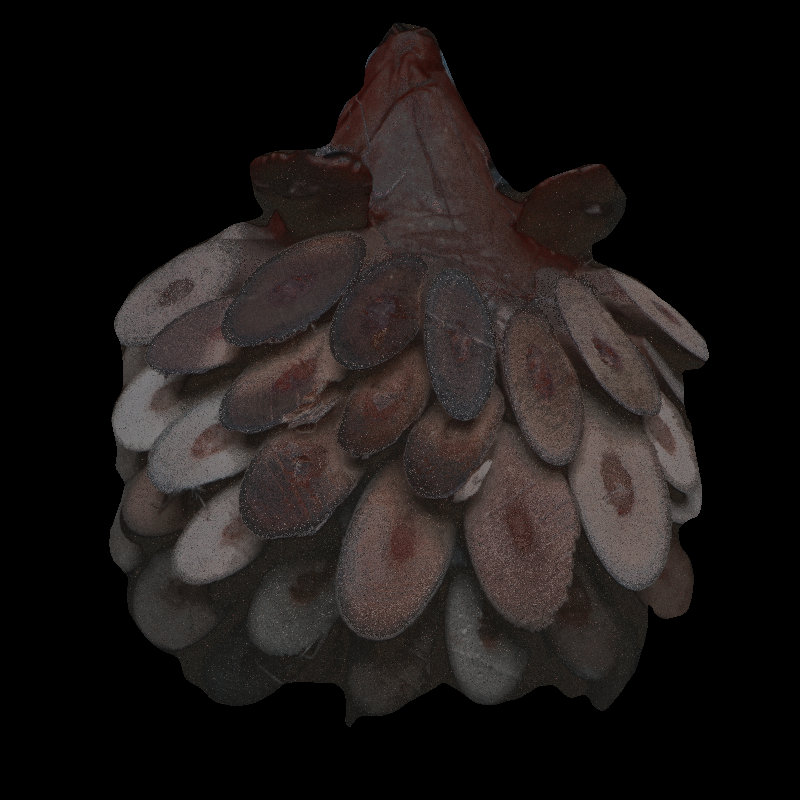}
        \end{minipage}
        \begin{minipage}{.24\linewidth}
            \centering
            \includegraphics[width=\linewidth]{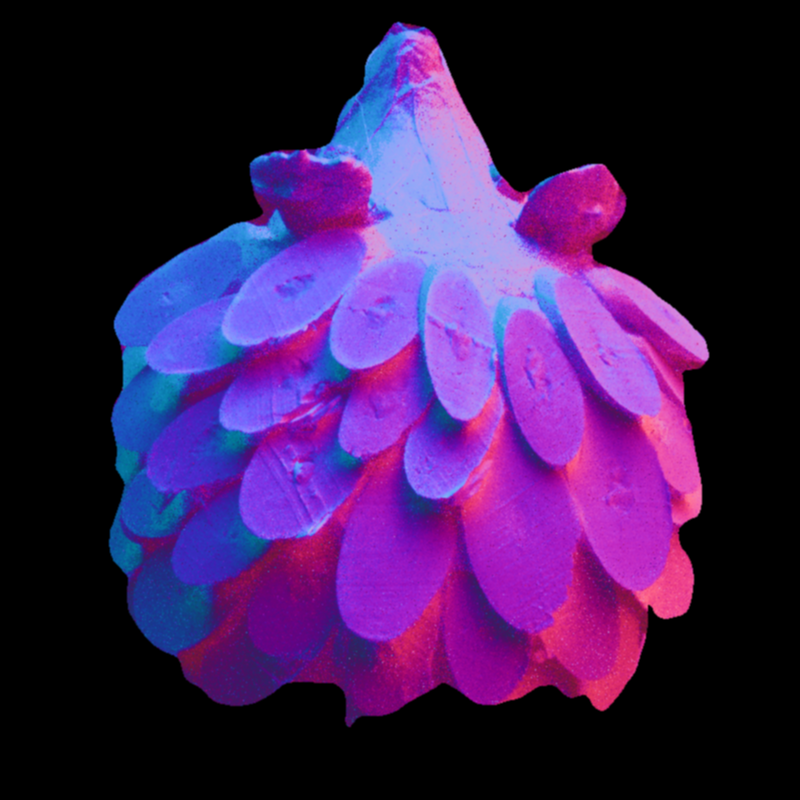}
        \end{minipage}
        \begin{minipage}{.24\linewidth}
            \centering
            \includegraphics[width=\linewidth]{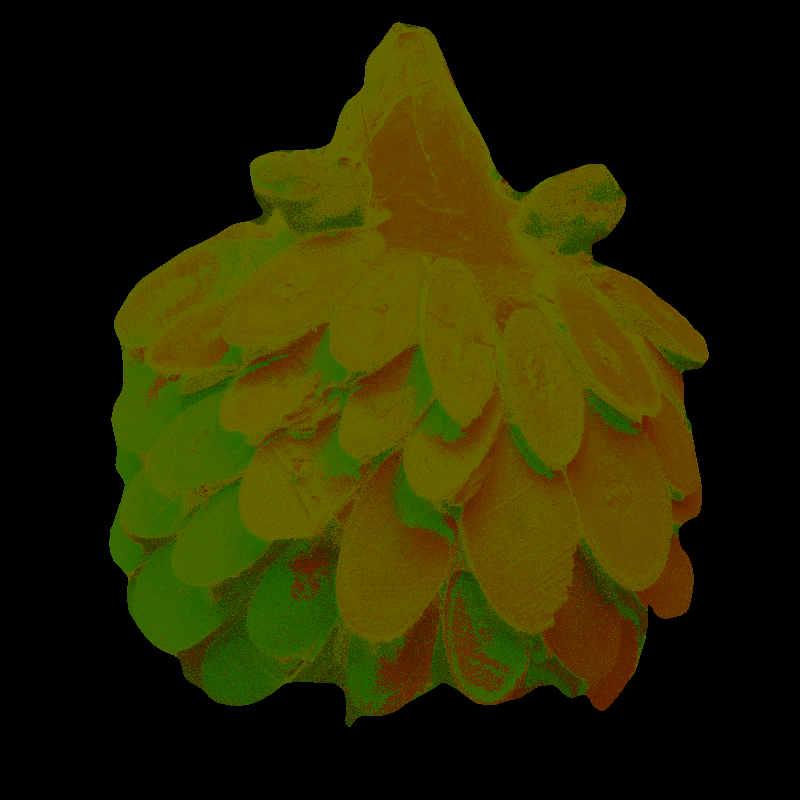}
        \end{minipage}
    \end{minipage}

    \begin{minipage}{\linewidth}
        \center
        \begin{minipage}{.24\linewidth}
            \centering
            \includegraphics[width=\linewidth]{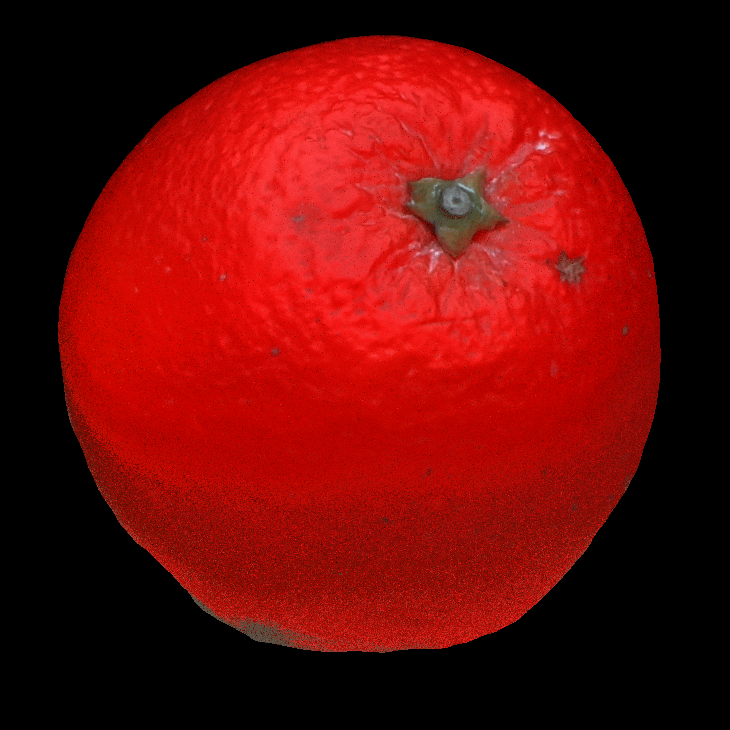}
        \end{minipage}
        \begin{minipage}{.24\linewidth}
            \centering
            \includegraphics[width=\linewidth]{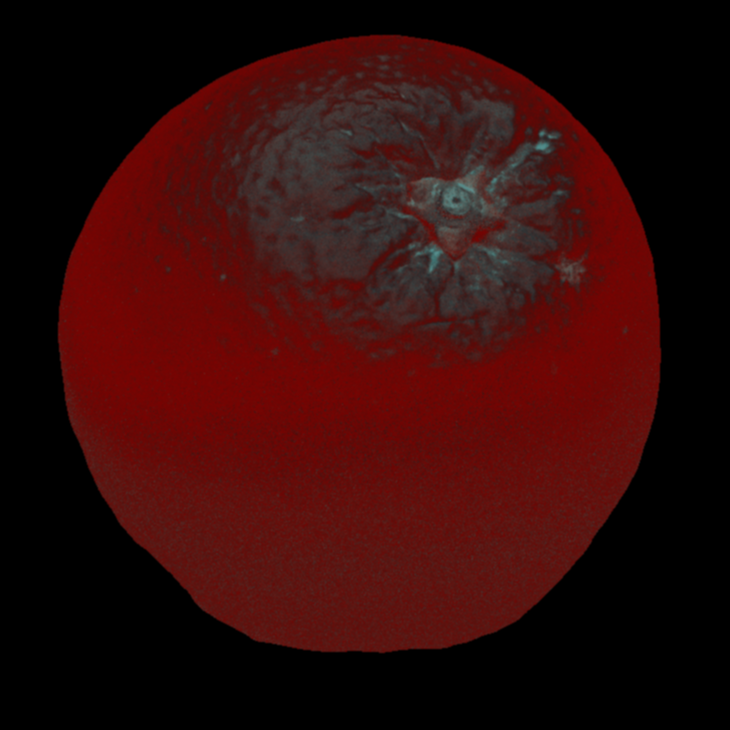}
        \end{minipage}
        \begin{minipage}{.24\linewidth}
            \centering
            \includegraphics[width=\linewidth]{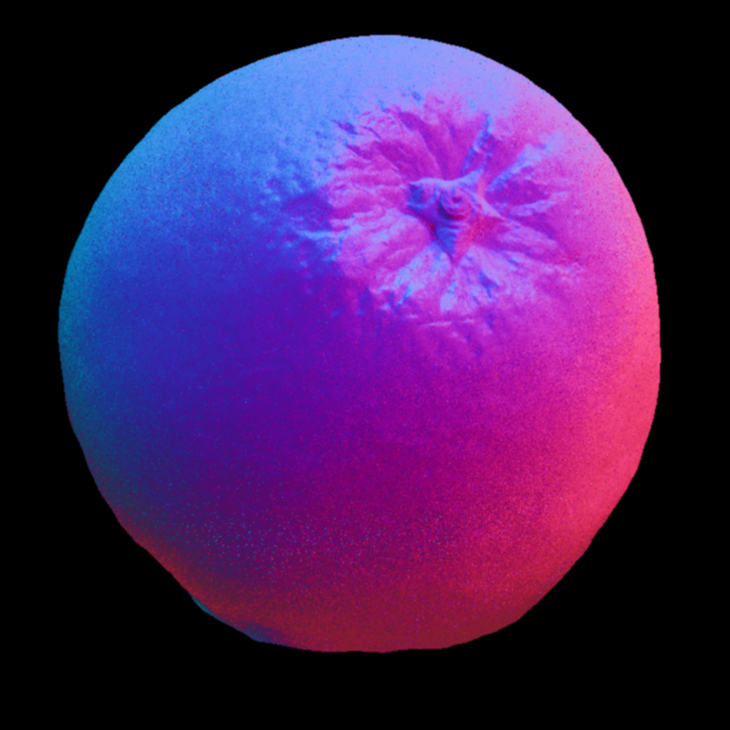}
        \end{minipage}
        \begin{minipage}{.24\linewidth}
            \centering
            \includegraphics[width=\linewidth]{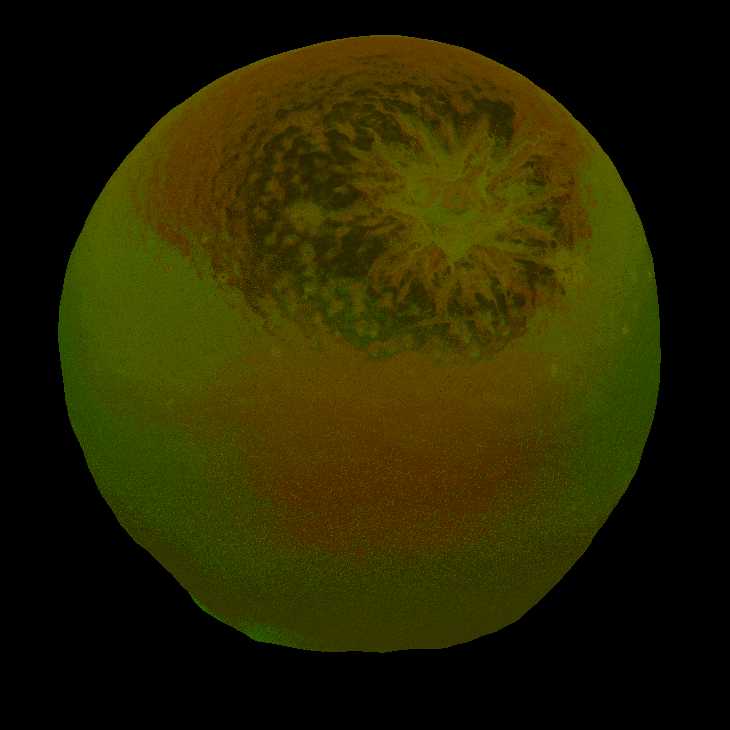}
        \end{minipage}
    \end{minipage}

    \caption{Reflectance results represented as GGX BRDF parameters map. Tangent maps are not shown here due to limited space.}   
    \label{fig:texturemap}
\end{figure}

%% file: figs/geo_main.tex
\begin{figure*}
    \begin{minipage}{\textwidth}
        \begin{minipage}{0.03in}
            \hspace{0.03in}
        \end{minipage}	
        \captionsetup[subfigure]{justification=centering}
        \begin{minipage}{\textwidth}
            \centering
            \begin{minipage}{\textwidth}
                \centering
                \begin{minipage}{.133\textwidth}
                    \centering
                    \subcaption*{\small Ours(Adaptive)}
                \end{minipage}
                \begin{minipage}{.133\textwidth}
                    \centering
                    \subcaption*{\small Ours \\(Non-Adaptive)}
                \end{minipage}
                \begin{minipage}{.133\textwidth}
                    \centering
                    \subcaption*{\small Xu et al.~\cite{xxm_2023_unified} }
                \end{minipage}
                \begin{minipage}{.133\textwidth}
                    \centering
                    \subcaption*{\small MPS~\cite{gupta_2012_mpsSL}}
                \end{minipage}
                \begin{minipage}{.133\textwidth}
                    \centering
                    \subcaption*{\small Photograph}
                \end{minipage}
                \begin{minipage}{.133\textwidth}
                    \centering
                    \subcaption*{\small Ours}
                \end{minipage}
                \begin{minipage}{.133\textwidth}
                    \centering
                    \subcaption*{\small Xu et al.~\cite{xxm_2023_unified}}
                \end{minipage}
            \end{minipage}
        \end{minipage}       
    \end{minipage}

    \begin{minipage}{\textwidth}
        \begin{minipage}{0.03in}	
            \centering
            \rotatebox{90}{\small \textsc{Pig}}
        \end{minipage}
        \begin{minipage}{\textwidth}
            \centering
            \begin{minipage}{.133\textwidth}
                \includegraphics[width=\textwidth]{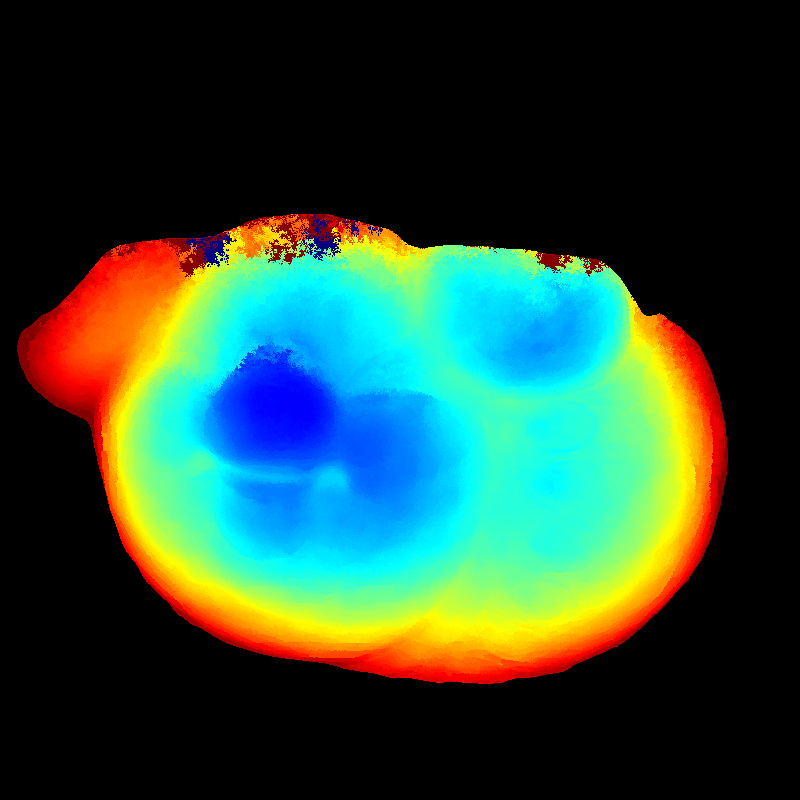}
                \put(-13,3){\scalebox{.8}{\color{black} 0.9}}
            \end{minipage}
            \begin{minipage}{.133\textwidth}
                \includegraphics[width=\textwidth]{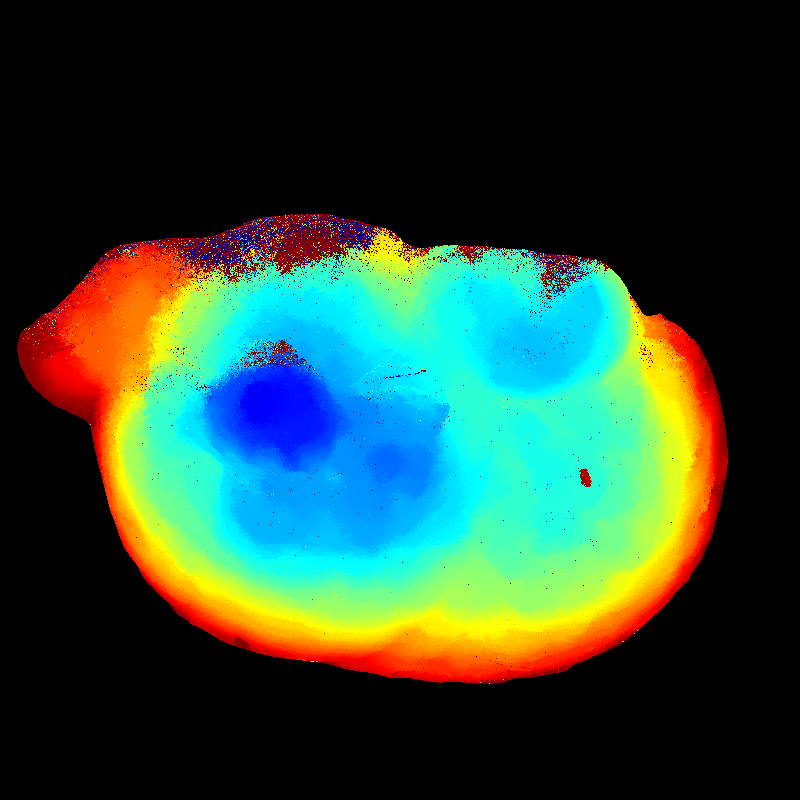}
                \put(-13,3){\scalebox{.8}{\color{black} 0.9}}
            \end{minipage}
            \begin{minipage}{.133\textwidth}
                \includegraphics[width=\textwidth]{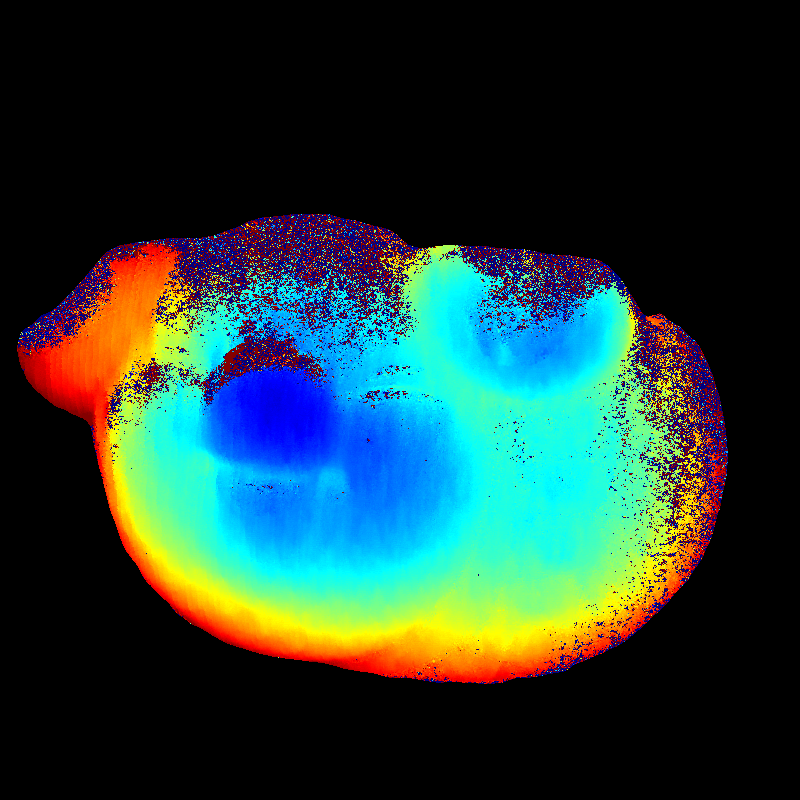}
                \put(-13,3){\scalebox{.8}{\color{black} 0.9}}
            \end{minipage}
            \begin{minipage}{.133\textwidth}
                \includegraphics[width=\textwidth]{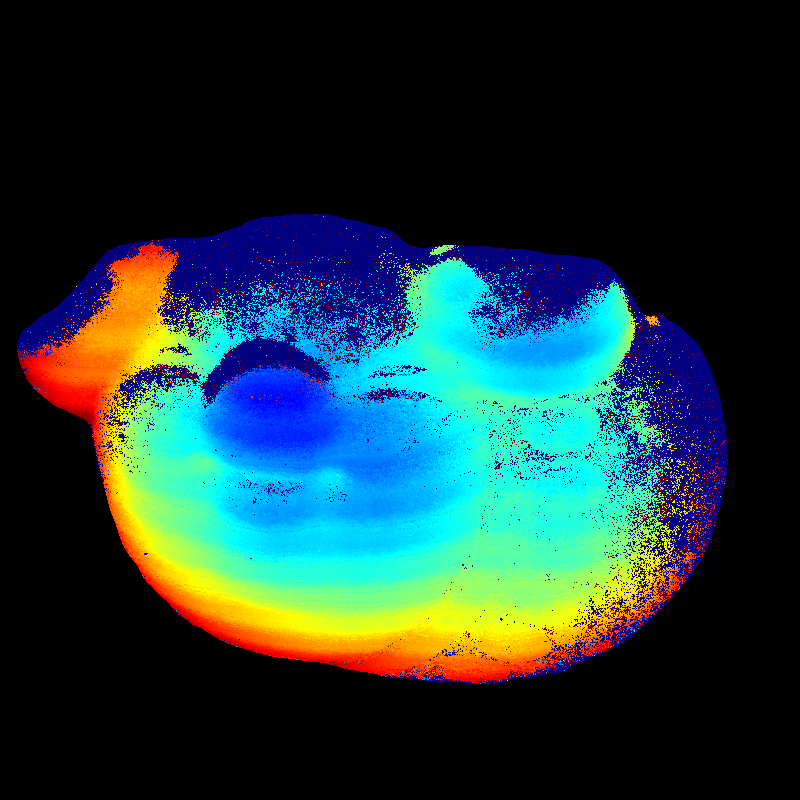}
                \put(-13,3){\scalebox{.8}{\color{black} 4.2}}
            \end{minipage}
            \begin{minipage}{.133\textwidth}
                \includegraphics[width=\textwidth]{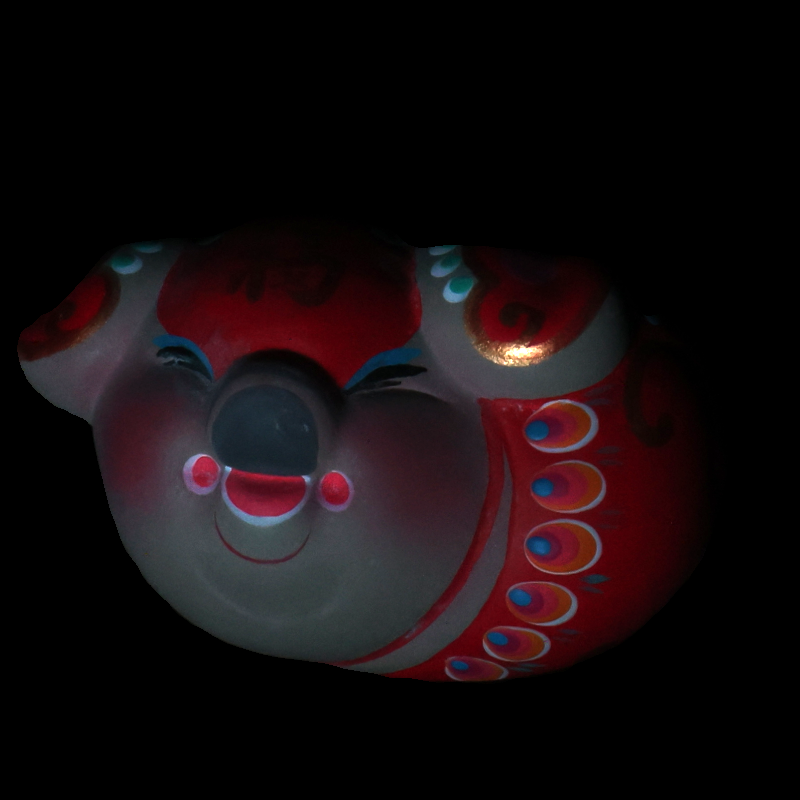}
                \put(-16,3){\scalebox{.8}{\color{black} 0.93}}
            \end{minipage}
            \begin{minipage}{.133\textwidth}
                \includegraphics[width=\textwidth]{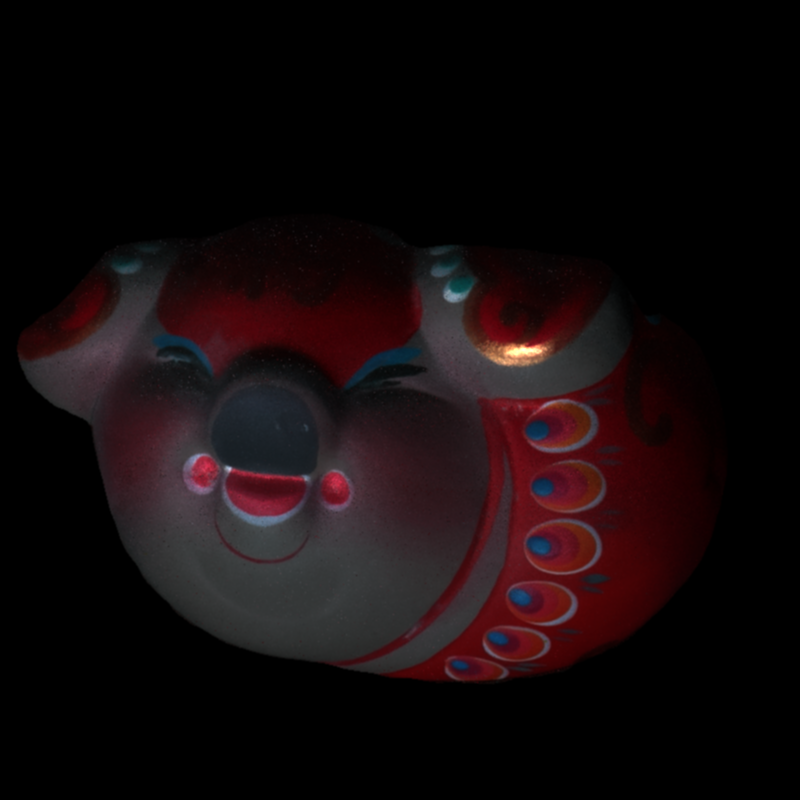}
                \put(-16,3){\scalebox{.8}{\color{black} 48.7}}
            \end{minipage}
            \begin{minipage}{.133\textwidth}
                \includegraphics[width=\textwidth]{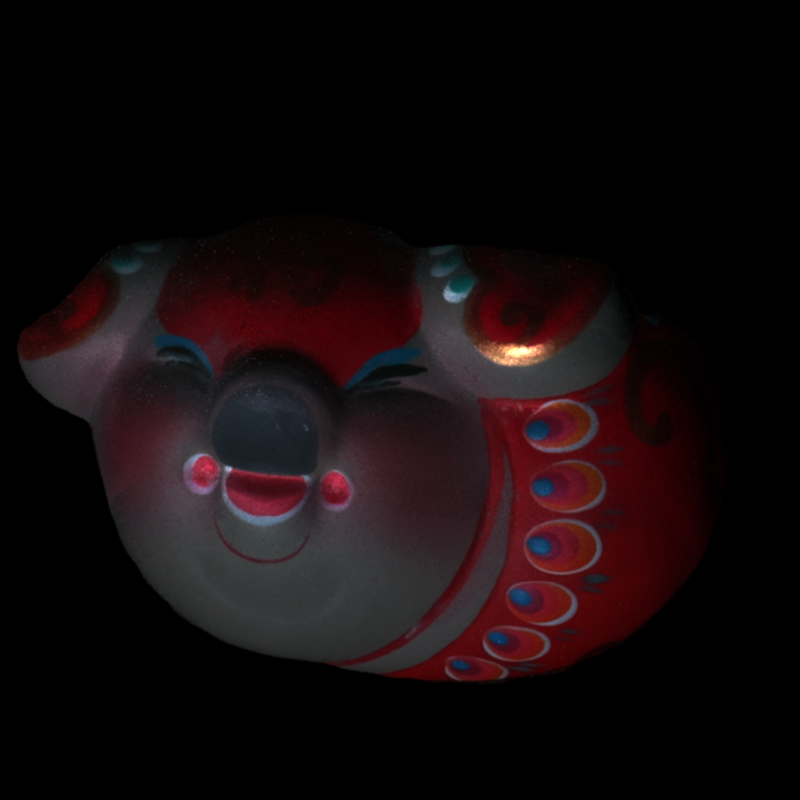}
                \put(-16,3){\scalebox{.8}{\color{black} 0.93}}
            \end{minipage}
        \end{minipage}
    \end{minipage}

    \begin{minipage}{\textwidth}
        \vspace{-1mm}
        \begin{minipage}{0.03in}
            \hspace{0.03in}
        \end{minipage}	
        \captionsetup[subfigure]{justification=centering}
        \begin{minipage}{\textwidth}
            \centering
            \begin{minipage}{\textwidth}
                \centering
                \begin{minipage}{.133\textwidth}
                    \centering
                    \subcaption*{\scriptsize RMSE = 3.75mm \\  
                    RMSE(\%inliers) = 0.31mm (97\% ) }
                \end{minipage}
                \begin{minipage}{.133\textwidth}
                    \centering
                    \subcaption*{\scriptsize RMSE = 4.68mm \\  
                     RMSE(\%inliers) = 0.47mm (94\% )}
                \end{minipage}
                \begin{minipage}{.133\textwidth}
                    \centering
                    \subcaption*{\scriptsize RMSE = 12.12mm \\  
                     RMSE(\%inliers) = 0.60mm (81\% ) }
                \end{minipage}
                \begin{minipage}{.133\textwidth}
                    \centering
                    \subcaption*{\scriptsize RMSE = 66.39mm \\  
                    RMSE(\%inliers) = 1.13mm (60\% )}
                \end{minipage}
                \begin{minipage}{.133\textwidth}
                    \centering
                    \subcaption*{\scriptsize }
                \end{minipage}
                \begin{minipage}{.133\textwidth}
                    \centering
                    \subcaption*{\scriptsize SSIM = 0.96 \\
                        LPIPS = 0.036 \\
                        PSNR = 34.03}
                \end{minipage}
                \begin{minipage}{.133\textwidth}
                    \centering
                    \subcaption*{\scriptsize  SSIM = 0.96 \\
                        LPIPS = 0.034 \\
                        PSNR = 34.40}
                \end{minipage}
            \end{minipage}
        \end{minipage}       
    \end{minipage}

    \begin{minipage}{\textwidth}
        \begin{minipage}{0.03in}	
            \centering
            \rotatebox{90}{\small \textsc{Rabbit}}
        \end{minipage}
        \begin{minipage}{\textwidth}
            \centering
            \begin{minipage}{.133\textwidth}
                \includegraphics[width=\textwidth]{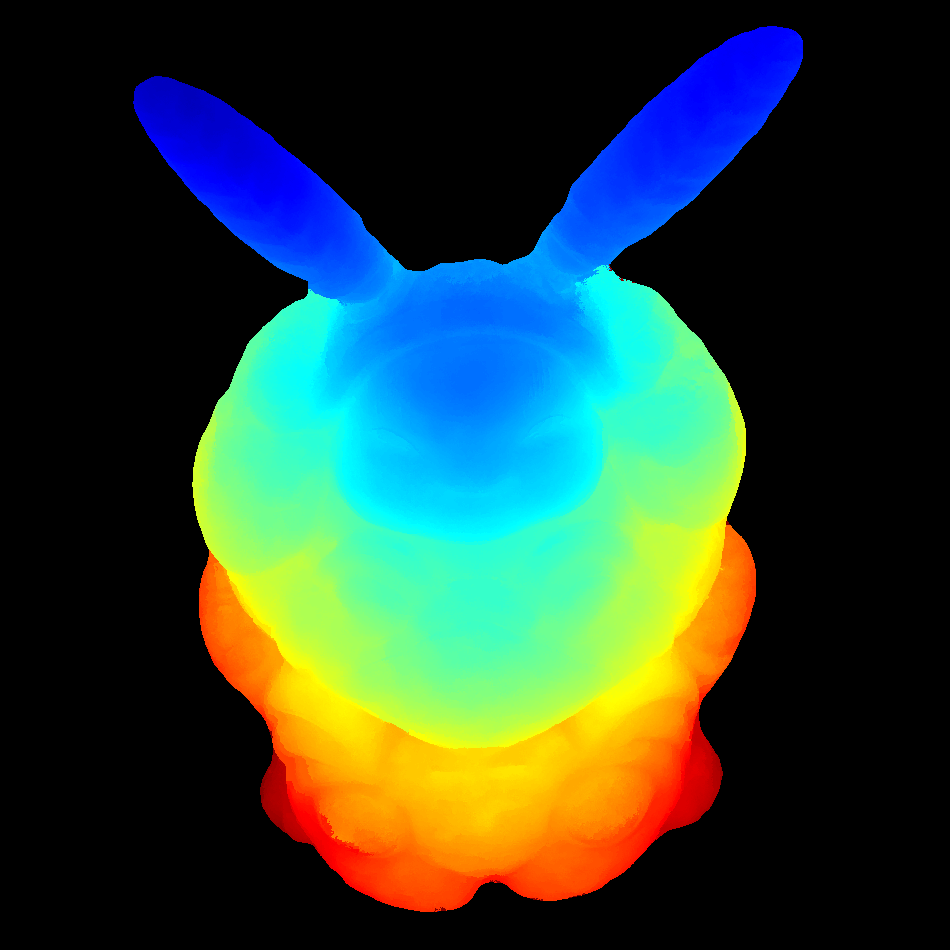}
                \put(-13,3){\scalebox{.8}{\color{black} 6.2}}
            \end{minipage}
            \begin{minipage}{.133\textwidth}
                \includegraphics[width=\textwidth]{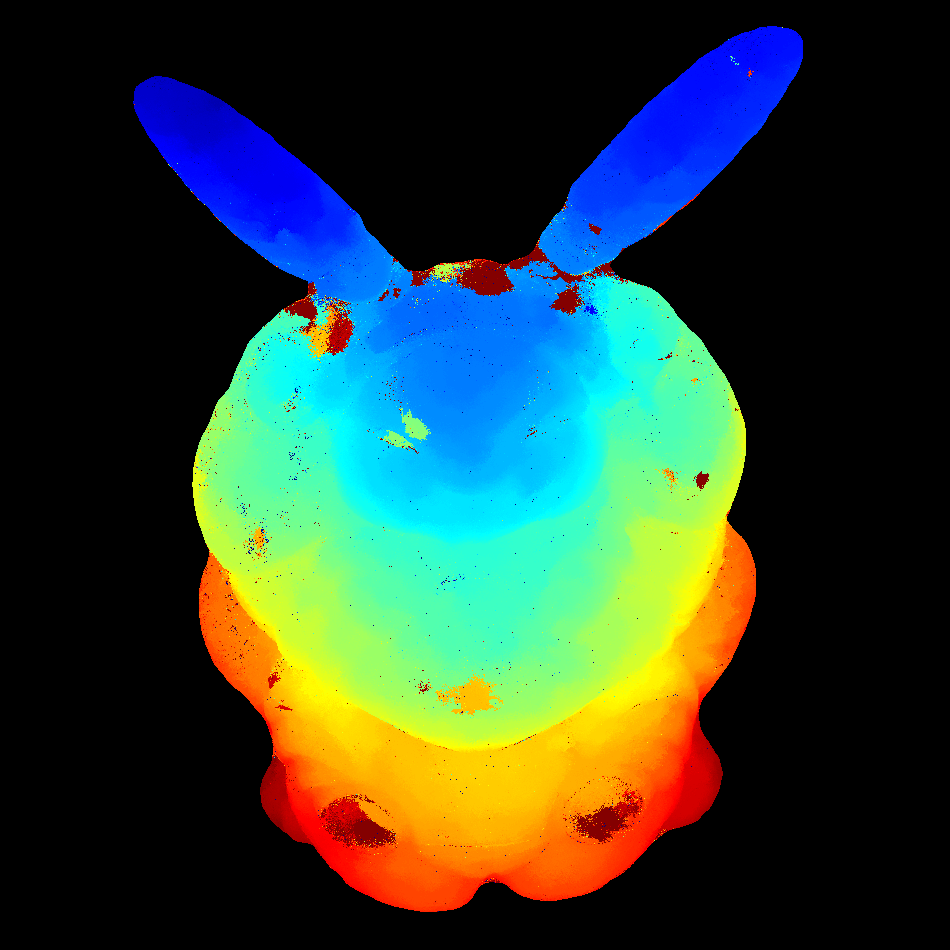}
                \put(-13,3){\scalebox{.8}{\color{black} 6.2}}
            \end{minipage}
            \begin{minipage}{.133\textwidth}
                \includegraphics[width=\textwidth]{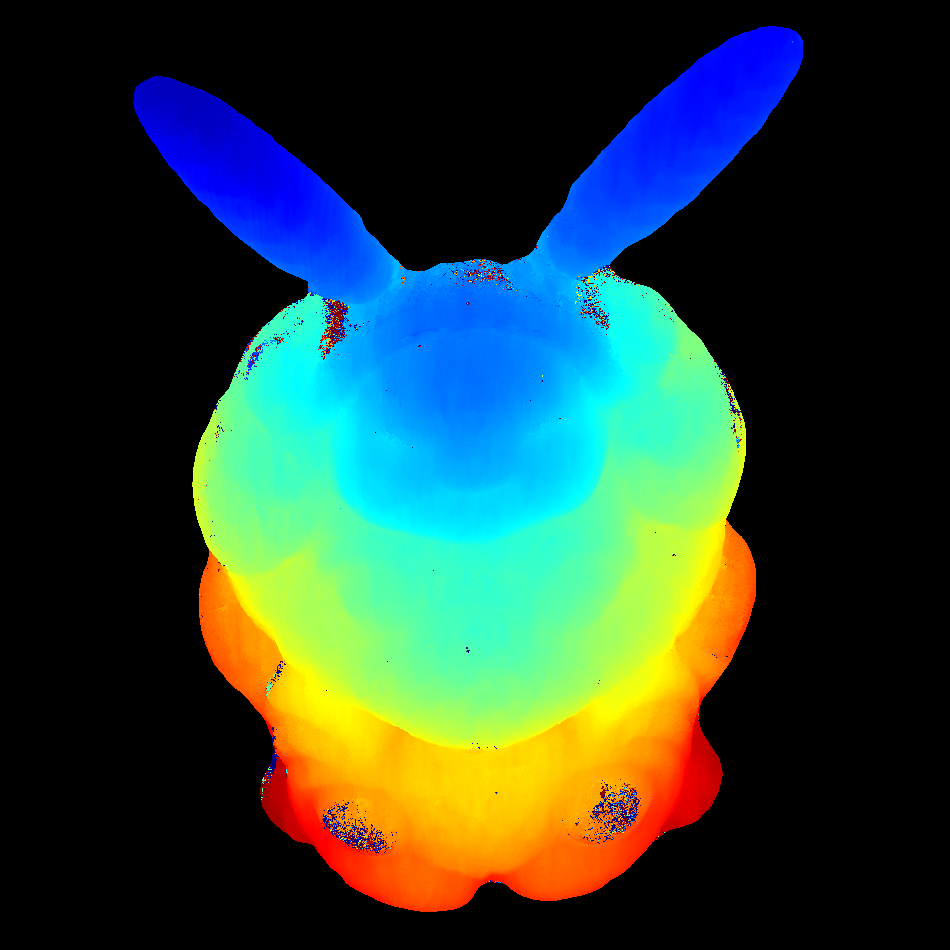}
                \put(-13,3){\scalebox{.8}{\color{black} 6.2}}
            \end{minipage}
            \begin{minipage}{.133\textwidth}
                \includegraphics[width=\textwidth]{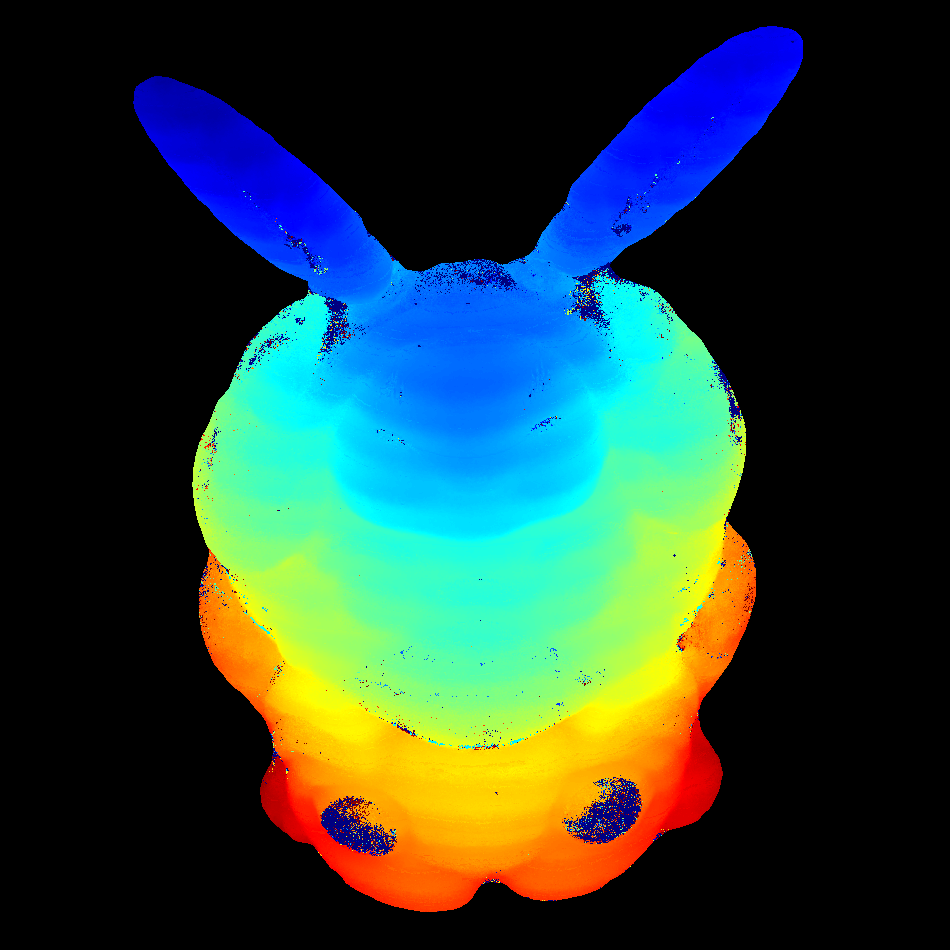}
                \put(-13,3){\scalebox{.8}{\color{black} 6.2}}
            \end{minipage}
            \begin{minipage}{.133\textwidth}
                \includegraphics[width=\textwidth]{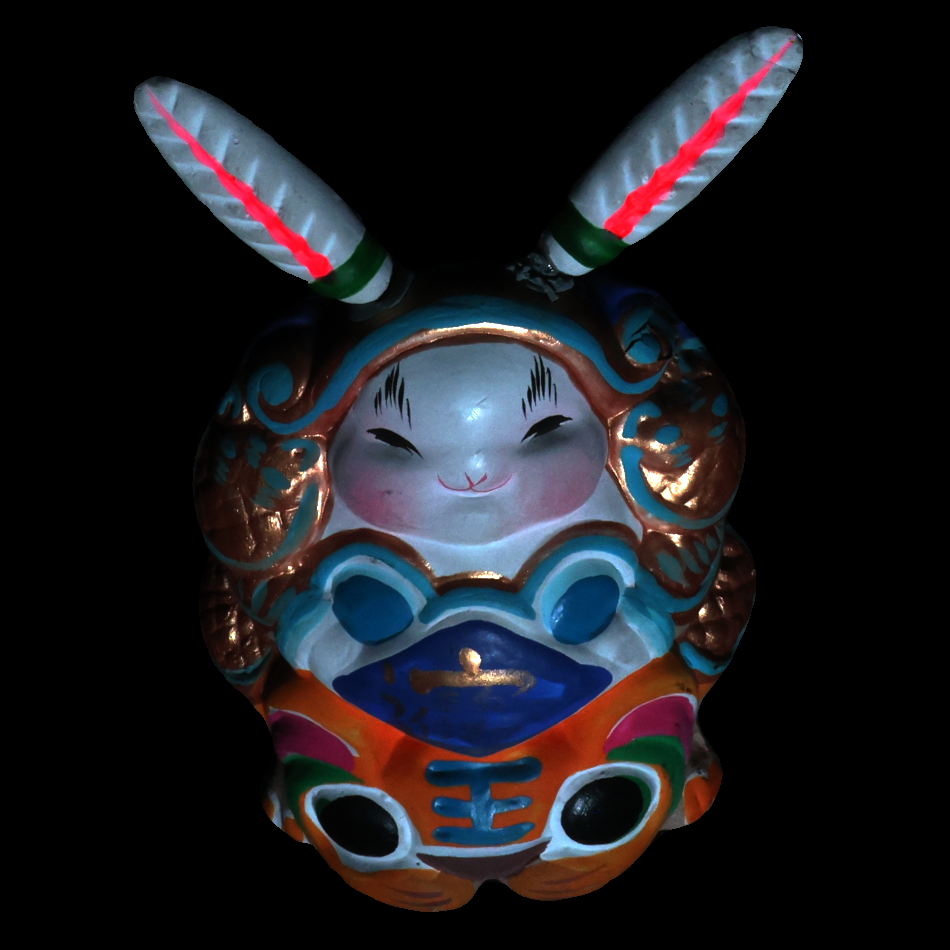}
                \put(-16,3){\scalebox{.8}{\color{black} 78.0}}
            \end{minipage}
            \begin{minipage}{.133\textwidth}
                \includegraphics[width=\textwidth]{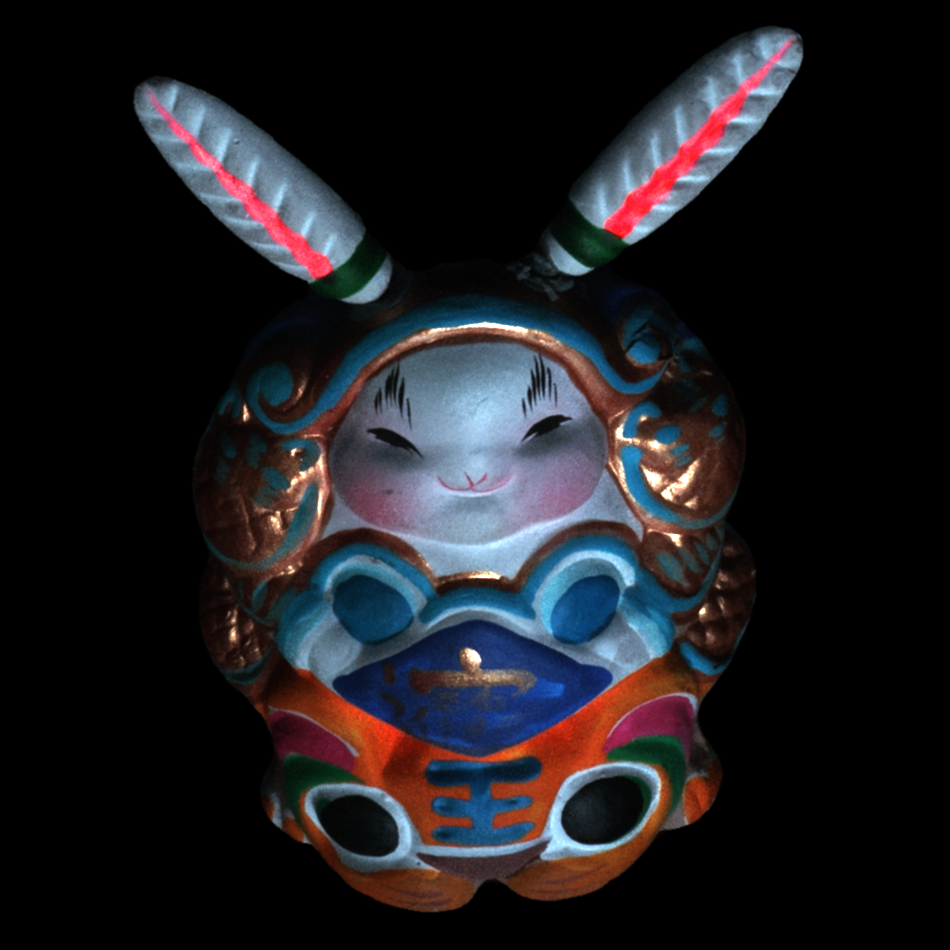}
                \put(-16,3){\scalebox{.8}{\color{black} 15.4}}
            \end{minipage}
            \begin{minipage}{.133\textwidth}
                \includegraphics[width=\textwidth]{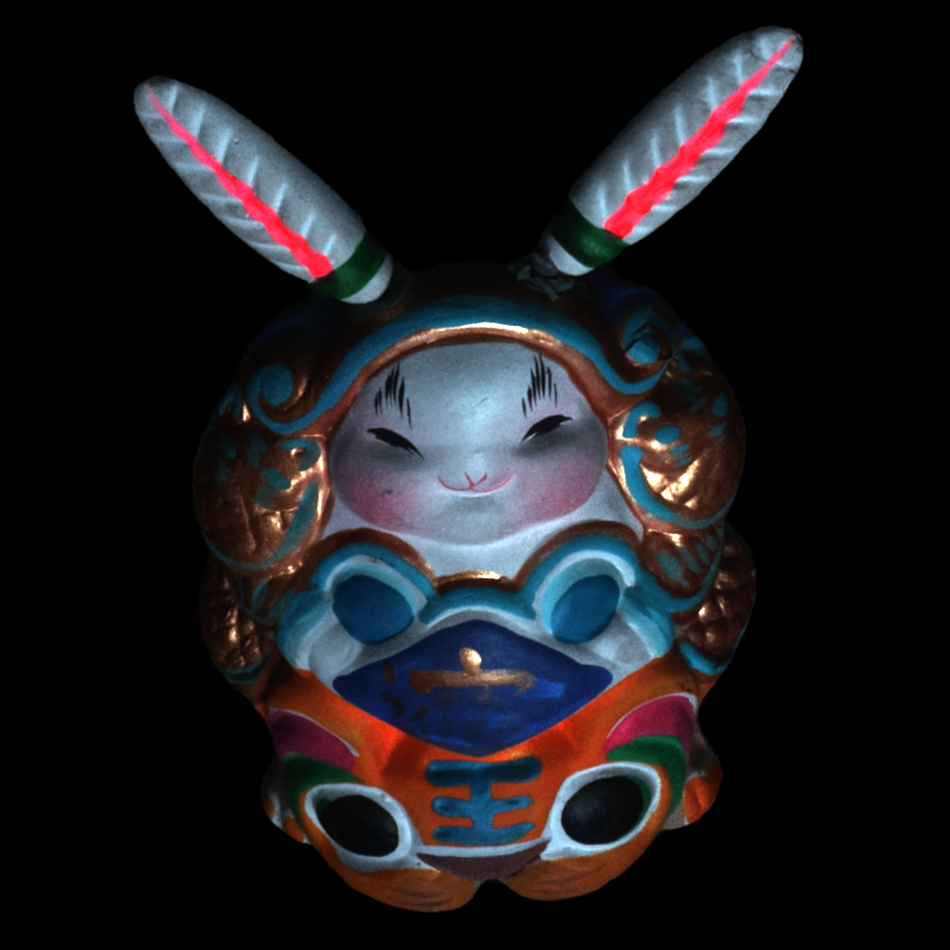}
                \put(-16,3){\scalebox{.8}{\color{black} 20.0}}
            \end{minipage}
        \end{minipage}
    \end{minipage}

    \begin{minipage}{\textwidth}
        \vspace{-1mm}
        \begin{minipage}{0.03in}
            \hspace{0.03in}
        \end{minipage}	
        \captionsetup[subfigure]{justification=centering}
        \begin{minipage}{\textwidth}
            \centering
            \begin{minipage}{\textwidth}
                \centering
                \begin{minipage}{.133\textwidth}
                    \centering
                    \subcaption*{\scriptsize RMSE = 2.26mm \\  
                    RMSE(\%inliers) = 0.28mm (99\% ) }
                \end{minipage}
                \begin{minipage}{.133\textwidth}
                    \centering
                    \subcaption*{\scriptsize RMSE = 3.14mm \\  
                     RMSE(\%inliers) = 0.44mm (97\% )}
                \end{minipage}
                \begin{minipage}{.133\textwidth}
                    \centering
                    \subcaption*{\scriptsize RMSE = 4.21mm \\  
                     RMSE(\%inliers) = 0.52mm (94\% ) }
                \end{minipage}
                \begin{minipage}{.133\textwidth}
                    \centering
                    \subcaption*{\scriptsize RMSE = 8.99mm \\  
                    RMSE(\%inliers) = 1.04mm (95\% )}
                \end{minipage}
                \begin{minipage}{.133\textwidth}
                    \centering
                    \subcaption*{\scriptsize }
                \end{minipage}
                \begin{minipage}{.133\textwidth}
                    \centering
                    \subcaption*{\scriptsize SSIM = 0.95 \\
                        LPIPS = 0.043 \\
                        PSNR = 31.81}
                \end{minipage}
                \begin{minipage}{.133\textwidth}
                    \centering
                    \subcaption*{\scriptsize  SSIM = 0.94 \\
                        LPIPS = 0.049 \\
                        PSNR = 29.69}
                \end{minipage}
            \end{minipage}
        \end{minipage}       
    \end{minipage}

    \begin{minipage}{\textwidth}
        \begin{minipage}{0.03in}	
            \centering
            \rotatebox{90}{\small \textsc{Maruko}}
        \end{minipage}
        \begin{minipage}{\textwidth}
            \centering
            \begin{minipage}{.133\textwidth}
                \includegraphics[width=\textwidth]{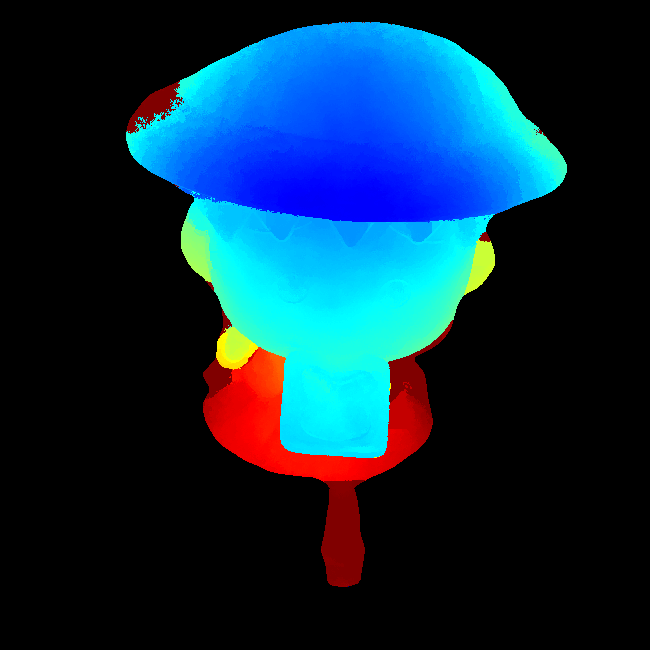}
                \put(-13,3){\scalebox{.8}{\color{black} 6.2}}
            \end{minipage}
            \begin{minipage}{.133\textwidth}
                \includegraphics[width=\textwidth]{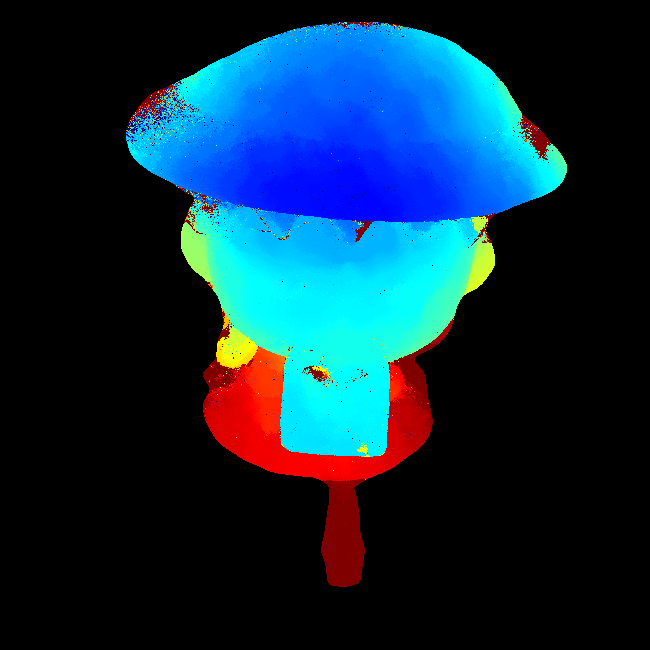}
                \put(-13,3){\scalebox{.8}{\color{black} 6.2}}
            \end{minipage}
            \begin{minipage}{.133\textwidth}
                \includegraphics[width=\textwidth]{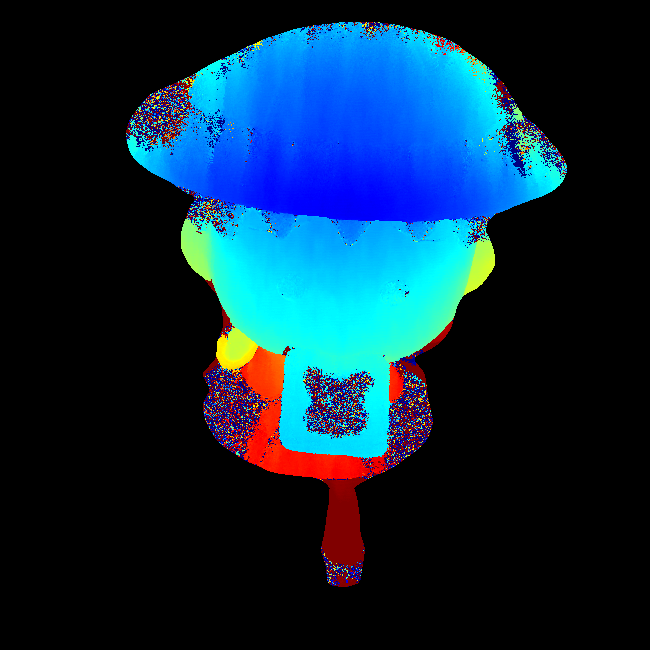}
                \put(-13,3){\scalebox{.8}{\color{black} 6.2}}
            \end{minipage}
            \begin{minipage}{.133\textwidth}
                \includegraphics[width=\textwidth]{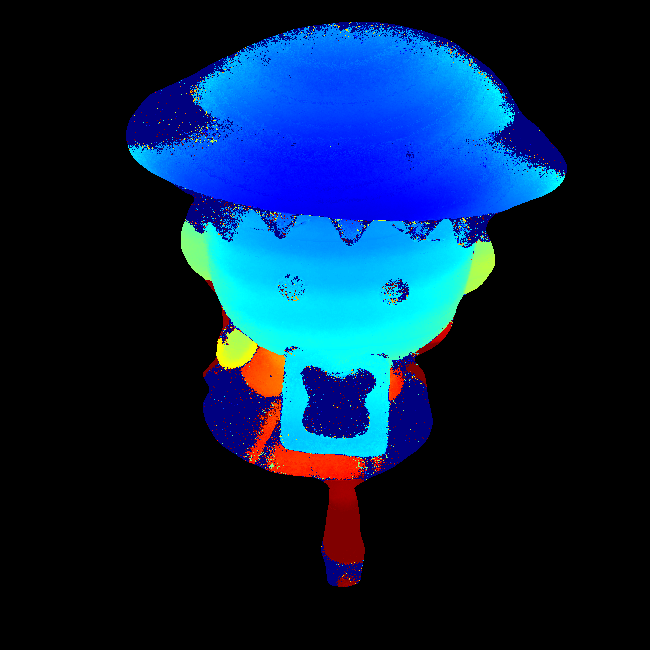}
                \put(-13,3){\scalebox{.8}{\color{black} 6.2}}
            \end{minipage}
            \begin{minipage}{.133\textwidth}
                \includegraphics[width=\textwidth]{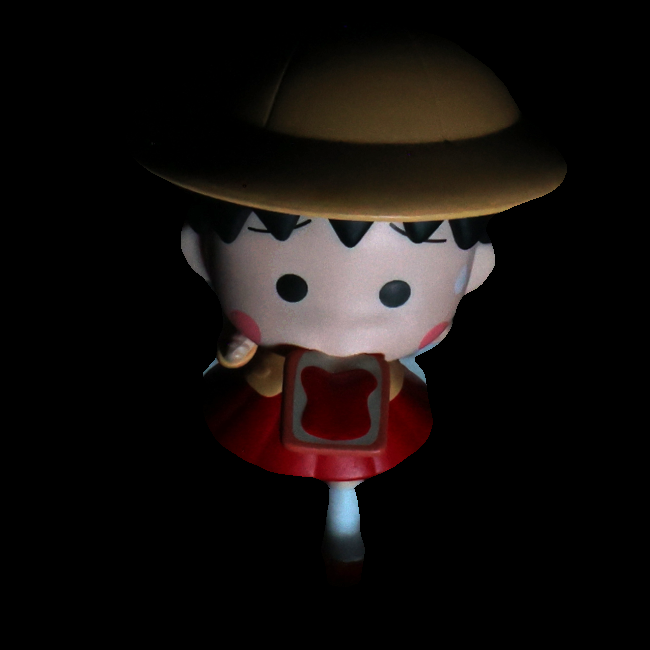}
                \put(-16,3){\scalebox{.8}{\color{black} 78.0}}
            \end{minipage}
            \begin{minipage}{.133\textwidth}
                \includegraphics[width=\textwidth]{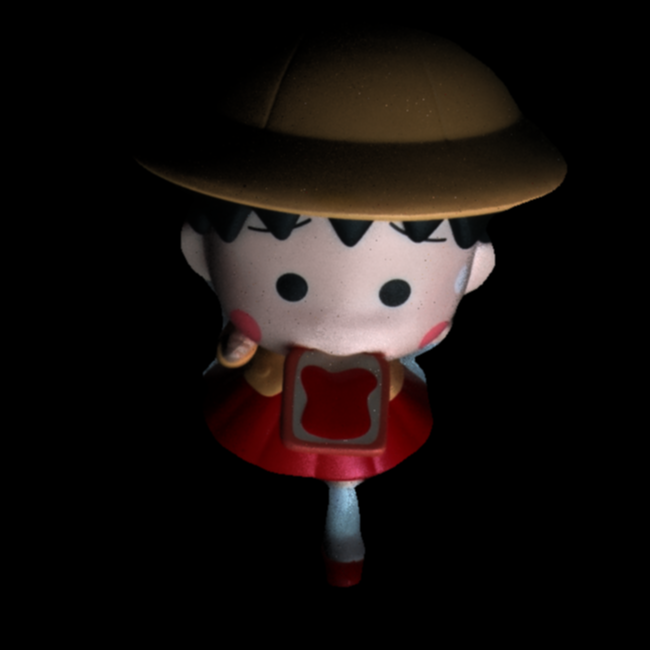}
                \put(-16,3){\scalebox{.8}{\color{black} 15.4}}
            \end{minipage}
            \begin{minipage}{.133\textwidth}
                \includegraphics[width=\textwidth]{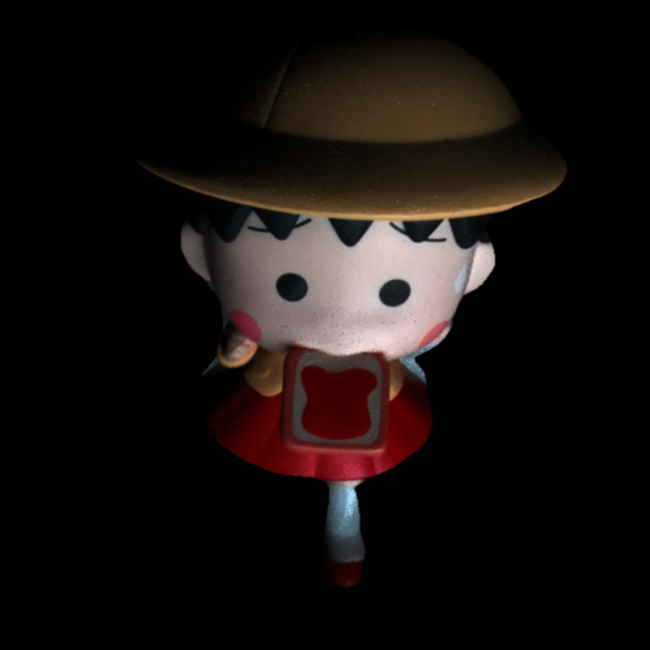}
                \put(-16,3){\scalebox{.8}{\color{black} 20.0}}
            \end{minipage}
        \end{minipage}
    \end{minipage}

    \begin{minipage}{\textwidth}
        \vspace{-1mm}
        \begin{minipage}{0.03in}
            \hspace{0.03in}
        \end{minipage}	
        \captionsetup[subfigure]{justification=centering}
        \begin{minipage}{\textwidth}
            \centering
            \begin{minipage}{\textwidth}
                \centering
                \begin{minipage}{.133\textwidth}
                    \centering
                    \subcaption*{\scriptsize RMSE = 4.78mm \\  
                    RMSE(\%inliers) = 0.42mm (94\% ) }
                \end{minipage}
                \begin{minipage}{.133\textwidth}
                    \centering
                    \subcaption*{\scriptsize RMSE = 5.61mm \\  
                     RMSE(\%inliers) = 0.56mm (92\% )}
                \end{minipage}
                \begin{minipage}{.133\textwidth}
                    \centering
                    \subcaption*{\scriptsize RMSE = 9.97mm \\  
                     RMSE(\%inliers) = 0.44mm (83\% ) }
                \end{minipage}
                \begin{minipage}{.133\textwidth}
                    \centering
                    \subcaption*{\scriptsize RMSE = 56.53mm \\  
                    RMSE(\%inliers) = 1.19mm (76\% )}
                \end{minipage}
                \begin{minipage}{.133\textwidth}
                    \centering
                    \subcaption*{\scriptsize }
                \end{minipage}
                \begin{minipage}{.133\textwidth}
                    \centering
                    \subcaption*{\scriptsize  SSIM = 0.96 \\
                        LPIPS = 0.033 \\
                        PSNR = 31.81}
                \end{minipage}
                \begin{minipage}{.133\textwidth}
                    \centering
                    \subcaption*{\scriptsize SSIM = 0.96 \\
                        LPIPS = 0.034 \\
                        PSNR = 31.42}
                \end{minipage}

            \end{minipage}
        \end{minipage}       
    \end{minipage}

    \begin{minipage}{\textwidth}
        \begin{minipage}{0.03in}	
            \centering
            \rotatebox{90}{\small \textsc{Hedgehog}}
        \end{minipage}
        \begin{minipage}{\textwidth}
            \centering
            \begin{minipage}{.133\textwidth}
                \includegraphics[width=\textwidth]{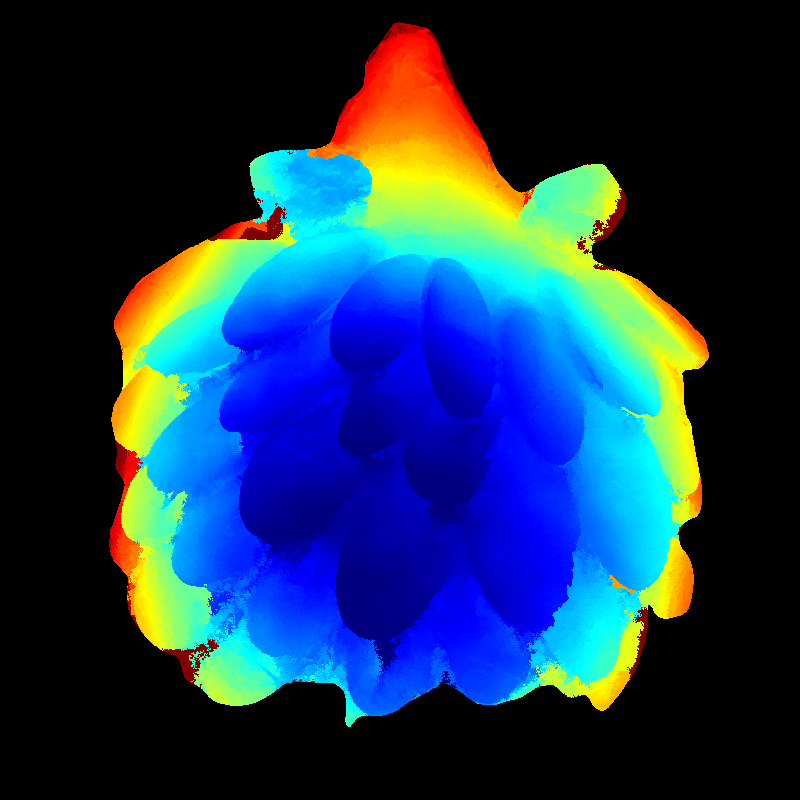}
                \put(-13,3){\scalebox{.8}{\color{black} 6.2}}
            \end{minipage}
            \begin{minipage}{.133\textwidth}
                \includegraphics[width=\textwidth]{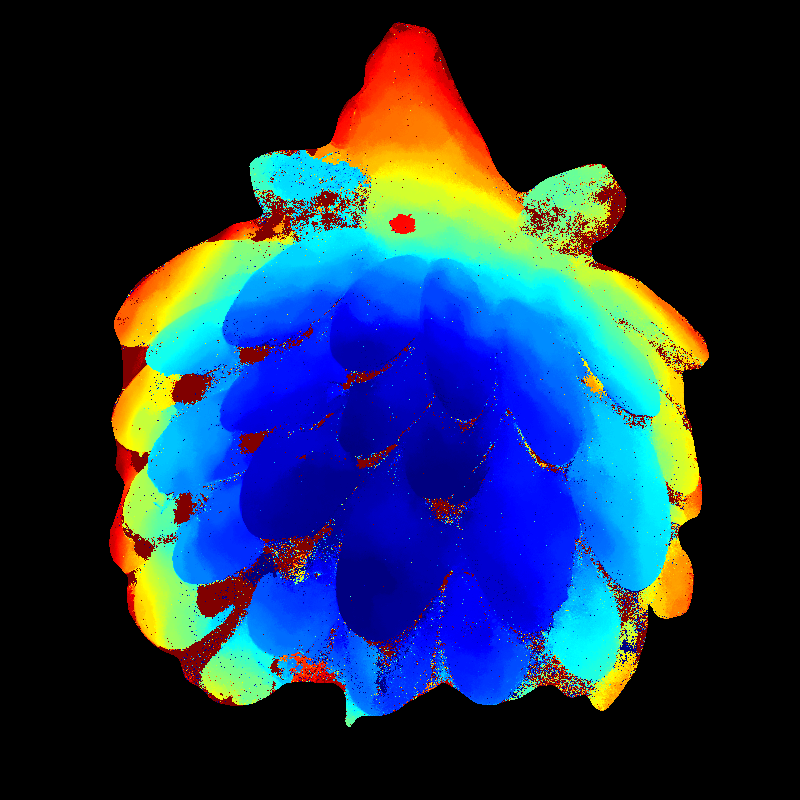}
                \put(-13,3){\scalebox{.8}{\color{black} 6.2}}
            \end{minipage}
            \begin{minipage}{.133\textwidth}
                \includegraphics[width=\textwidth]{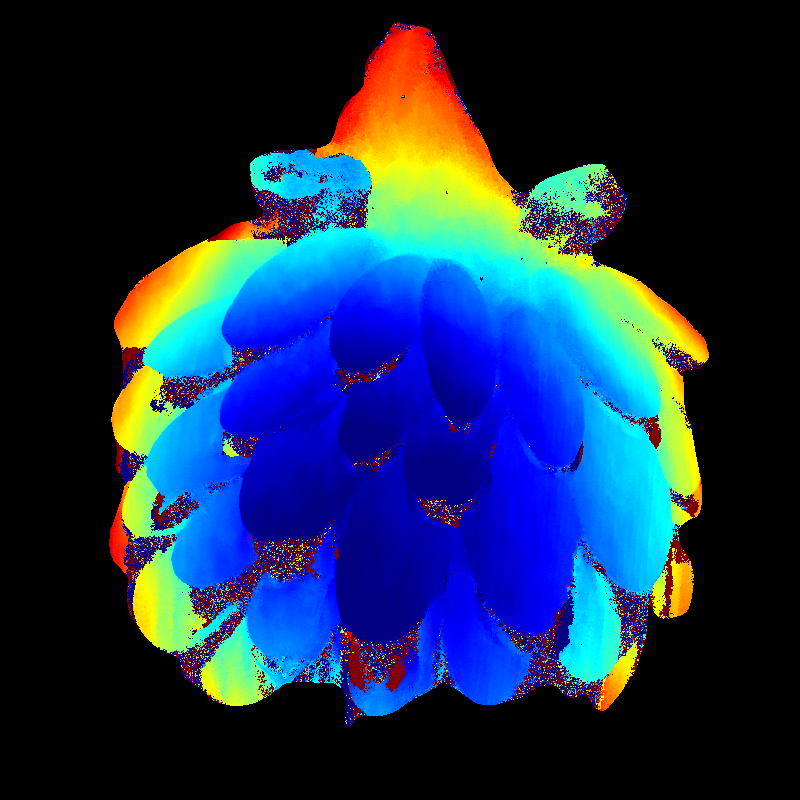}
                \put(-13,3){\scalebox{.8}{\color{black} 6.2}}
            \end{minipage}
            \begin{minipage}{.133\textwidth}
                \includegraphics[width=\textwidth]{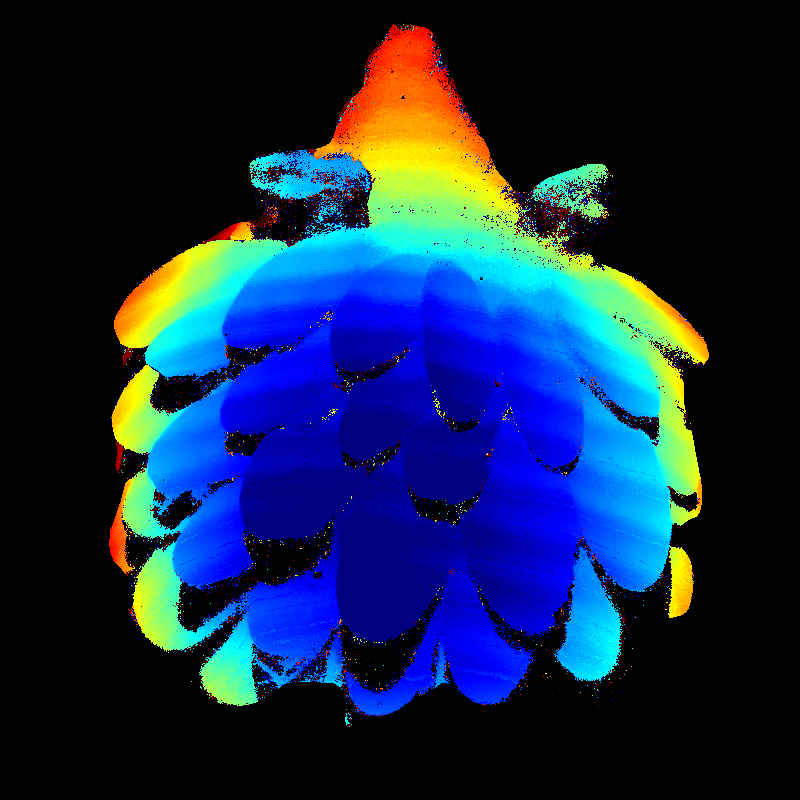}
                \put(-13,3){\scalebox{.8}{\color{black} 6.2}}
            \end{minipage}
            \begin{minipage}{.133\textwidth}
                \includegraphics[width=\textwidth]{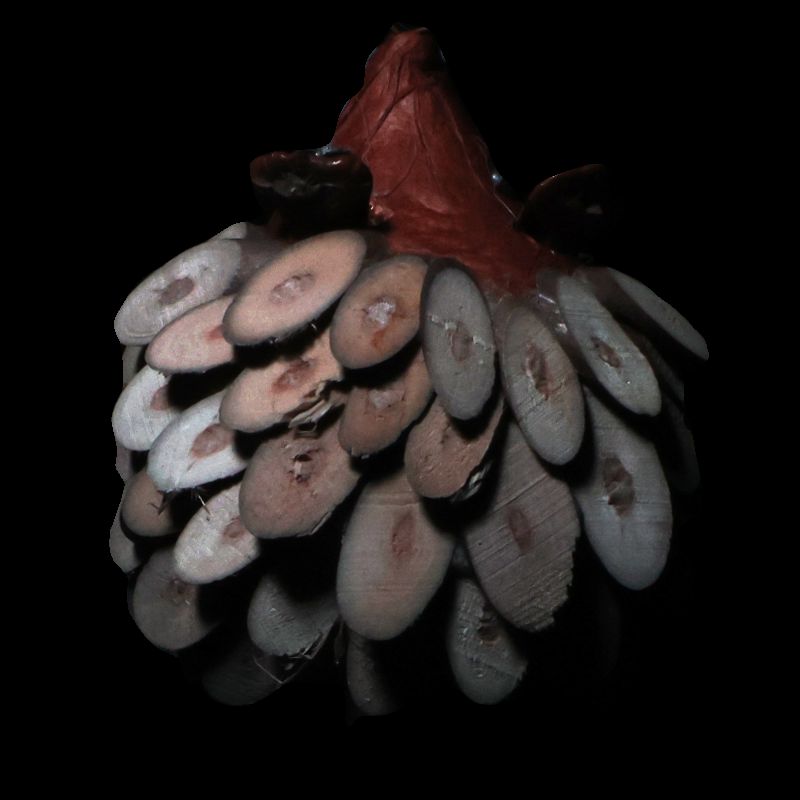}
                \put(-16,3){\scalebox{.8}{\color{black} 78.0}}
            \end{minipage}
            \begin{minipage}{.133\textwidth}
                \includegraphics[width=\textwidth]{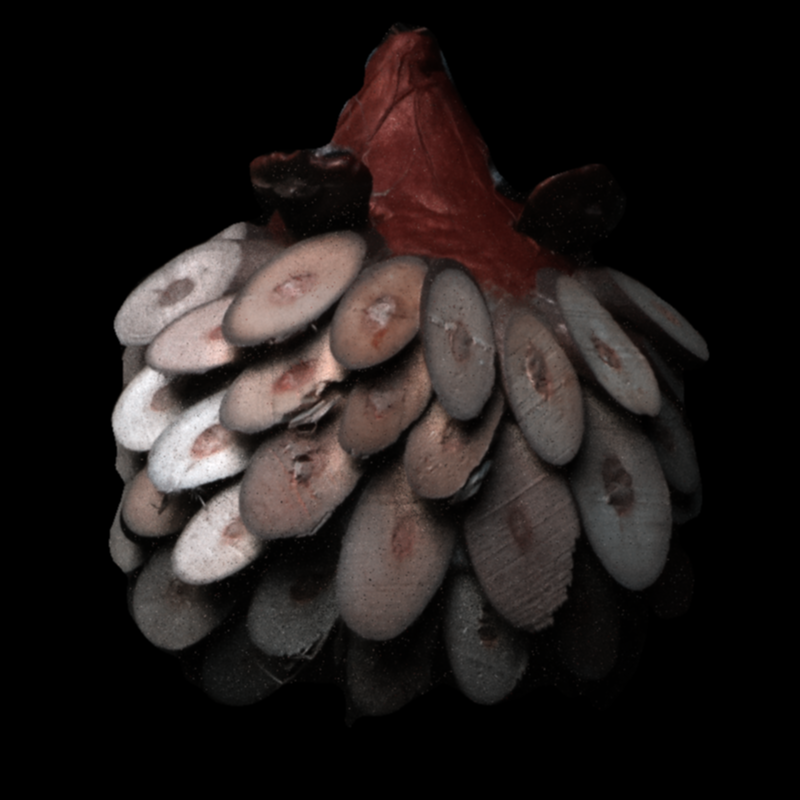}
                \put(-16,3){\scalebox{.8}{\color{black} 15.4}}
            \end{minipage}
            \begin{minipage}{.133\textwidth}
                \includegraphics[width=\textwidth]{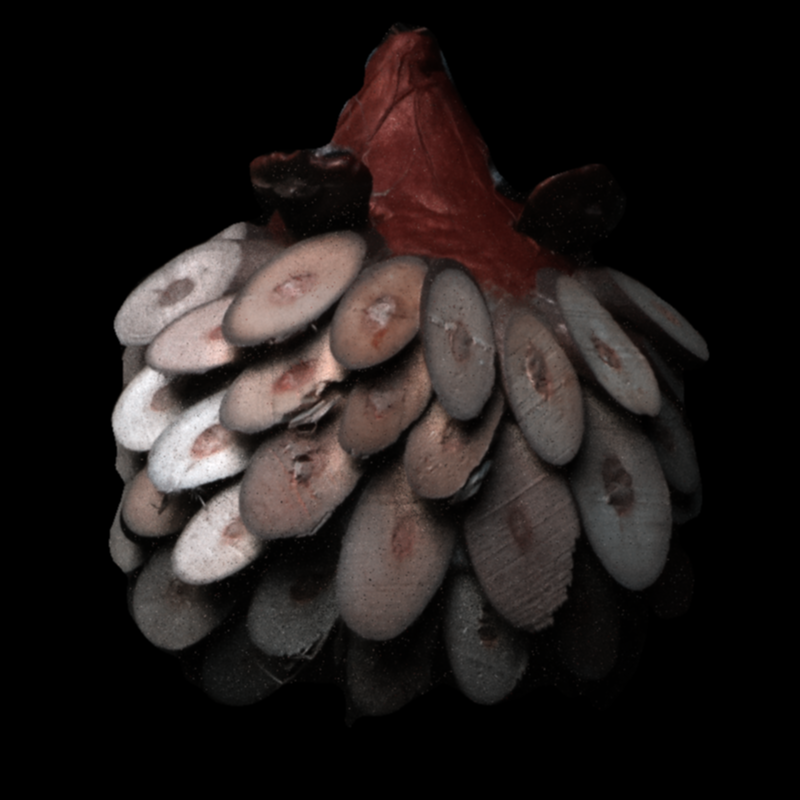}
                \put(-16,3){\scalebox{.8}{\color{black} 20.0}}
            \end{minipage}
        \end{minipage}
    \end{minipage}

    \begin{minipage}{\textwidth}
        \vspace{-1mm}
        \begin{minipage}{0.03in}
            \hspace{0.03in}
        \end{minipage}	
        \captionsetup[subfigure]{justification=centering}
        \begin{minipage}{\textwidth}
            \centering
            \begin{minipage}{\textwidth}
                \centering
                \begin{minipage}{.133\textwidth}
                    \centering
                    \subcaption*{\scriptsize RMSE = 7.30mm \\  
                    RMSE(\%inliers) = 0.50mm (91\% ) }
                \end{minipage}
                \begin{minipage}{.133\textwidth}
                    \centering
                    \subcaption*{\scriptsize RMSE = 11.64mm \\  
                     RMSE(\%inliers) = 0.63mm (85\% ) }
                \end{minipage}
                \begin{minipage}{.133\textwidth}
                    \centering
                    \subcaption*{\scriptsize RMSE = 11.66mm \\  
                     RMSE(\%inliers) = 0.61mm (82\% )}
                \end{minipage}
                \begin{minipage}{.133\textwidth}
                    \centering
                    \subcaption*{\scriptsize RMSE = 44.05mm \\  
                    RMSE(\%inliers) = 1.32mm (77\% )}
                \end{minipage}
                \begin{minipage}{.133\textwidth}
                    \centering
                    \subcaption*{\scriptsize }
                \end{minipage}
                \begin{minipage}{.133\textwidth}
                    \centering
                    \subcaption*{\scriptsize SSIM = 0.90 \\
                        LPIPS = 0.082 \\
                        PSNR = 28.72}
                \end{minipage}
                \begin{minipage}{.133\textwidth}
                    \centering
                    \subcaption*{\scriptsize  SSIM = 0.90 \\
                        LPIPS = 0.082 \\
                        PSNR = 29.05}
                \end{minipage}
            \end{minipage}
        \end{minipage}       
    \end{minipage}

  \begin{minipage}{\textwidth}
        \begin{minipage}{0.03in}	
            \centering
            \rotatebox{90}{\small \textsc{Orange}}
        \end{minipage}
        \begin{minipage}{\textwidth}
            \centering
            \begin{minipage}{.133\textwidth}
                \includegraphics[width=\textwidth]{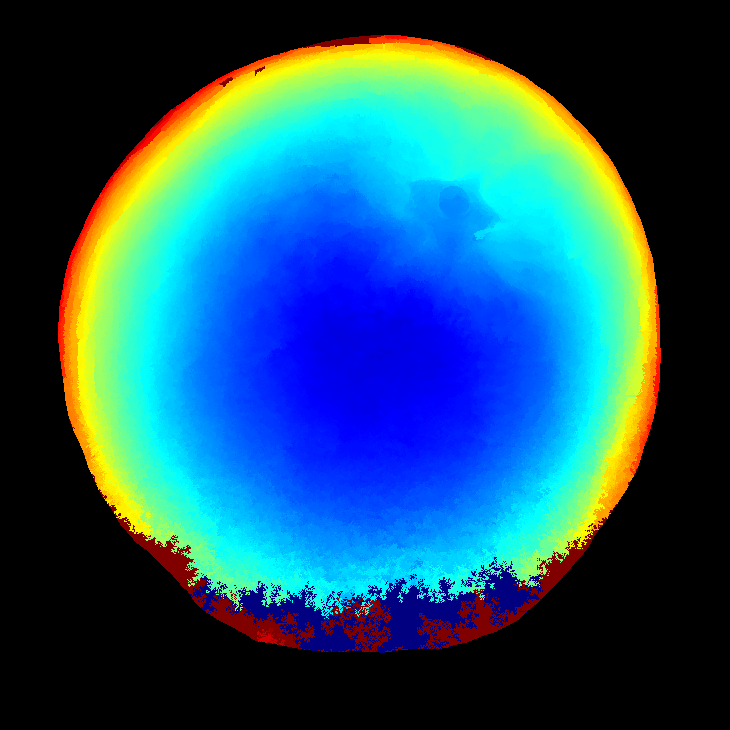}
                \put(-13,3){\scalebox{.8}{\color{black} 6.2}}
            \end{minipage}
            \begin{minipage}{.133\textwidth}
                \includegraphics[width=\textwidth]{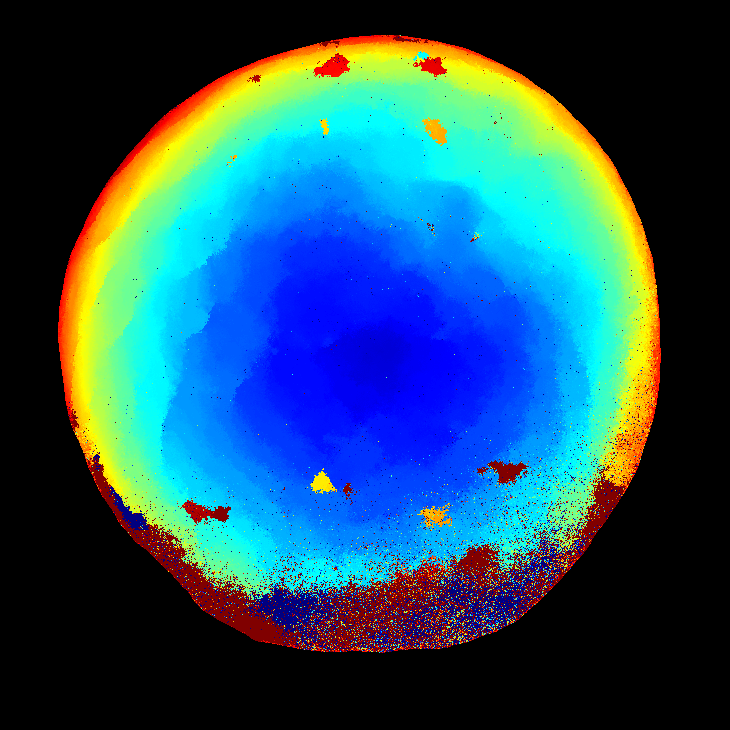}
                \put(-13,3){\scalebox{.8}{\color{black} 6.2}}
            \end{minipage}
            \begin{minipage}{.133\textwidth}
                \includegraphics[width=\textwidth]{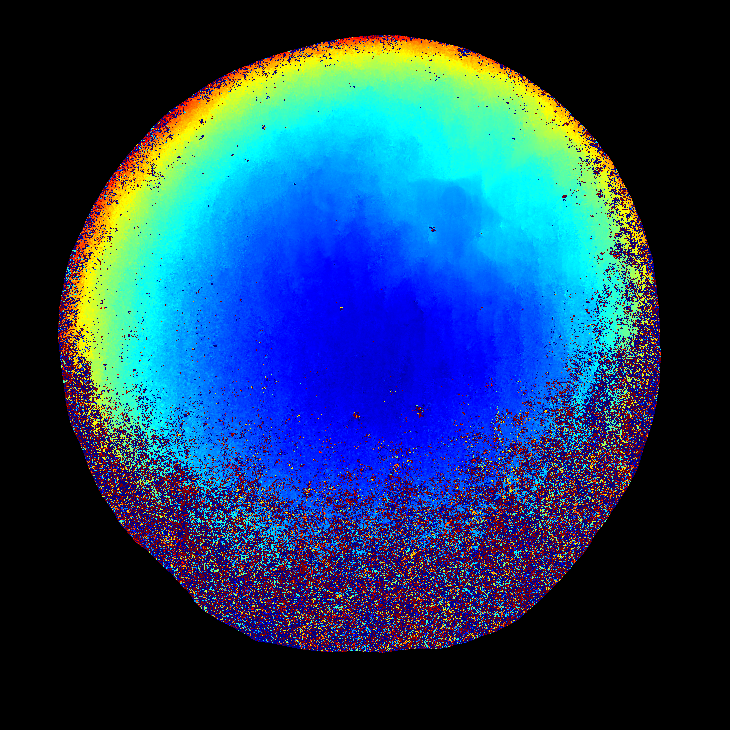}
                \put(-13,3){\scalebox{.8}{\color{black} 6.2}}
            \end{minipage}
            \begin{minipage}{.133\textwidth}
                \includegraphics[width=\textwidth]{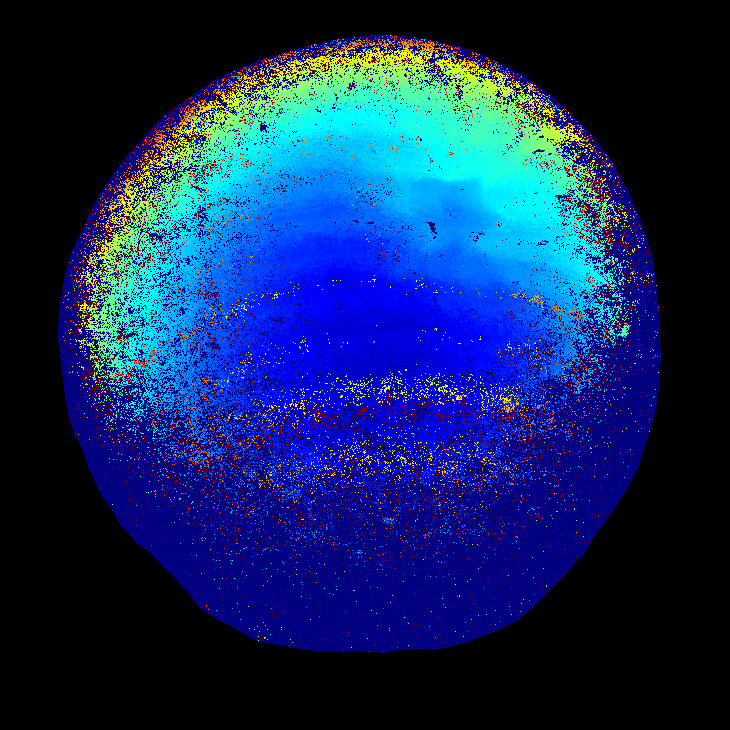}
                \put(-13,3){\scalebox{.8}{\color{black} 6.2}}
            \end{minipage}
            \begin{minipage}{.133\textwidth}
                \includegraphics[width=\textwidth]{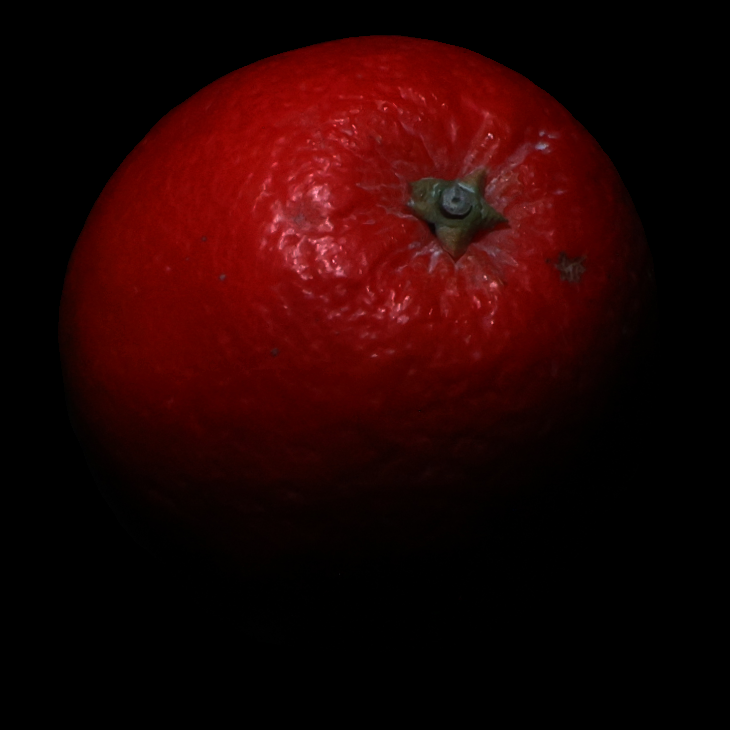}
                \put(-16,3){\scalebox{.8}{\color{black} 78.0}}
            \end{minipage}
            \begin{minipage}{.133\textwidth}
                \includegraphics[width=\textwidth]{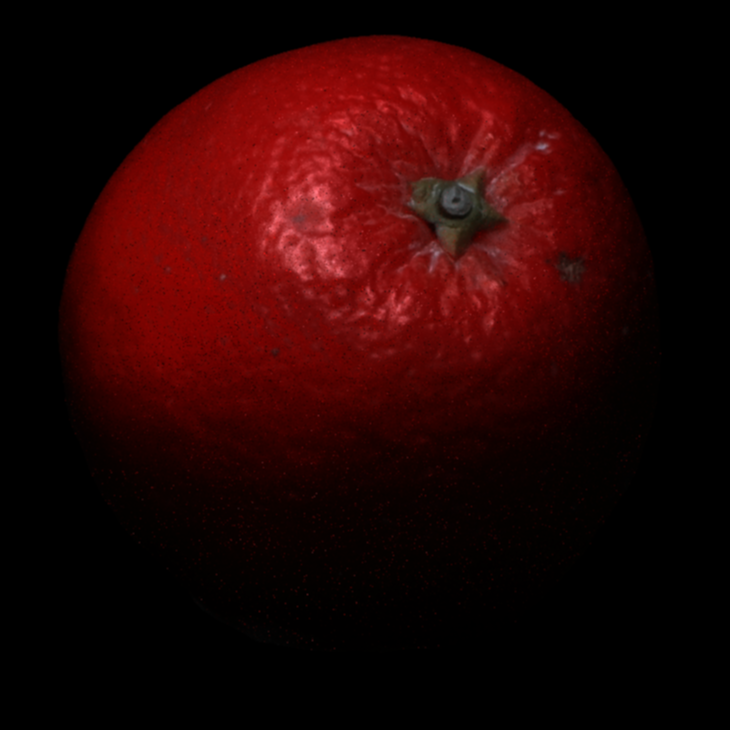}
                \put(-16,3){\scalebox{.8}{\color{black} 15.4}}
            \end{minipage}
            \begin{minipage}{.133\textwidth}
                \includegraphics[width=\textwidth]{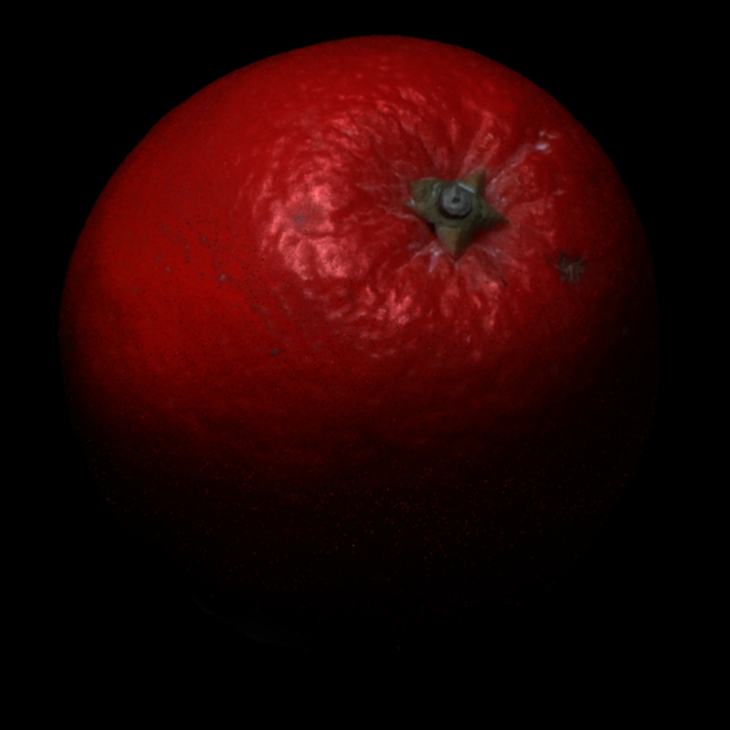}
                \put(-16,3){\scalebox{.8}{\color{black} 20.0}}
            \end{minipage}
        \end{minipage}
    \end{minipage}

   \begin{minipage}{\textwidth}
        \vspace{-1mm}
        \begin{minipage}{0.03in}
            \hspace{0.03in}
        \end{minipage}	
        \captionsetup[subfigure]{justification=centering}
        \begin{minipage}{\textwidth}
            \centering
            \begin{minipage}{\textwidth}
                \centering
               \begin{minipage}{.133\textwidth}
                    \centering
                    \subcaption*{\scriptsize }
                \end{minipage}
                \begin{minipage}{.133\textwidth}
                    \centering
                    \subcaption*{\scriptsize }
                \end{minipage}
               \begin{minipage}{.133\textwidth}
                    \centering
                    \subcaption*{\scriptsize }
                \end{minipage}
                \begin{minipage}{.133\textwidth}
                    \centering
                    \subcaption*{\scriptsize }
                \end{minipage}
                \begin{minipage}{.133\textwidth}
                    \centering
                    \subcaption*{\scriptsize }
                \end{minipage}
                \begin{minipage}{.133\textwidth}
                    \centering
                    \subcaption*{\scriptsize SSIM = 0.91 \\
                        LPIPS = 0.077 \\
                        PSNR = 35.16}
                \end{minipage}
                \begin{minipage}{.133\textwidth}
                    \centering
                    \subcaption*{\scriptsize  SSIM = 0.91 \\
                        LPIPS = 0.066 \\
                        PSNR = 35.28}
                \end{minipage}
            \end{minipage}
        \end{minipage}       
    \end{minipage}

    \caption{Comparisons with state-of-the-art techniques on shape and reflectance capture. From the left column to right: depth reconstruction with our approach (adaptive/non-adaptive patterns), ~\cite{xxm_2023_unified} and MPS~\cite{gupta_2012_mpsSL}; photograph under a lighting condition not used in optimization, rendering with the reflectance results of our approach and~\cite{xxm_2023_unified}. Quantitative errors are listed below each related image.}
    
    \label{fig:geo_main}
\end{figure*}

%% file: figs/cmp_shadow.tex
\begin{figure}[htbp]
    \centering
    \captionsetup[subfigure]{justification=centering}
    \begin{minipage}{\linewidth}
        \centering
        \begin{minipage}{\textwidth}
            \centering
            \begin{minipage}{.24\textwidth}
                \centering 
                \subcaption*{\small Ours}
            \end{minipage}
            \begin{minipage}{.72\textwidth}
                \centering
                \subcaption*{\small ~\cite{xxm_2023_unified} with 1 different LED on at a time}
            \end{minipage}
        \end{minipage}
    \end{minipage}  
    \begin{minipage}{\linewidth}
        \center
        \begin{minipage}{.24\linewidth}
            \centering
            \includegraphics[width=\linewidth]{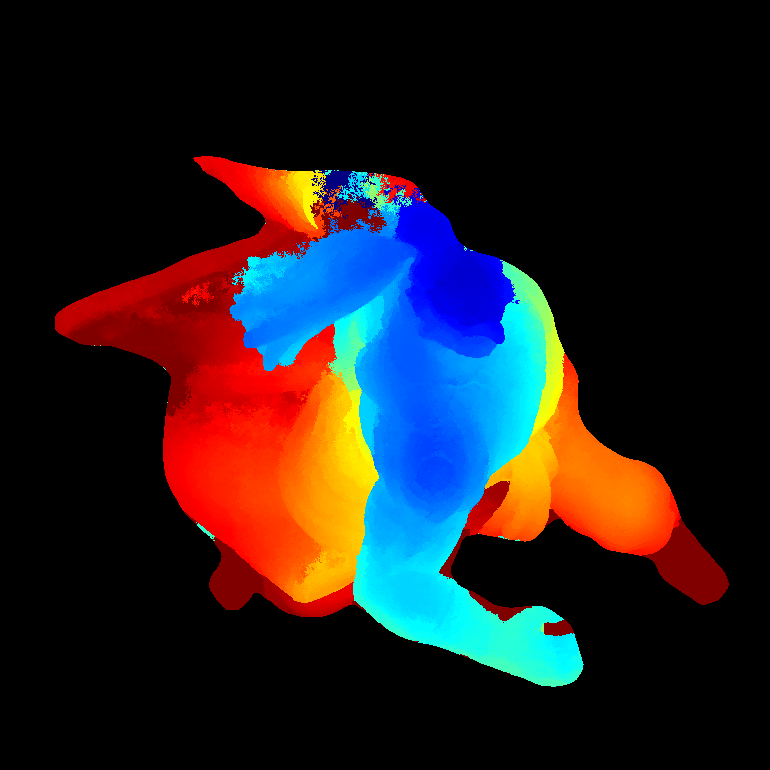}
        \end{minipage}
        \begin{minipage}{.24\linewidth}
            \centering
            \includegraphics[width=\linewidth]{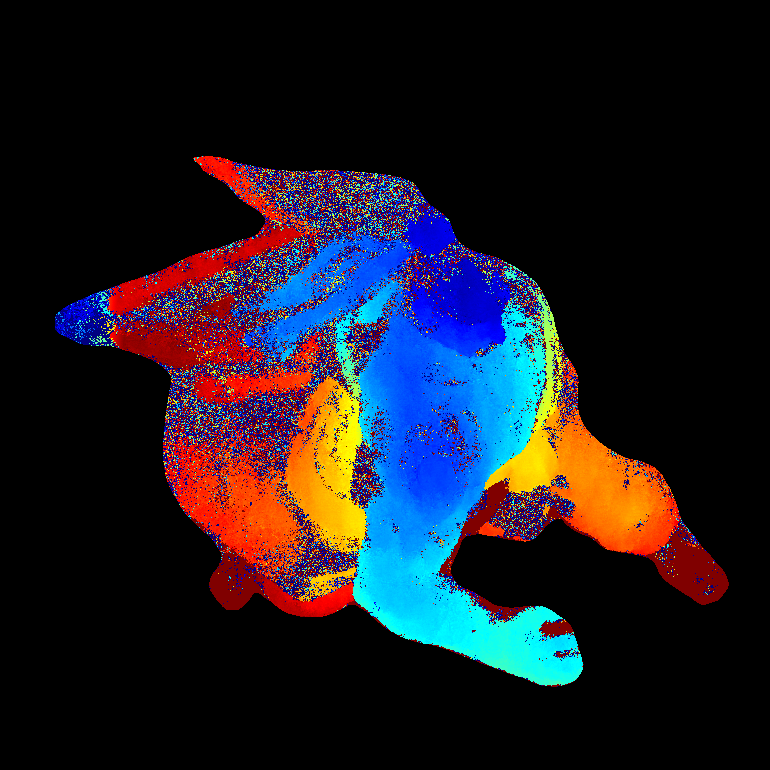}
        \end{minipage}
        \begin{minipage}{.24\linewidth}
            \centering
            \includegraphics[width=\linewidth]{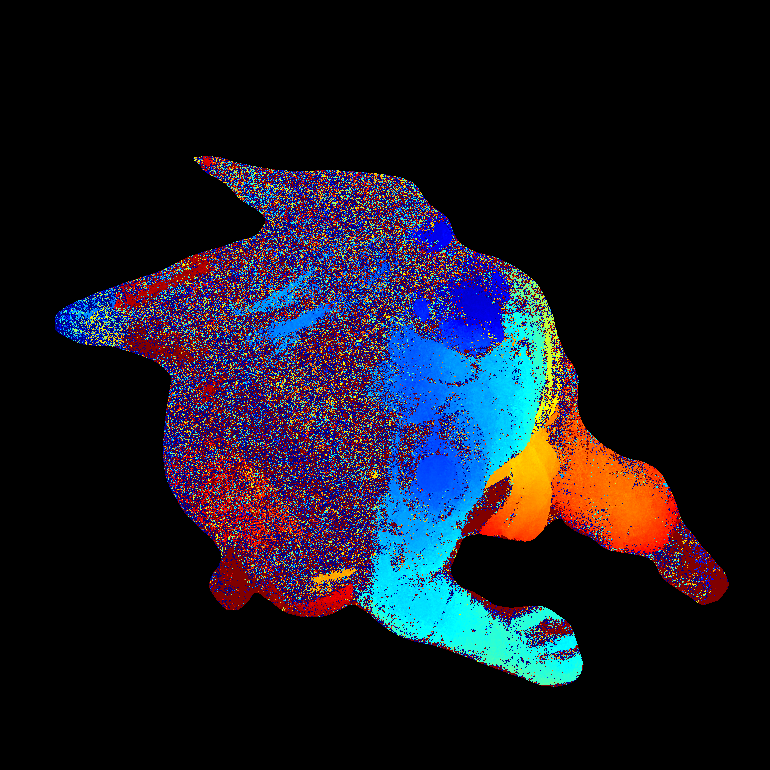}
        \end{minipage}
        \begin{minipage}{.24\linewidth}
            \centering
            \includegraphics[width=\linewidth]{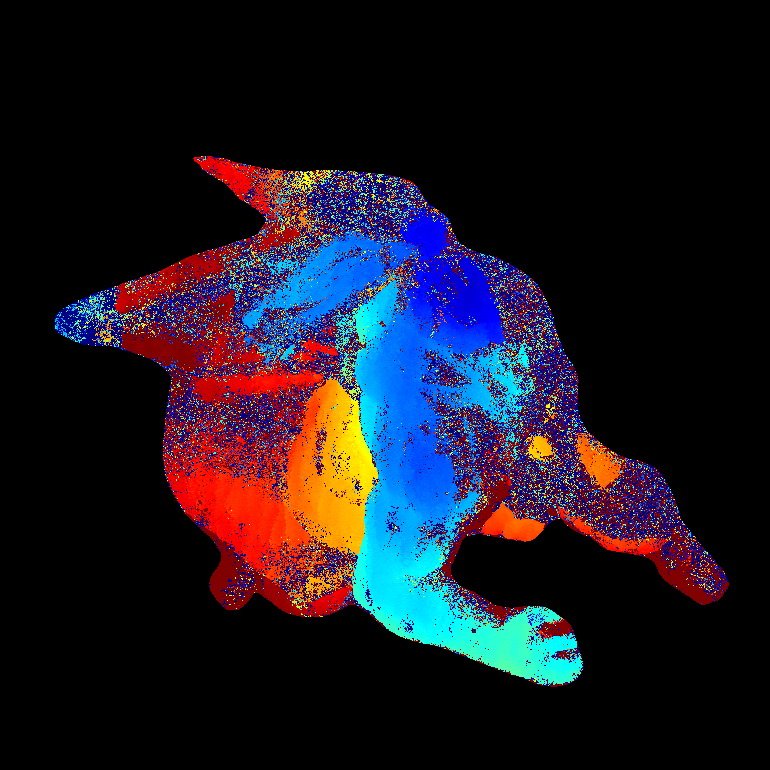}
        \end{minipage}
    \end{minipage}
    \begin{minipage}{\linewidth}
        \centering
        \begin{minipage}{\textwidth}
            \centering
            \begin{minipage}{.24\textwidth}
                    \centering
                    \subcaption*{\scriptsize RMSE = 10.28mm \\  
                    RMSE(\%inliers) = 0.48mm (89\% ) }
                \end{minipage}
           \begin{minipage}{.24\textwidth}
                    \centering
                    \subcaption*{\scriptsize RMSE = 22.03mm \\  
                    RMSE(\%inliers) = 0.79mm (64\% ) }
                \end{minipage}
           \begin{minipage}{.24\textwidth}
                    \centering
                    \subcaption*{\scriptsize RMSE = 33.72mm \\  
                    RMSE(\%inliers) = 1.07mm (41\% ) }
                \end{minipage}
            \begin{minipage}{.24\textwidth}
                    \centering
                    \subcaption*{\scriptsize RMSE = 29.26mm \\  
                    RMSE(\%inliers) = 0.85mm (55\% ) }
                \end{minipage}
        \end{minipage}
    \end{minipage}  
  
    \caption{Comparison with a single-source structured light~\cite{xxm_2023_unified}. From the left to right: our result, the result of~\cite{xxm_2023_unified} when the LED at the center, left, or right corner of the LED array is on.}   
    \label{fig:cmp_shadow}
\end{figure}

%% file: figs/abl_pattern_num.tex
\begin{figure}[htbp]
    \centering
    \captionsetup[subfigure]{justification=centering}
    \begin{minipage}{\linewidth}
        \centering
        \begin{minipage}{\textwidth}
            \centering
            \begin{minipage}{.32\textwidth}
                \centering 
                \subcaption*{\small 36 patterns}
            \end{minipage}
            \begin{minipage}{.32\textwidth}
                \centering
                \subcaption*{\small 54 patterns}
            \end{minipage}
            \begin{minipage}{.32\textwidth}
                \centering
                \subcaption*{\small 72 patterns}
            \end{minipage}
        \end{minipage}
    \end{minipage}  

    \begin{minipage}{\linewidth}
        \centering
        \begin{minipage}{.32\linewidth}
            \centering
            \includegraphics[width=\linewidth]{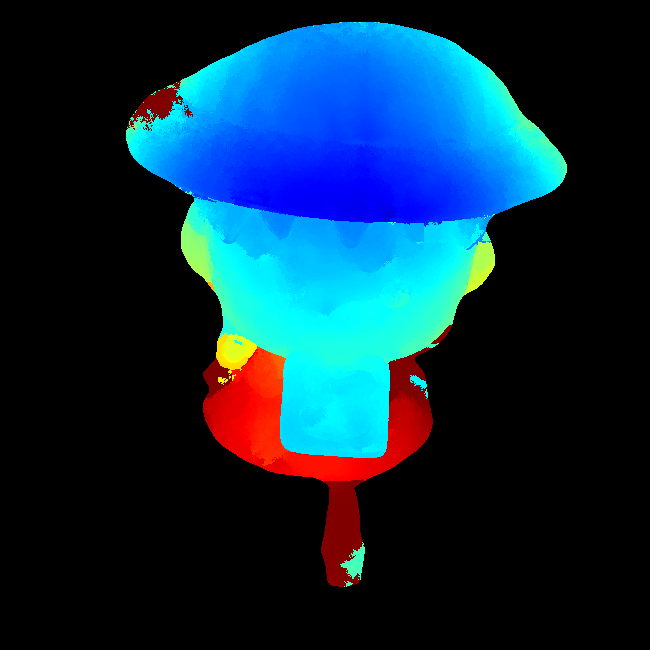}
        \end{minipage}
        \begin{minipage}{.32\linewidth}
            \centering
            \includegraphics[width=\linewidth]{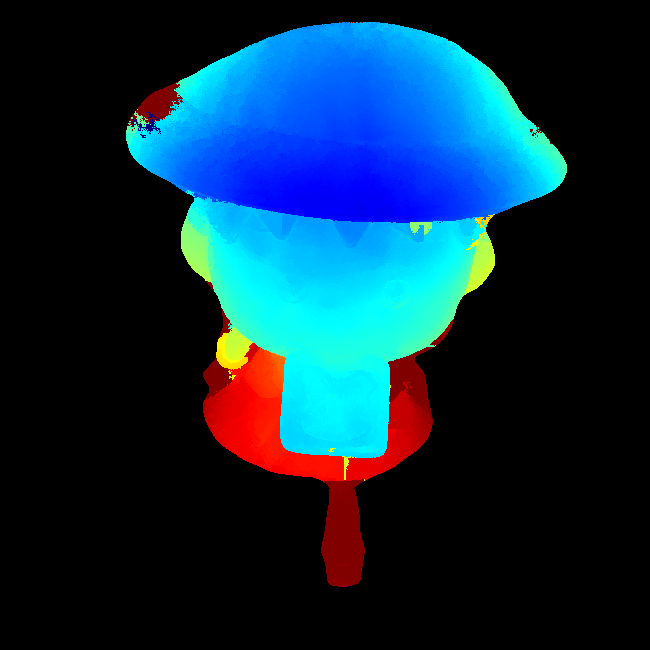}
        \end{minipage}
        \begin{minipage}{.32\linewidth}
            \centering
            \includegraphics[width=\linewidth]{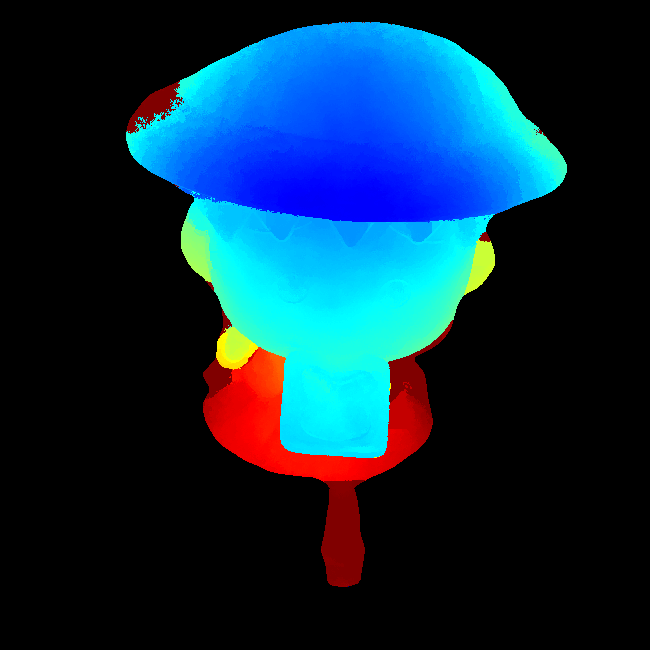}
        \end{minipage}
    \end{minipage}
    \begin{minipage}{\linewidth}
        \centering
        \begin{minipage}{\textwidth}
            \centering
            \begin{minipage}{.31\textwidth}
                    \centering
                    \subcaption*{\scriptsize RMSE = 4.95mm \\  
                    RMSE(\%inliers) = 0.46mm (93\% ) }
                \end{minipage}
            \begin{minipage}{.31\textwidth}
                    \centering
                    \subcaption*{\scriptsize RMSE = 4.94mm \\  
                     RMSE(\%inliers) = 0.42mm (93\% ) }
                \end{minipage}
            \begin{minipage}{.31\textwidth}
                    \centering
                    \subcaption*{\scriptsize RMSE = 4.78mm \\  
                     RMSE(\%inliers) = 0.42mm (94\% )}
                \end{minipage}
        \end{minipage}
    \end{minipage}  


    \caption{Impact of the total number of adaptive light/mask patterns over the depth quality.}   
    \label{fig:abl_pattern_num}
\end{figure}

%% file: figs/abl_sample_num.tex
\begin{figure}[htbp]
    \centering
    \captionsetup[subfigure]{justification=centering}
    \begin{minipage}{\linewidth}
        \centering
        \begin{minipage}{\textwidth}
            \centering
            \begin{minipage}{.30\textwidth}
                \centering 
                \subcaption*{\small $n_{\operatorname{sample}}$ = 100}
            \end{minipage}
            \begin{minipage}{.30\textwidth}
                \centering
                \subcaption*{\small $n_{\operatorname{sample}}$ = 300}
            \end{minipage}
            \begin{minipage}{.30\textwidth}
                \centering
                \subcaption*{\small $n_{\operatorname{sample}}$ = 600}
            \end{minipage}
        \end{minipage}
    \end{minipage}  
    \begin{minipage}{\linewidth}
        \center
        \begin{minipage}{.32\linewidth}
            \centering
            \includegraphics[width=\linewidth]{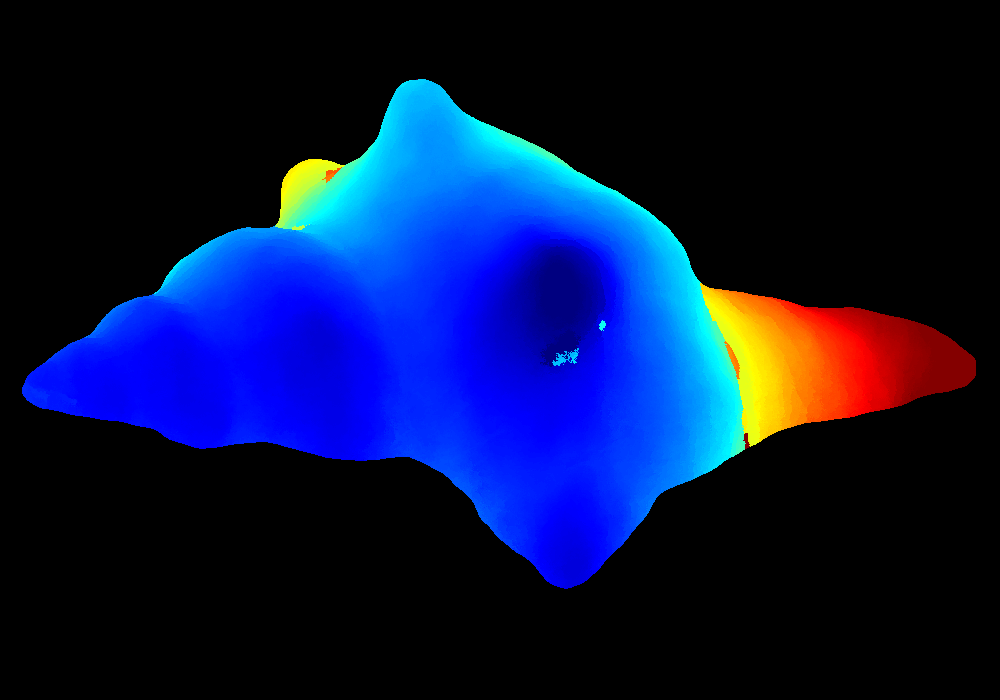}
        \end{minipage}
        \begin{minipage}{.32\linewidth}
            \centering
            \includegraphics[width=\linewidth]{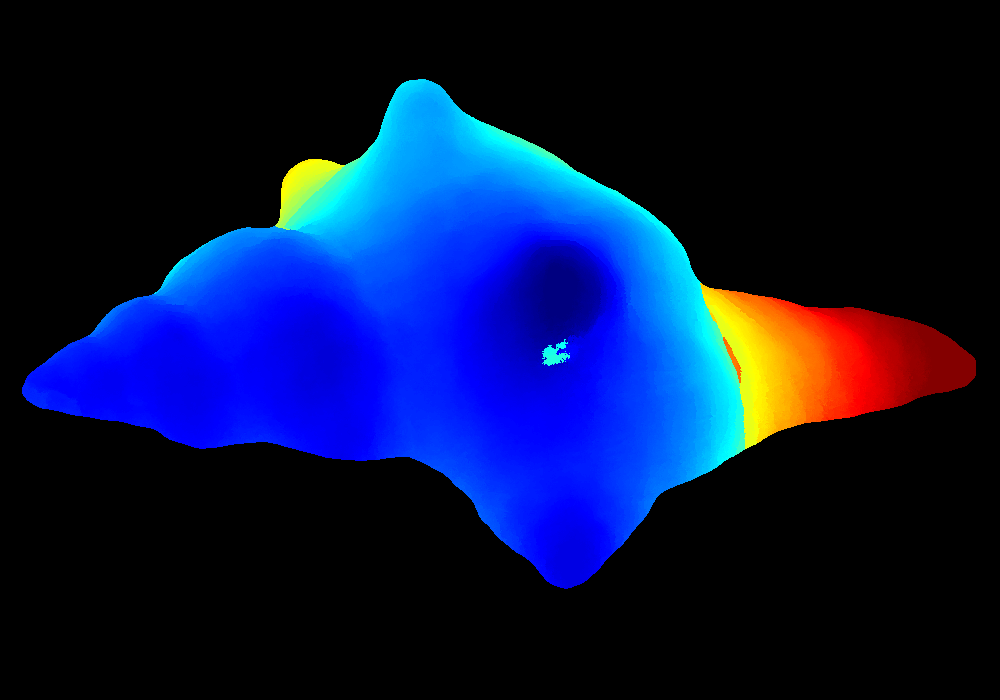}
        \end{minipage}
        \begin{minipage}{.32\linewidth}
            \centering
            \includegraphics[width=\linewidth]{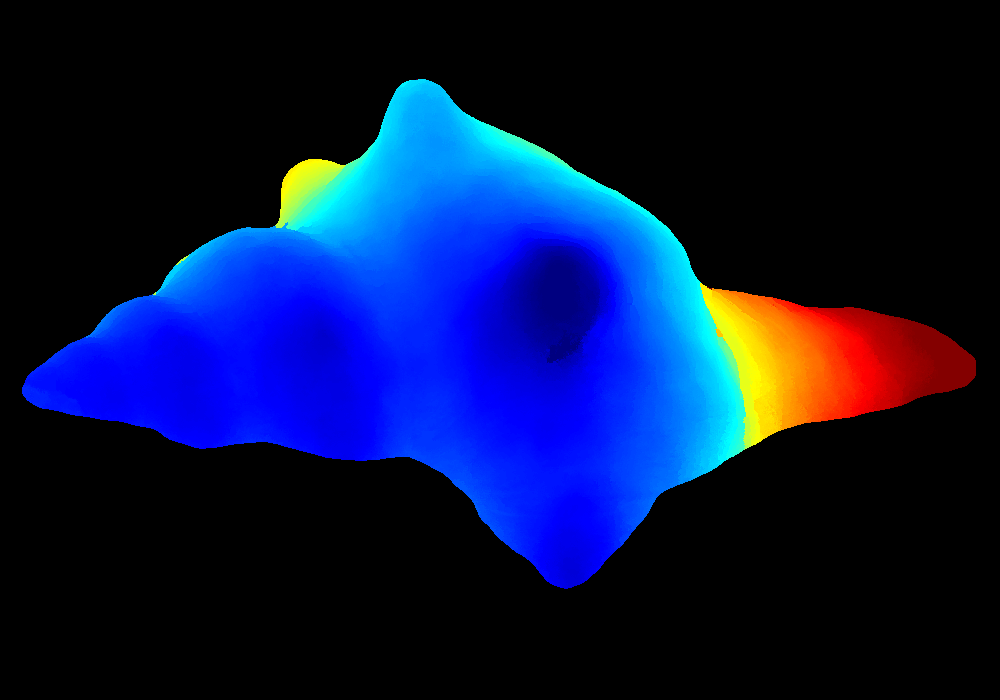}
        \end{minipage}
    \end{minipage}

    \begin{minipage}{\linewidth}
        \centering
        \begin{minipage}{\textwidth}
            \centering
            \begin{minipage}{.31\textwidth}
                    \centering
                    \subcaption*{\scriptsize RMSE = 1.87mm \\  
                    RMSE(\%inliers) = 0.28mm (98.6\% ) }
                \end{minipage}
             \begin{minipage}{.31\textwidth}
                    \centering
                    \subcaption*{\scriptsize RMSE = 1.79mm \\  
                    RMSE(\%inliers) = 0.27mm (98.7\% ) }
                \end{minipage}
            \begin{minipage}{.31\textwidth}
                    \centering
                    \subcaption*{\scriptsize RMSE = 1.75mm \\  
                    RMSE(\%inliers) = 0.26mm (99.1\% ) }
                \end{minipage}
        \end{minipage}
    \end{minipage}  
    \caption{Impact of $n_{\operatorname{sample}}$ over the depth quality.}   
    \label{fig:abl_sample_num}
\end{figure}


%% file: figs/abl_balance.tex
\begin{figure}[htbp]
    \centering
    \captionsetup[subfigure]{justification=centering}
    \begin{minipage}{\linewidth}
        \centering
        \begin{minipage}{\textwidth}
            \centering
            \begin{minipage}{.31\textwidth}
                \centering 
                \subcaption*{\small $n_{\operatorname{batch}}$  = 2}
            \end{minipage}
            \begin{minipage}{.31\textwidth}
                \centering
                \subcaption*{\small $n_{\operatorname{batch}}$  = 3}
            \end{minipage}
            \begin{minipage}{.31\textwidth}
                \centering
                \subcaption*{\small $n_{\operatorname{batch}}$  = 6}
            \end{minipage}
        \end{minipage}
    \end{minipage}  
    \begin{minipage}{\linewidth}
        \centering
        \begin{minipage}{.32\linewidth}
            \centering
            \includegraphics[width=\linewidth]{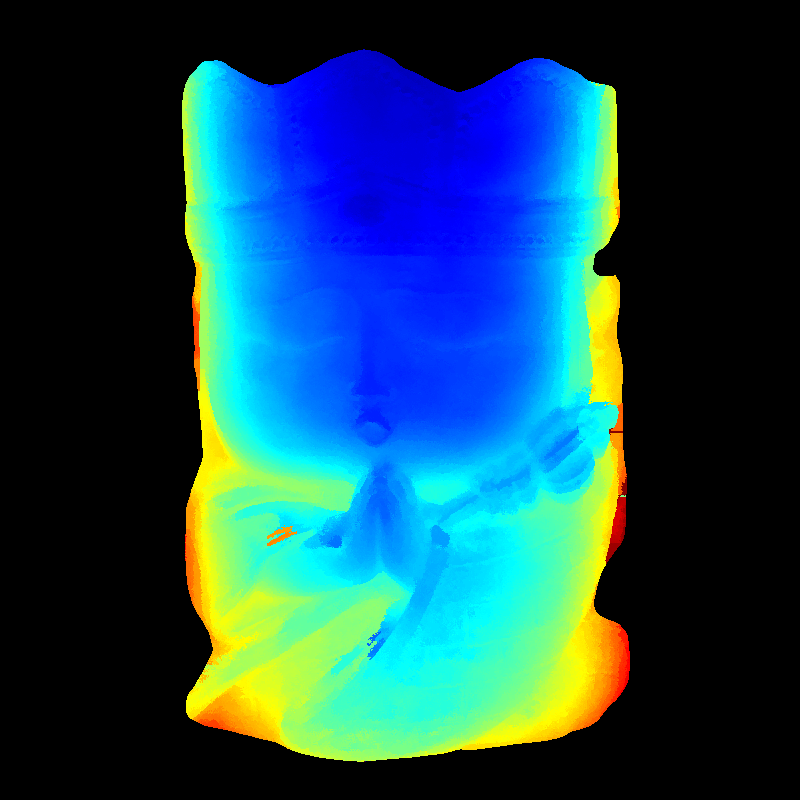}
        \end{minipage}
        \begin{minipage}{.32\linewidth}
            \centering
            \includegraphics[width=\linewidth]{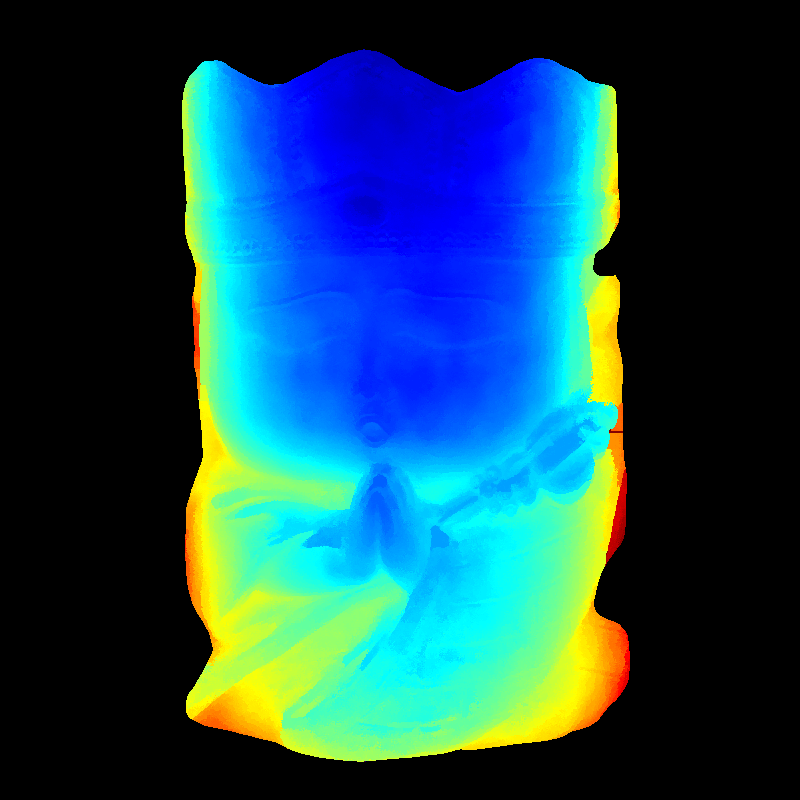}
        \end{minipage}
        \begin{minipage}{.32\linewidth}
            \centering
            \includegraphics[width=\linewidth]{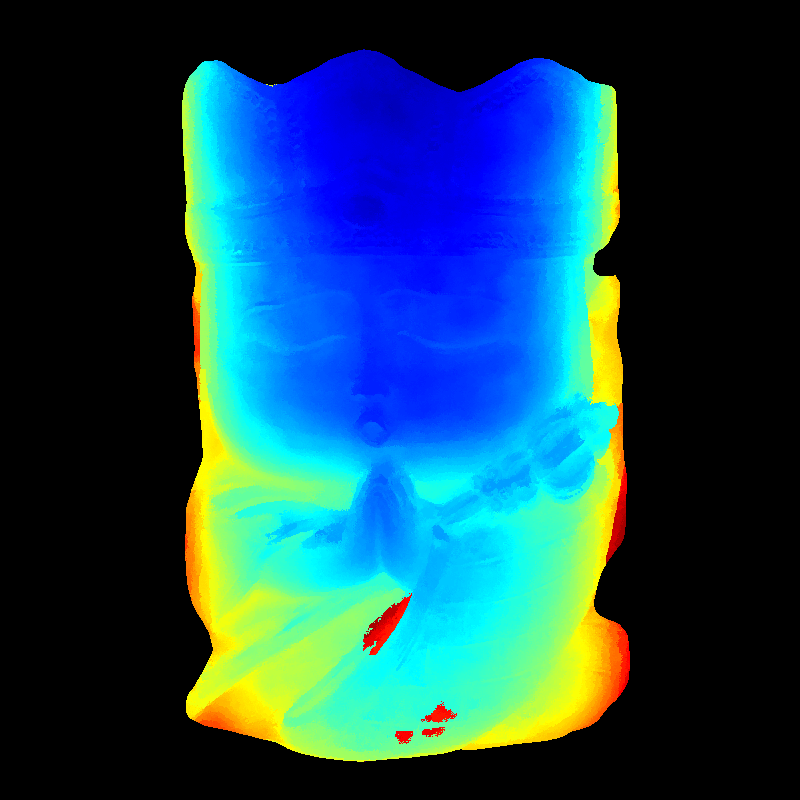}
        \end{minipage}
    \end{minipage}
    \begin{minipage}{\linewidth}
        \centering
        \begin{minipage}{\textwidth}
            \centering
               \begin{minipage}{.31\textwidth}
                    \centering
                    \subcaption*{\scriptsize RMSE = 3.57mm \\  
                    RMSE(\%inliers) = 0.43mm (97.9\% ) }
                \end{minipage}
            \begin{minipage}{.31\textwidth}
                    \centering
                    \subcaption*{\scriptsize RMSE = 3.54mm \\  
                    RMSE(\%inliers) = 0.40mm (98.3\% ) }
                \end{minipage}
            \begin{minipage}{.31\textwidth}
                    \centering
                    \subcaption*{\scriptsize RMSE = 3.59mm \\  
                    RMSE(\%inliers) = 0.45mm (97.6\% ) }
                \end{minipage}
        \end{minipage}
    \end{minipage}  
    
    \caption{Impact of the number of simultaneously optimized next patterns ($n_{\operatorname{batch}}$) over the depth quality, with the same total acquisition time.} 
    \label{fig:abl_balance}
\end{figure}

%% file: figs/abl_best_peak.tex
\begin{figure}[htbp]
    \centering
    \captionsetup[subfigure]{justification=centering}
    \begin{minipage}{\linewidth}
        \centering
        \begin{minipage}{\textwidth}
            \centering
            \begin{minipage}{.30\textwidth}
                \centering 
                \subcaption*{\small $n_{\operatorname{peak}}$ = 2}
            \end{minipage}
            \begin{minipage}{.30\textwidth}
                \centering
                \subcaption*{\small $n_{\operatorname{peak}}$ = 3}
            \end{minipage}
            \begin{minipage}{.30\textwidth}
                \centering
                \subcaption*{\small $n_{\operatorname{peak}}$ = 6}
            \end{minipage}
        \end{minipage}
    \end{minipage}  
    \begin{minipage}{\linewidth}
        \center
        \begin{minipage}{.32\linewidth}
            \centering
            \includegraphics[width=\linewidth]{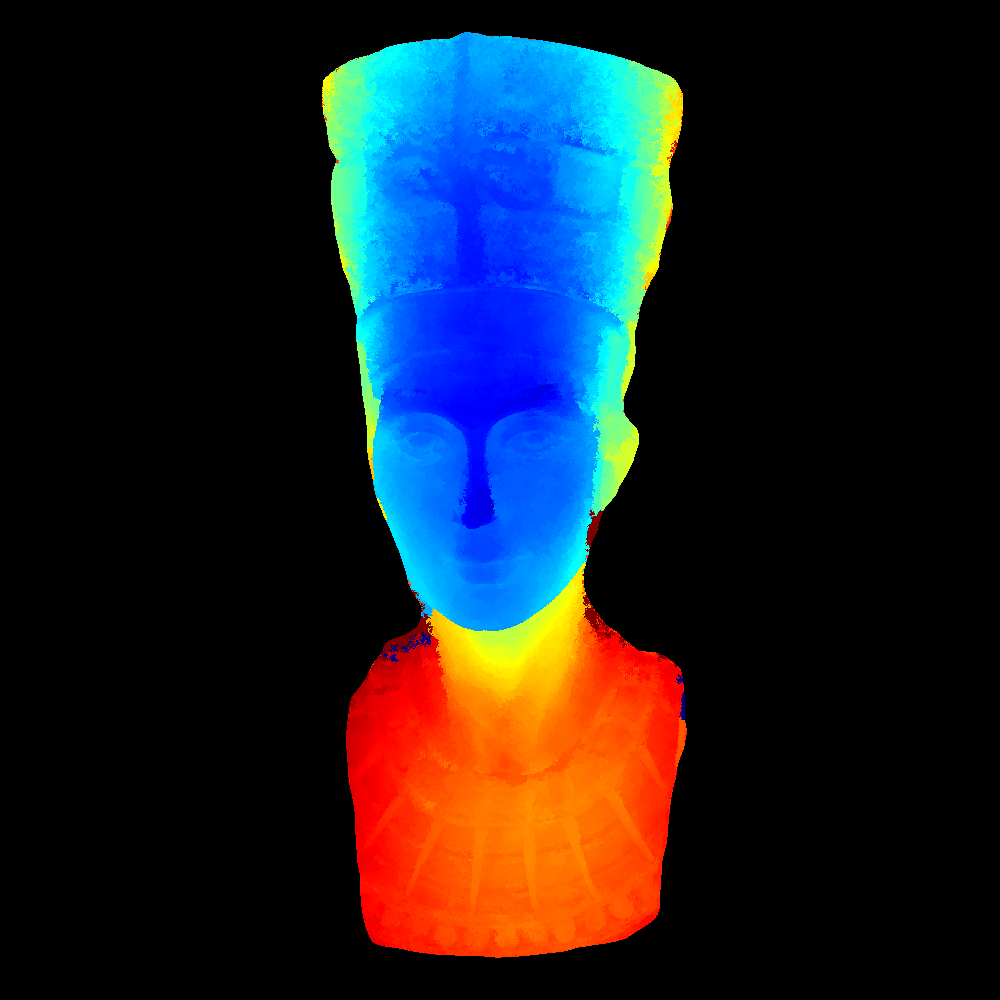}
        \end{minipage}
        \begin{minipage}{.32\linewidth}
            \centering
            \includegraphics[width=\linewidth]{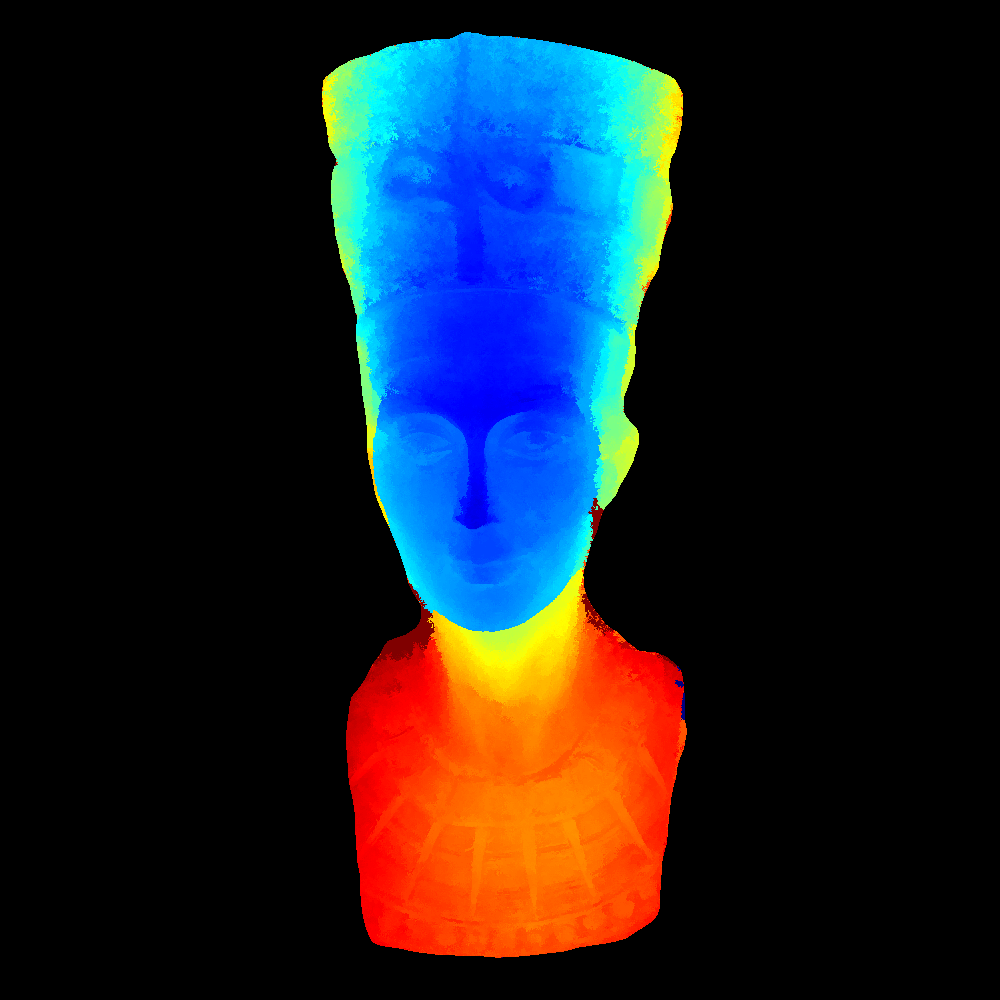}
        \end{minipage}
        \begin{minipage}{.32\linewidth}
            \centering
            \includegraphics[width=\linewidth]{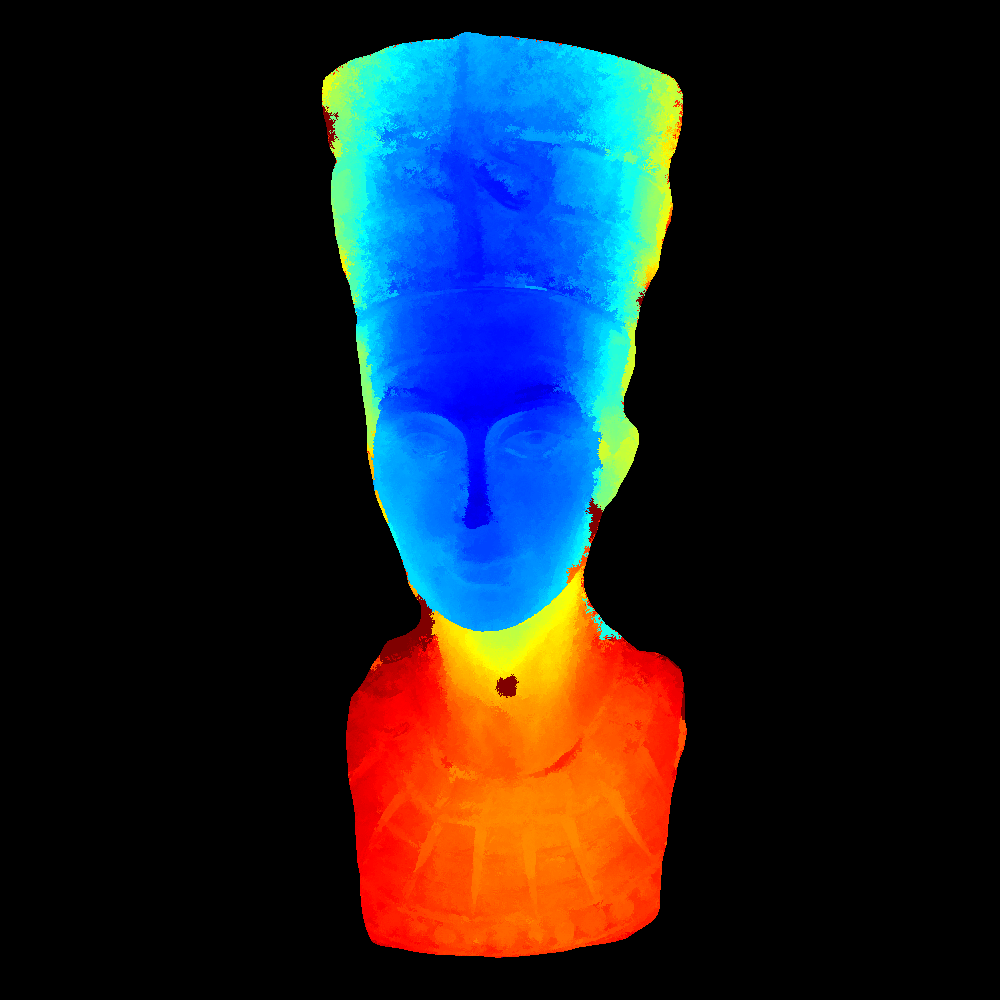}
        \end{minipage}
    \end{minipage}

    \begin{minipage}{\linewidth}
        \centering
        \begin{minipage}{\textwidth}
            \centering
            \begin{minipage}{.31\textwidth}
                    \centering
                    \subcaption*{\scriptsize RMSE = 2.40mm \\  
                    RMSE(\%inliers) = 0.48mm (97.4\% ) }
                \end{minipage}
             \begin{minipage}{.31\textwidth}
                    \centering
                    \subcaption*{\scriptsize RMSE = 2.30mm \\  
                    RMSE(\%inliers) = 0.47mm (97.9\% ) }
                \end{minipage}
            \begin{minipage}{.31\textwidth}
                    \centering
                    \subcaption*{\scriptsize RMSE = 2.40mm \\  
                    RMSE(\%inliers) = 0.52mm (96.9\% ) }
                \end{minipage}
        \end{minipage}
    \end{minipage}  
    \caption{Impact of $n_{\operatorname{peak}}$ over the depth quality.}    
    \label{fig:abl_best_peak}
\end{figure}

%% file: figs/abl_bin_num.tex
\begin{figure}[htbp]
    \centering
    \captionsetup[subfigure]{justification=centering}
    \begin{minipage}{\linewidth}
        \centering
        \begin{minipage}{\textwidth}
            \centering
            \begin{minipage}{.30\textwidth}
                \centering 
                \subcaption*{\small $n_{\operatorname{bin}}$ = 50}
            \end{minipage}
            \begin{minipage}{.30\textwidth}
                \centering
                \subcaption*{\small $n_{\operatorname{bin}}$ = 75}
            \end{minipage}
            \begin{minipage}{.30\textwidth}
                \centering
                \subcaption*{\small $n_{\operatorname{bin}}$ = 100}
            \end{minipage}
        \end{minipage}
    \end{minipage}  
    \begin{minipage}{\linewidth}
        \center
        \begin{minipage}{.32\linewidth}
            \centering
            \includegraphics[width=\linewidth]{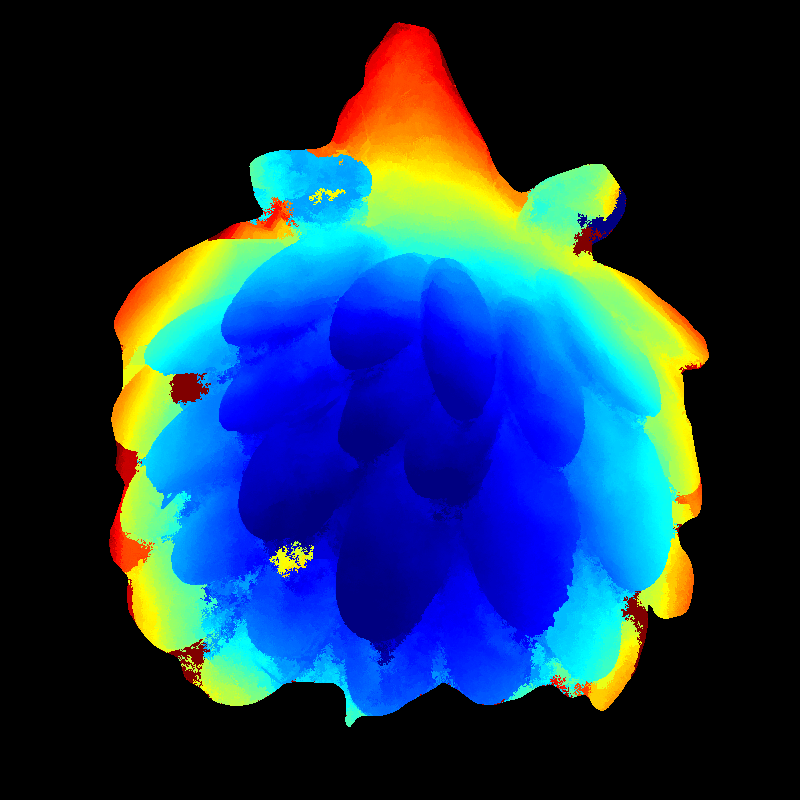}
        \end{minipage}
        \begin{minipage}{.32\linewidth}
            \centering
            \includegraphics[width=\linewidth]{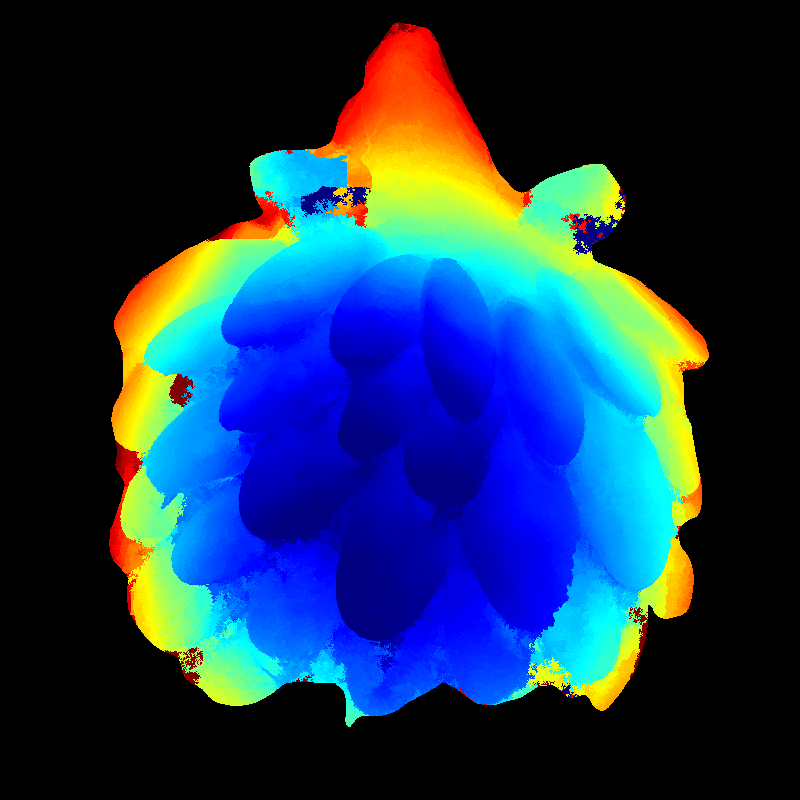}
        \end{minipage}
        \begin{minipage}{.32\linewidth}
            \centering
            \includegraphics[width=\linewidth]{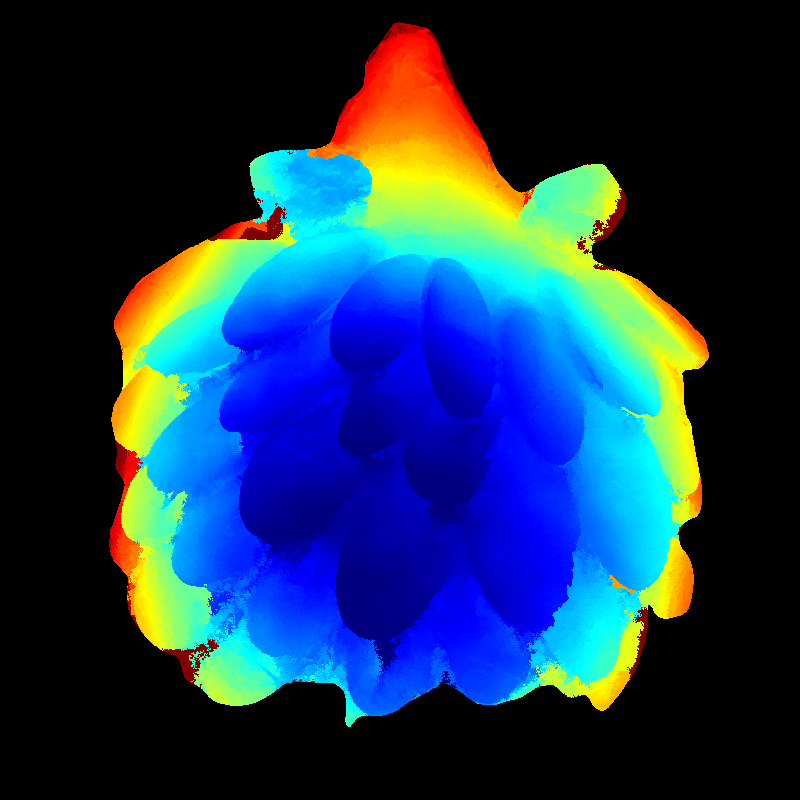}
        \end{minipage}
    \end{minipage}

    \begin{minipage}{\linewidth}
        \centering
        \begin{minipage}{\textwidth}
            \centering
            \begin{minipage}{.31\textwidth}
                    \centering
                    \subcaption*{\scriptsize RMSE = 7.58mm \\  
                    RMSE(\%inliers) = 0.48mm (90.2\% ) }
                \end{minipage}
             \begin{minipage}{.31\textwidth}
                    \centering
                    \subcaption*{\scriptsize RMSE = 7.53mm \\  
                    RMSE(\%inliers) = 0.48mm (90.3\% ) }
                \end{minipage}
            \begin{minipage}{.31\textwidth}
                    \centering
                    \subcaption*{\scriptsize RMSE = 7.30mm \\  
                    RMSE(\%inliers) = 0.48mm (91.0\% ) }
                \end{minipage}
        \end{minipage}
    \end{minipage}  
    \caption{Impact of $n_{\operatorname{bin}}$ over the depth quality.}
     \label{fig:abl_bin_num}
\end{figure}


%% file: figs/abl_mipmap.tex
\begin{figure}[htbp]
    \centering
    \captionsetup[subfigure]{justification=centering}
    \begin{minipage}{\linewidth}
        \centering
        \begin{minipage}{\textwidth}
            \centering
            \begin{minipage}{.32\textwidth}
                \centering 
                \subcaption*{\small 512$\times$256}
            \end{minipage}
            \begin{minipage}{.32\textwidth}
                \centering
                \subcaption*{\small 254$\times$128}
            \end{minipage}
            \begin{minipage}{.32\textwidth}
                \centering
                \subcaption*{\small 127$\times$64}
            \end{minipage}
        \end{minipage}
    \end{minipage}  
    \begin{minipage}{\linewidth}
        \center
        \begin{minipage}{.32\linewidth}
            \centering
            \includegraphics[width=\linewidth]{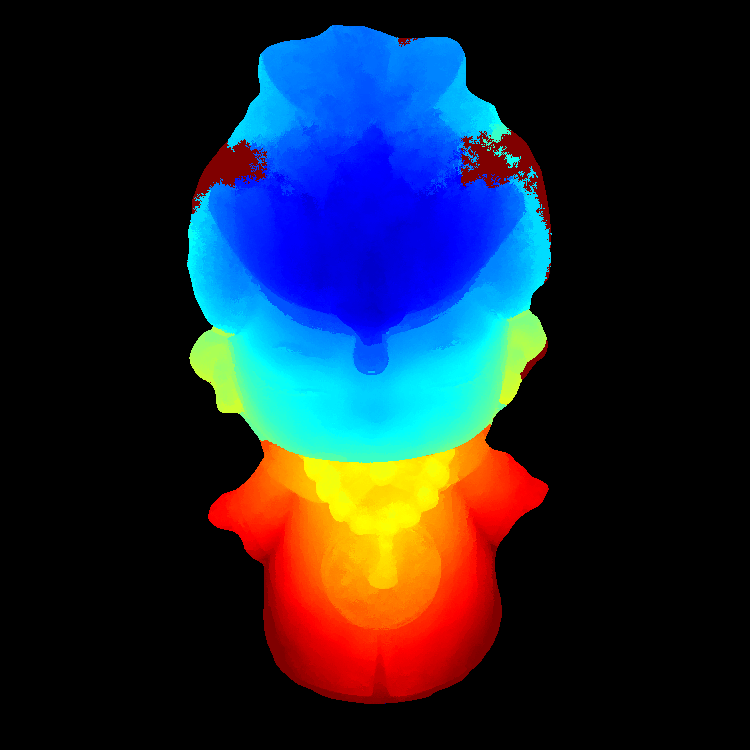}
        \end{minipage}
        \begin{minipage}{.32\linewidth}
            \centering
            \includegraphics[width=\linewidth]{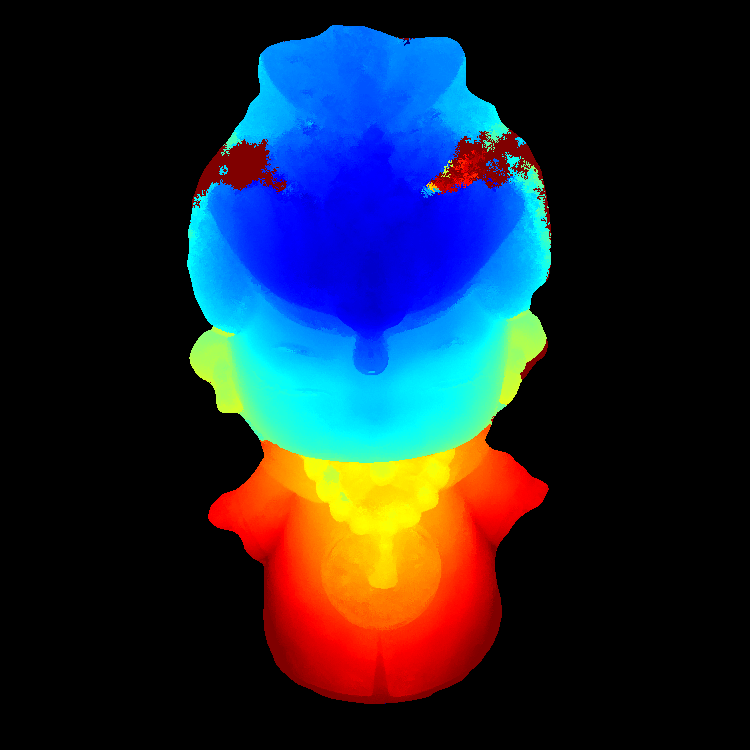}
        \end{minipage}
        \begin{minipage}{.32\linewidth}
            \centering
            \includegraphics[width=\linewidth]{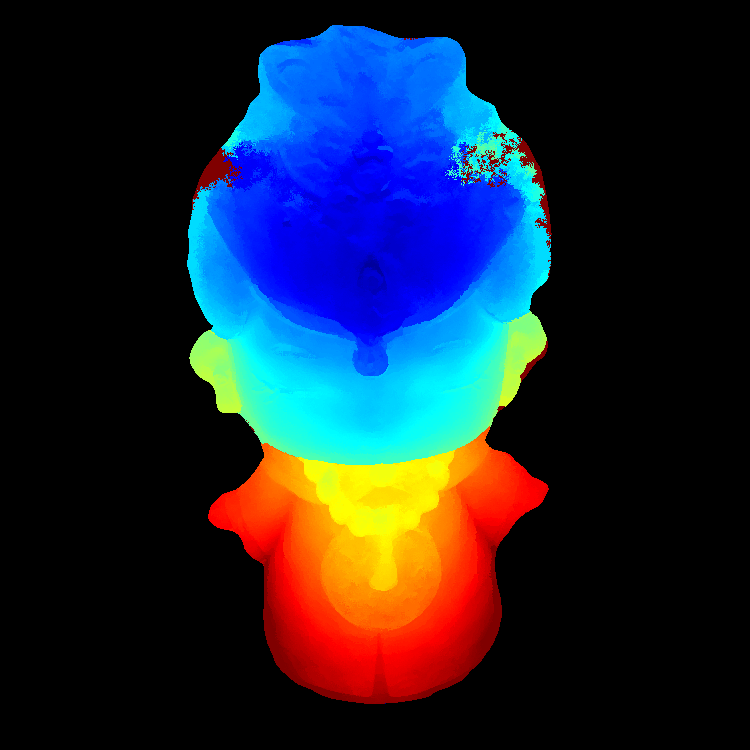}
        \end{minipage}
    \end{minipage}

    \begin{minipage}{\linewidth}
        \centering
        \begin{minipage}{\textwidth}
            \centering
            \begin{minipage}{.31\textwidth}
                    \centering
                    \subcaption*{\scriptsize RMSE = 4.20mm \\  
                    RMSE(\%inliers) = 0.38mm (94.4\% ) } 
                \end{minipage}
             \begin{minipage}{.31\textwidth}
                    \centering
                    \subcaption*{\scriptsize RMSE = 4.17mm \\  
                    RMSE(\%inliers) = 0.40mm (94.5\% ) }
                \end{minipage}
            \begin{minipage}{.31\textwidth}
                    \centering
                    \subcaption*{\scriptsize RMSE = 3.71mm \\  
                    RMSE(\%inliers) = 0.67mm (93.3\% ) }
                \end{minipage}
        \end{minipage}
    \end{minipage}  
    \caption{Impact of the image resolution for computing adaptive patterns (listed on top) over the depth quality. Note that the same reconstruction resolution is used in all cases.}
    \label{fig:abl_mipmap}
\end{figure}
